\documentclass[final]{core/cvpr}

\usepackage{times}
\usepackage{epsfig}
\usepackage{graphicx}
\usepackage{amsmath}
\usepackage{amssymb}

\usepackage{mathtools}
\usepackage{balance}
\usepackage{core/visinfdefs}
\newcommand{\myparagraph}[1]{\smallskip\noindent\textbf{#1}}

\renewcommand{\cf}{\emph{cf}\onedot}

\usepackage{etoolbox,siunitx}
\usepackage[normalem]{ulem}
\robustify\uline
\robustify\bfseries
\usepackage[super]{nth}
\usepackage{newtxtt}

\def\Uline#1{#1\llap{\uline{\phantom{#1}}}}

\let\oldFootnote\footnote
\newcommand\nextToken\relax

\renewcommand\footnote[1]{%
    \oldFootnote{#1}\futurelet\nextToken\isFootnote}

\newcommand\isFootnote{%
    \ifx\footnote\nextToken\textsuperscript{,}\fi}

\usepackage{pifont}
\newcommand{\cmark}{\ding{51}}

\usepackage{booktabs}
\usepackage{tabularx}
\usepackage{multirow}
\usepackage{makecell}

\usepackage{xcolor}
\definecolor{customgray}{rgb}{0.31, 0.31, 0.31}
\definecolor{customblue}{rgb}{0.34, 0.70, 0.91}
\definecolor{customred}{rgb}{0.83, 0.31, 0}
\definecolor{customyellow}{rgb}{0.94, 0.89, 0.26}

\usepackage[format=plain,labelformat=simple,labelsep=period,font=small,skip=4pt,compatibility=false]{caption}
\usepackage[font=scriptsize,skip=2pt,subrefformat=parens]{subcaption}

\usepackage[pagebackref=true,breaklinks=true,colorlinks,bookmarks=false]{hyperref}
\usepackage[capitalize]{cleveref}
\crefname{section}{Sec.}{Section}

\usepackage[inline]{enumitem}

\def\cvprPaperID{2077}
\def\confYear{CVPR 2021}

\usepackage{fancyhdr}
\usepackage{setspace}

\begin{document}

%% Main paper title
\title{Self-Supervised Multi-Frame Monocular Scene Flow}

\author{Junhwa Hur$^1$ \qquad\qquad Stefan Roth$^{1,2}$\\
$\ ^1$Department of Computer Science, TU Darmstadt \hspace{1cm} $\ ^2$ hessian.AI}
\maketitle
\thispagestyle{fancy}

%% Abstract
\begin{abstract}
   % !TEX root = ../arxiv.tex
Estimating 3D scene flow from a sequence of monocular images has been gaining increased attention due to the simple, economical capture setup.
Owing to the severe ill-posedness of the problem, the accuracy of current methods has been limited, especially that of efficient, real-time approaches.
In this paper, we introduce a multi-frame monocular scene flow network based on self-supervised learning, improving the accuracy over previous networks while retaining real-time efficiency.
Based on an advanced two-frame baseline with a split-decoder design, we propose \emph{(i)} a multi-frame model using a triple frame input and convolutional LSTM connections, \emph{(ii)} an occlusion-aware census loss for better accuracy, and \emph{(iii)} a gradient detaching strategy to improve training stability.
On the KITTI dataset, we observe state-of-the-art accuracy among monocular scene flow methods based on self-supervised learning.

\end{abstract}

%% Main paper body
% !TEX root = ../arxiv.tex
\section{Introduction}
\label{sec:introduction}

Scene flow estimation, that is the task of estimating 3D structure and 3D motion of a dynamic scene, has been receiving increased attention together with a growing interest and demand for autonomous navigation systems.
Many approaches have been proposed, based on various input data such as stereo images \cite{Behl:2017:BBS,Jiang:2019:SAS,Ma:2019:DRI,Schuster:2018:SFF,Vogel:2015:3SF}, RGB-D sequences \cite{Hornacek:2014:SF6,Lv:2018:LRD,Qiao:2018:SFN}, or 3D point clouds \cite{Behl:2019:PFN,Gu:2019:HPL,Liu:2019:FN3,Wu:2020:PPN}.

Recently, \emph{monocular} scene flow approaches \cite{Brickwedde:2019:MSF,Hur:2020:SSM,Luo:2019:EPC,Yang:2020:UOF} have shown the possibility of estimating 3D scene flow from a pair of temporally consecutive monocular frames only, obviating complicated, expensive sensor setups such as a stereo rig, RGB-D sensors, or a LiDAR scanner.
Only a simple, affordable monocular camera is needed.
The availability of ground-truth data has been another key challenge for scene flow estimation in general.
To address this, methods based on self-supervised learning \cite{Hur:2020:SSM,Luo:2019:EPC} have shown it possible to train CNNs for jointly estimating depth and scene flow without expensive 3D annotations.
Yet, their accuracy is bounded by the limitation of only using two frames as input, their underlying proxy loss, and training instabilities due to the difficulty of optimizing CNNs for multiple tasks, particularly in a self-supervised manner \cite{Luo:2019:EPC}.

Semi-supervised methods have demonstrated promising accuracy by combining CNNs with energy minimization \cite{Brickwedde:2019:MSF} or sequentially estimating optical flow and depth to infer 3D scene flow \cite{Yang:2020:UOF}.
Those methods, however, do not reach real-time efficiency due to iterative energy minimization \cite{Brickwedde:2019:MSF} or the additional processing time from pre-computing depth and optical flow \cite{Yang:2020:UOF} beforehand.
Yet, computational efficiency is important for autonomous navigation applications.

In this paper, we introduce a self-supervised monocular scene flow approach that substantially advances the previously most accurate real-time method of Hur~\etal~\cite{Hur:2020:SSM}, while keeping its advantages (\eg, computational efficiency and training on unlabeled data).
We first analyze the technical design, revealing some limitations, and propose an improved two-frame backbone network to overcome them.
Next, we introduce a multi-frame formulation that temporally propagates the estimate from the previous time step for more accurate and reliable results in consecutive frames.
Previous monocular methods \cite{Brickwedde:2019:MSF,Hur:2020:SSM,Luo:2019:EPC,Yang:2020:UOF} utilize only two frames as input, which is a minimal setup for demonstrating the underlying ideas.
In contrast, our approach is the first to demonstrate how to exploit multiple consecutive frames, which are naturally available in most real-world scenarios.

\begin{figure}[t]
\centering
\includegraphics[width=\linewidth]{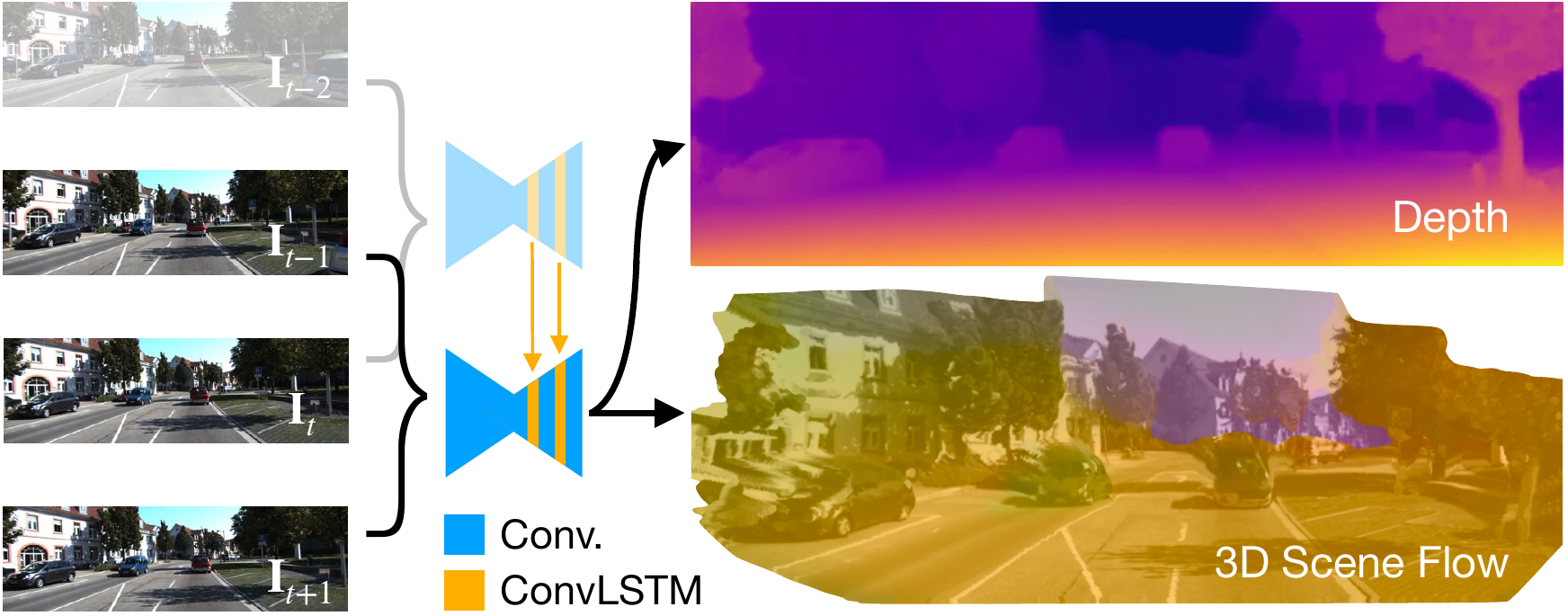}
\caption{\textbf{Our multi-frame monocular scene flow approach} inputs three frames and estimates depth and scene flow (visualized in 3D \cite{Zhou:2018:O3D} with scene flow color coding in \cref{supp:sec:color_coding}).}
\label{fig:teaser}
%\vspace{-0.5em}
\end{figure}

We make the following main contributions:
\begin{enumerate*}[label=(\roman*), font=\itshape]
  \item We uncover some limitations of the baseline architecture of \cite{Hur:2020:SSM} and introduce an advanced two-frame basis with a \emph{split-decoder design}.
  Contradicting the finding of \cite{Hur:2020:SSM} on using a single joint decoder, our split decoder is not only faster and more stable to train, but delivers competitive accuracy.
  \item Next, we introduce our \emph{multi-frame network} based on overlapping triplets of frames as input and temporally propagating the previous estimate via a convolutional LSTM \cite{Hochreiter:1997:LSTM, Shi:2015:CLSTM} (\cf~\cref{fig:teaser}).
  \item Importantly, we propagate the hidden states using \emph{forward warping}, which is especially beneficial for handling occlusion; it is also more stable to train than using backward warping in a self-supervised setup.
  \item We propose an \emph{occlusion-aware census transform} to take occlusion cues into account, providing a more robust measure for brightness difference as a self-supervised proxy loss.
  \item Lastly, we introduce a \emph{gradient detaching strategy} that improves not only the accuracy but also the training stability, which self-supervised methods for multi-task learning can benefit from.
  We successfully validate all our design choices through an ablation study.
\end{enumerate*}

Training on KITTI raw \cite{Geiger:2013:VMR} in a self-supervised manner, our model improves the accuracy of the direct baseline of \cite{Hur:2020:SSM} by \num{15.4}\%.
Owing to its self-supervised design, our approach also generalizes to different datasets.
We can optionally perform (semi-)supervised fine-tuning on \emph{KITTI Scene Flow Training} \cite{Menze:2015:J3E,Menze:2018:OSF}, where we also outperform \cite{Hur:2020:SSM}, reducing the accuracy gap to semi-supervised methods \cite{Brickwedde:2019:MSF,Yang:2020:UOF} while remaining many times more efficient.

% !TEX root = ../arxiv.tex
\section{Related Work}
\label{sec:relatedwork}

\paragraph{Scene flow.}
First introduced by Vedula~\etal~\cite{Vedula:1999:TDS,Vedula:2005:TDS}, scene flow aims to estimate dense 3D motion for each 3D point in the scene.
Depending on the available input data, approaches differ in their objectives and formulations.

Stereo-based methods \cite{Aleotti:2020:LEE,Behl:2017:BBS,Ma:2019:DRI,Schuster:2018:SFF,Wedel:2011:3SF,Zhang:2001:O3S} estimate a disparity map between stereo pairs to recover the 3D scene structure as well as a dense 3D scene flow field between a temporal frame pair.
Earlier work mainly uses variational formulations or graphical models, which yields limited accuracy \cite{Basha:2013:MVS,Huguet:2007:AVM} and/or slow runtime \cite{Menze:2015:OSF,Ren:2017:CSF,Vogel:2014:VC3,Vogel:2013:PRS,Vogel:2015:3SF}.
Recent CNN-based methods \cite{Ilg:2018:OMD,Jiang:2019:SAS,Mayer:2016:ALD,Saxena:2019:PDO} overcome these limitations: they attain state-of-the-art accuracy in real time by training CNNs on large synthetic datasets followed by fine-tuning on the target domain in a supervised manner.
Un-/self-supervised approaches \cite{Lee:2019:LRF,Liu:2019:ULS,Wang:2019:UOS}  aim to overcome the dependency on accurate, diverse labeled data, which is not easy to obtain.
Approaches using sequences of RGB-D images \cite{Goerlitz:2019:PRS,Hadfield:2011:KDP,Hornacek:2014:SF6,Lv:2018:LRD,Qiao:2018:SFN,Quiroga:2014:DSR} or 3D point clouds \cite{Behl:2019:PFN,Gu:2019:HPL,Liu:2019:FN3,Mittal:2020:JGW,Wang:2020:FN3,Wu:2020:PPN} have been also proposed, exploiting an already given 3D sparse point input.

In contrast, our approach jointly estimates both 3D scene structure and dense 3D scene flow from a monocular image sequence alone, which is a \emph{more practical, yet much more challenging} setup.

\myparagraph{Monocular scene flow.}
Estimating scene flow using a monocular image sequence has been gaining increased attention.
Multi-task CNN approaches \cite{Chen:2019:SSL,Lai:2019:BSM,Luo:2019:EPC,Ranjan:2019:CCJ,Yang:2018:EPC,Yin:2018:GNU,Zhu:2019:RMD,Zou:2018:DFN} jointly predict optical flow, depth, and camera motion from a monocular sequence;
scene flow can be reconstructed from those outputs.
However, such approaches have a critical limitation in that they cannot recover scene flow for occluded pixels.
Brickwedde~\etal~\cite{Brickwedde:2019:MSF} propose to combine CNNs for monocular depth prediction with an energy-based formulation for estimating scene flow from the given depth cue.
Yang~\etal~\cite{Yang:2020:UOF} introduce an integrated pipeline that obtains scene flow from given optical flow and depth cues via determining motion in depth from observing changes in object sizes.
Using a single network, Hur~\etal~\cite{Hur:2020:SSM} directly estimate depth and scene flow with a joint decoder design, trained in a self-supervised manner.

All above methods \cite{Brickwedde:2019:MSF,Hur:2020:SSM,Yang:2020:UOF} are limited to using two frames.
In contrast, we demonstrate how to leverage \emph{multiple consecutive frames} for more accurate and consistent results, which is desirable in real applications.

\myparagraph{Multi-frame estimation.}
Multi-frame approaches to optical flow typically exploit a constant velocity or acceleration assumption \cite{Black:1991:RDM,Garg:2013:AVA,Janai:2017:SFE,Kennedy:2015:OFG,Ricco:2012:DLM} to encourage temporally smooth and reliable estimates. % Volz:2011:MTC,Werlberger:2009:AHO
Using CNNs, propagation approaches have shown how to exploit previous predictions for the current time step, either by explicitly fusing the two outputs \cite{Maurer:2018:PFL,Ren:2019:FAM} or using them as input for the current estimation \cite{Neoral:2018:COO}.
Better temporal consistency has also been achieved using a bi-directional cost volume \cite{Janai:2018:ULM,Liu:2019:SFS} and convolutional LSTMs \cite{Godet:2020:STA,Guan:2019:ULO}.
%(\ie~from $t$ to $t-1$ and from $t$ to $t+1$)
Overall, these multi-frame approaches improve the optical flow accuracy, especially for occluded or out-of-bound areas.

For scene flow (stereo or RGB-D based), relatively few multi-frame methods have been introduced so far, all using classical energy minimization.
A consistent rigid motion assumption has been proposed for temporal consistency \cite{Neoral:2017:OSF,Vogel:2014:VC3}.
Other approaches include jointly estimating camera pose and motion segmentation \cite{Taniai:2017:FMF}, matching and visibility reasoning among multiple frames \cite{Schuster:2020:SFF}, or an integrated energy formulation \cite{Golyanik:2017:MSF}.
These methods are robust against outliers and occlusion, improving the accuracy; yet, their runtime is slow due to iterative energy minimization.

We introduce a \emph{CNN-based multi-frame scene flow approach in the challenging monocular setup}, ensuring real-time efficiency.
Building on the two-frame network of \cite{Hur:2020:SSM}, our method utilizes a bi-directional cost volume with convolutional LSTM connections, ensuring temporal consistency through overlapping frame triplets and temporal propagation of intermediate outputs (\cf~\cref{fig:teaser}).
Moreover, we propose a forward-warping strategy for LSTMs.

% !TEX root = ../arxiv.tex
\section{Multi-frame Monocular Scene Flow}
\label{sec:approach}

Given $N$ temporally consecutive frames, $\{ \mv{I}_{t-(N-2)}$, $\mv{I}_{t-(N-1)}$, ..., $\mv{I}_t$, $\mv{I}_{t+1} \}$,
our main objective is to estimate 3D surface points $\mv{P} = (P_x, P_y, P_z)$ for each pixel $\mv{p} = (p_x, p_y)$ in the reference frame $\mv{I}_t$ and the 3D scene flow $\mv{s}=(s_x, s_y, s_z)$ of each 3D point to the target frame $\mv{I}_{t+1}$.

\subsection{Refined backbone architecture}
\label{sec:approach:architecture}

\paragraph{Advanced two-frame baseline.}
Our network architecture is based on the integrated two-frame network of Hur~\etal~\cite{Hur:2020:SSM}, which uses PWC-Net \cite{Sun:2018:PWC} as a basis and runs in real time.
The network constructs a feature pyramid for each input frame, calculates the cost volume, and estimates the residual scene flow and disparity with a joint decoder over the pyramid levels.
While maintaining the core backbone, we first investigate whether recent advances in self-supervised optical flow can be carried over to monocular 3D scene flow.

Jonschkowski~\etal~\cite{Jonschkowski:2020:WMU} systematically analyze the key factors for highly accurate self-supervised optical flow, identifying crucial steps such as cost volume normalization, level dropout, data distillation, using a square resolution, \etc
While we do not aim for a comprehensive review of such factors in the context of monocular scene flow, we performed a simple empirical study of their key findings.\footnote{\label{footnote_supp1}See \cref{supp:sec:architecture}.}
We found cost volume normalization and using one less pyramid level (\ie~$6$ instead of $7$) to be helpful, and employ them for our advanced baseline.
Other findings were less effective for monocular scene flow;
hence we do not adopt them.

Moreover, we observed that the context network, a post-processing module with dilated convolutions \cite{Sun:2018:PWC}, is a source of training instability in the self-supervised setup.\textsuperscript{\ref{footnote_supp1}}
We thus discard the context network for stable convergence.

\begin{figure}[t]
\centering
\subcaptionbox{A single joint decoder \cite{Hur:2020:SSM} \label{fig:decoder_joint}}[0.4\linewidth]{\includegraphics[height=0.6\linewidth]{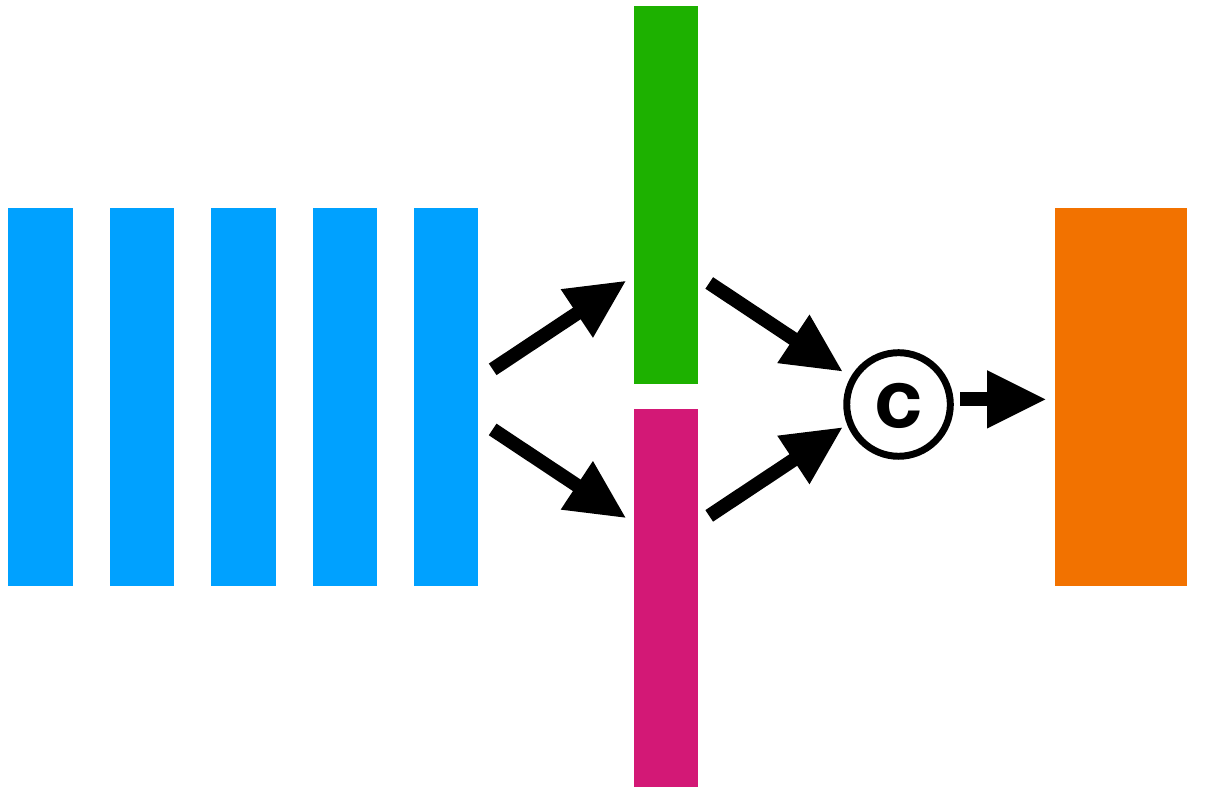}} \hspace{0.1em}
\subcaptionbox{No context network\label{fig:decoder_no_context}}[0.3\linewidth]{\includegraphics[height=0.8\linewidth]{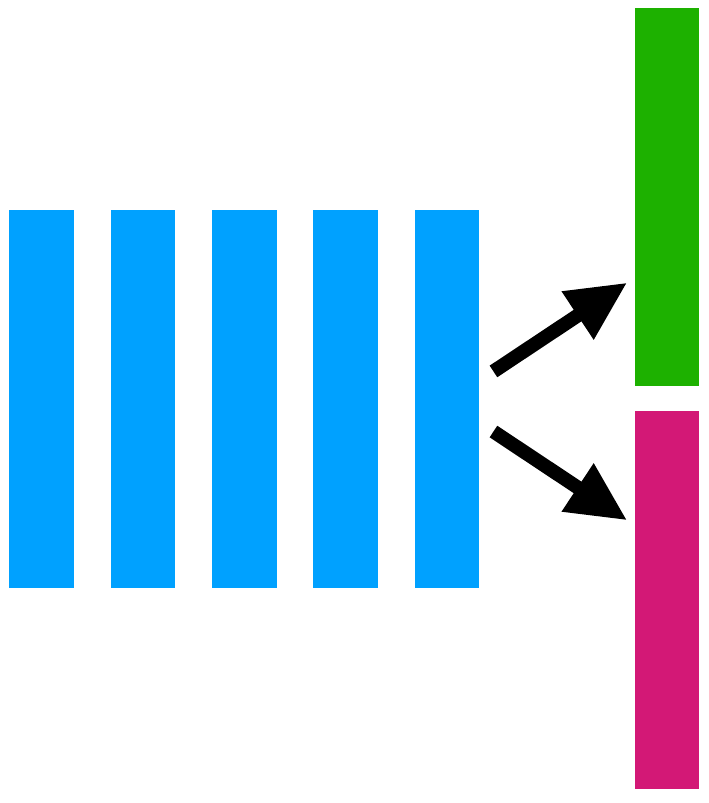}} \hspace{0.1em}
\subcaptionbox{Our split decoder\label{fig:decoder_split}}[0.25\linewidth]{\includegraphics[height=0.96\linewidth]{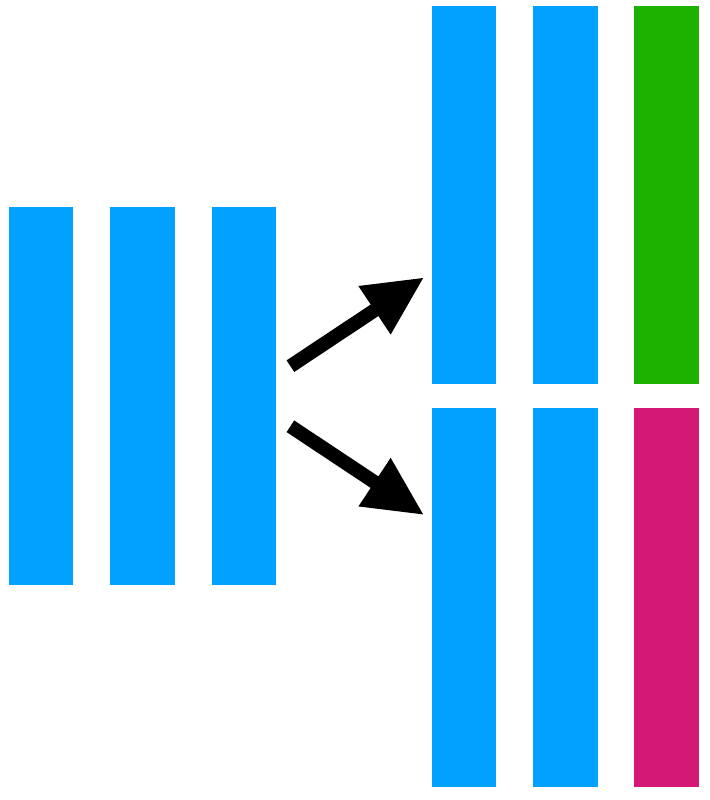}} \\ [0.5em]
\subcaptionbox*{\label{fig:decoder_legend}}{\includegraphics[width=0.95\linewidth]{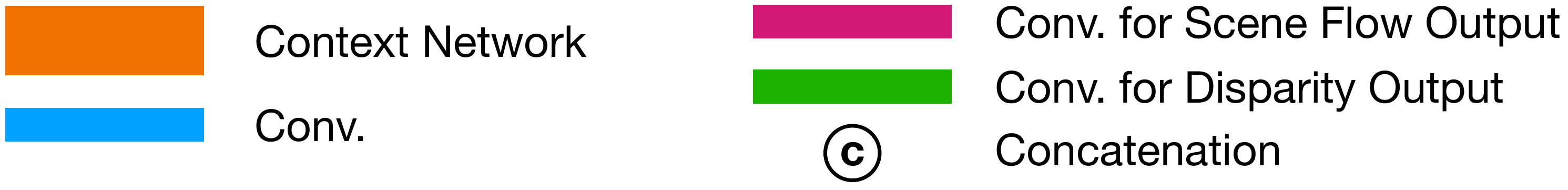}}
\vspace{-1em}
\caption{\textbf{Decoder configuration}: \emph{(a)} A single joint decoder \cite{Hur:2020:SSM}, \emph{(b)} removing the context network, and \emph{(c)} our split decoder design.}
\label{fig:decoder}
%\vspace{-0.25em}
\end{figure}

\myparagraph{Split-decoder design.}
We further probe the decoder design in detail and introduce a split-decoder model that converges faster and more stably.
Hur~\etal~\cite{Hur:2020:SSM} propose to use a single decoder (\cf~\cref{fig:decoder_joint}) that jointly predicts both scene flow and disparity based on the observation that separating the decoder for each task leads to balancing issues.
This can result in a trivial prediction for the disparity (\eg, outputting a constant value for all pixels).
However, we observe that this issue mainly stems from the context network, which we discard (see above) due to stability concerns.

After discarding the context network (\cref{fig:decoder_no_context}), we find a better decoder configuration.
We gradually split the decoder starting from the last layer into two separate decoders for each task and compare the scene flow accuracy in the experimental setting of \cite{Hur:2020:SSM}.
\Cref{table:spliting_decoders} reports the result (lower is better).
Discarding the context network degrades the accuracy by \num{4.1}\%, but in the end, splitting the decoder at the \nth{2}-to-last layer yields an accuracy competitive to the one of \cite{Hur:2020:SSM}.
We choose this configuration (\ie~\cref{fig:decoder_split}) as our decoder design.
Our findings suggest that the conclusions of \cite{Hur:2020:SSM} regarding the decoder design only hold in the presence of a context network.
The benefit of our split decoder is that competitive accuracy is achieved more stably and in fewer training iterations (at \num{56}\% of the full training schedule), with a lighter network ($\sim$ 10\% fewer parameters).\textsuperscript{\ref{footnote_supp1}}

\begin{table}[t]
\centering
\footnotesize
\begin{tabularx}{\columnwidth}{@{}XS[table-format=2.2]@{\hskip 0.4em}S[table-format=2.2]@{\hskip 0.4em}S[table-format=2.2]@{\hskip 0.4em}S[table-format=2.2]@{}}
	\toprule
	Configuration & {D$1$-all} & {D$2$-all} & {Fl-all} & {SF-all} \\
	\midrule
	Single joint decoder (\cref{fig:decoder_joint}, \cite{Hur:2020:SSM}) & 31.25 & \Uline{34.86} & \bfseries 23.49 & \bfseries 47.05 \\
	\midrule
	Removing the context network (\cref{fig:decoder_no_context})			& 33.04 & 36.45 & 36.45 & 48.97 \\
	Splitting at the last layer 											& 34.11 & 37.28 & \Uline{24.49} & 49.92 \\
	Splitting at the \nth{2}-to-last layer (\cref{fig:decoder_split}) 		& \bfseries 30.50 & \bfseries 34.72 & 24.72 & \Uline{47.53} \\
	Splitting at the \nth{3}-to-last layer 	& \Uline{31.04} & 35.79 & 24.57 & 47.85 \\
	Splitting at the \nth{4}-to-last layer 	& 32.21 & 36.28 & 24.65 & 48.84 \\
	Splitting into two separate decoders 	& 32.12 & 36.14 & 25.02 & 48.66 \\
	\bottomrule
\end{tabularx}
\caption{\textbf{Scene flow accuracy of split-decoder designs}: Removing the context network degrades the accuracy (\cf~\cref{sec:exp:ablation} for a description of the metrics), but splitting the decoder at the \nth{2}-to-last layer yields competitive accuracy while being stable to train.}
\label{table:spliting_decoders}
\vspace{-0.5em}
\end{table}

\subsection{Multi-frame estimation}
\label{sec:approach:multiframe}

\begin{figure}[t]
\centering
\includegraphics[width=\linewidth]{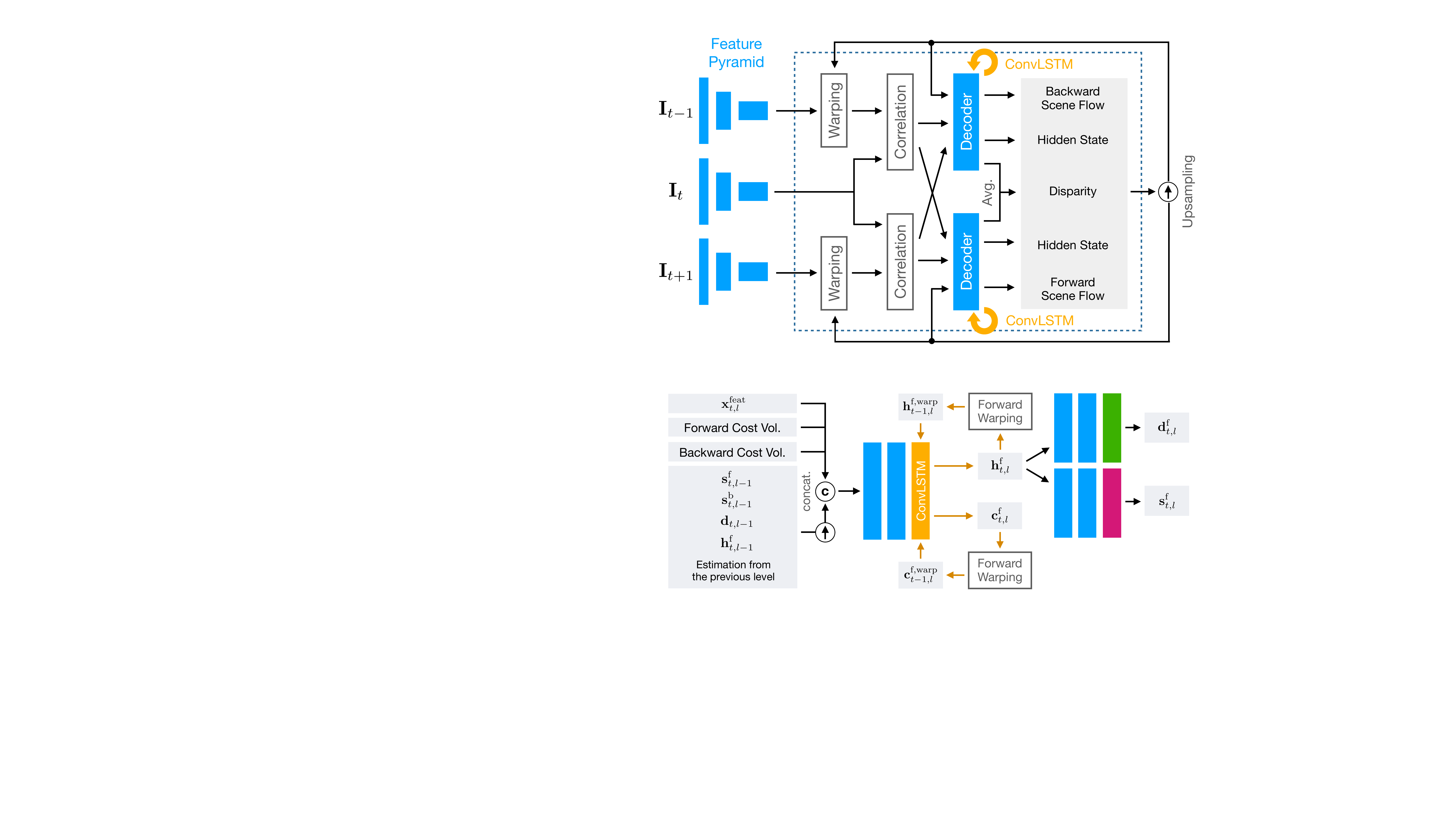}
% \subcaptionbox{A detailed view of the decoder at pyramid level $l$\label{fig:architecture_decoder}}{\includegraphics[width=0.94\linewidth]{figures/architecture_decoder.pdf}}
\caption{\textbf{Our network architecture for multi-frame monocular scene flow based on PWC-Net} \cite{Sun:2018:PWC}: At each pyramid level (\emph{dashed square}), the two cost volumes are input to the decoder (shared for bi-directional estimation) to estimate the residual scene flow and disparity (averaged from the two decoders).}
\label{fig:architecture_overview}
\vspace{-0.5em}
\end{figure}

\paragraph{Three-frame estimation.}
Toward temporally consistent estimation over multiple frames, we first utilize three frames at each time step \cite{Janai:2018:ULM,Liu:2019:SFS}.
\cref{fig:architecture_overview} illustrates our network for multi-frame estimation in detail.
For simplicity, we only visualize one pyramid level (\ie~the dashed square in \cref{fig:architecture_overview}), noting that we iterate this across all pyramid levels.
Given the feature maps from each frame at times $t-1, t,$ and $t+1$, the forward cost volume (from $t$ to $t+1$) and the backward cost volume (from $t$ to $t-1$) are calculated from the correlation layer and fed into the decoder that estimates the forward scene flow $\mv{s}^\text{f}_t$ and disparity $\mv{d}^\text{f}_t$.
The remaining inputs of the decoders are the feature map $\mv{x}^\text{feat}_{t}$ from the encoder, upsampled estimates, and the hidden states in the convolutional LSTM (ConvLSTM) \cite{Shi:2015:CLSTM} module, see below, both from the previous pyramid level.
For backward scene flow $\mv{s}^\text{b}_t$ with disparity $\mv{d}^\text{b}_t$, the same decoder with shared weights is used by switching the order of the inputs.
We average the two disparity predictions for the final estimate, \ie~$\mv{d}_t =( \mv{d}^\text{f}_t + \mv{d}^\text{b}_t ) / 2$, as they correspond to the same view and should be consistent forward and backward in time.

\myparagraph{LSTM with forward warping.}
To further encourage temporal consistency, we employ a convolutional LSTM \cite{Shi:2015:CLSTM} in the decoder so that it can temporally propagate the hidden state across overlapping frame triplets (\cf~\cref{fig:teaser}) and implicitly exploit the previous estimates for the current time step.
\cref{fig:architecture_decoder} shows our decoder in detail, visualizing only the forward scene flow case $\mv{s}^\text{f}_{t,l}$ at pyramid level $l$ for simplicity.
Inside the decoder, we place the ConvLSTM module right before splitting into two separate decoders so that we can temporally propagate the joint intermediate representation of scene flow and depth.
The ConvLSTM is fed the feature map from previous layers, as well as cell state $\mv{c}^\text{warp}_{t-1, l}$ and hidden state $\mv{h}^\text{warp}_{t-1, l}$, both from the previous time step at the same pyramid level $l$.
The module outputs the current cell state $\mv{c}_{t,l}$ and hidden state $\mv{h}_{t,l}$, which are fed into the subsequent split decoder that outputs residual scene flow and (non-residual) disparity, respectively.
We use a leaky ReLU activation instead of tanh in the ConvLSTM, aiding faster convergence in our case.

Importantly, we forward-warp the previous states (\ie~$\mv{c}_{t-1, l}$ and $\mv{h}_{t-1, l}$) using the estimated scene flow at the previous time step $\mv{s}^f_{t-1}$ so that the coordinates of the states properly correspond.
Without warping, each pixel in the current frame will attend to the previous state from mismatched pixels, which does not ensure proper propagation of corresponding states.
Using backward warping based on backward scene flow at the previous pyramid level $\mv{s}^b_{t, l-1}$ may also be possible, but exhibits a challenge: using backward flow (at $l\!-\!1$) to warp the previous states to update itself (at $l$), which is not easy if the initial estimate is unreliable.
%but exhibits a challenge of warping the corresponding previous states using possibly less accurate backward flow from the previous level to estimate itself.
%but exhibits the chicken-and-egg problem of using backward flow to warp the previous states to estimate the backward flow, which is difficult to overcome.

\begin{figure}[b]
\centering
\vspace{-0.5em}
\subcaptionbox{Previous frame $\mv{I}_{t-1}$\label{fig:fw_source}}{\includegraphics[width=0.32\linewidth]{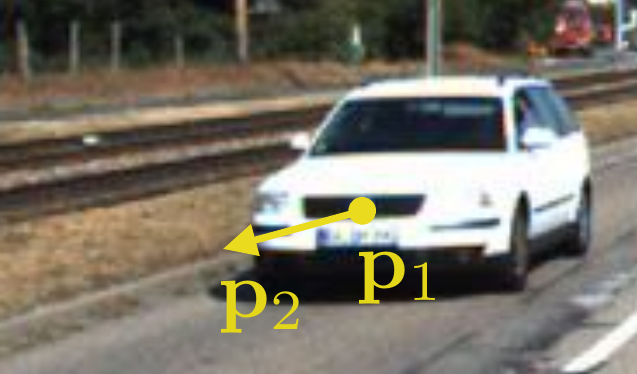}}
\subcaptionbox{Current frame $\mv{I}_{t}$\label{fig:fw_target}}{\includegraphics[width=0.32\linewidth]{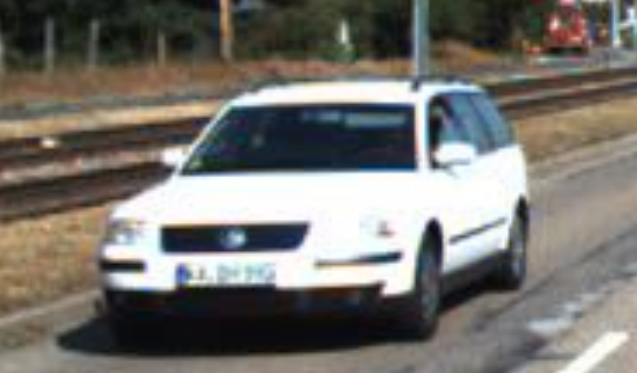}}
\subcaptionbox{Warping result $\mv{I}^\text{warp}_{t-1}$\label{fig:fw_warped}}{\includegraphics[width=0.32\linewidth]{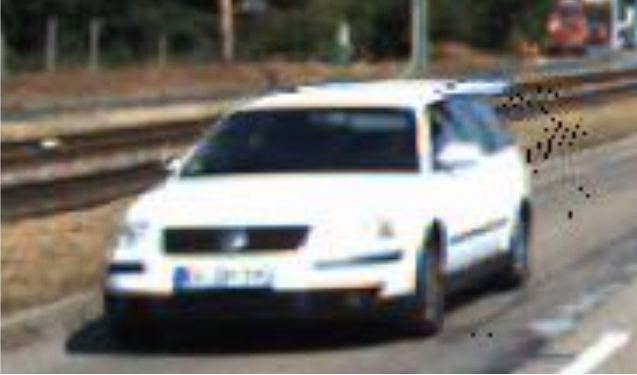}}
\caption{\textbf{An example of forward warping}: \emph{(a)} Previous frame, \emph{(b)} current frame, and \emph{(c)} forward-warped previous frame.}
\label{fig:forward_warping}
\end{figure}

\begin{figure}[t]
\centering
\includegraphics[width=\linewidth]{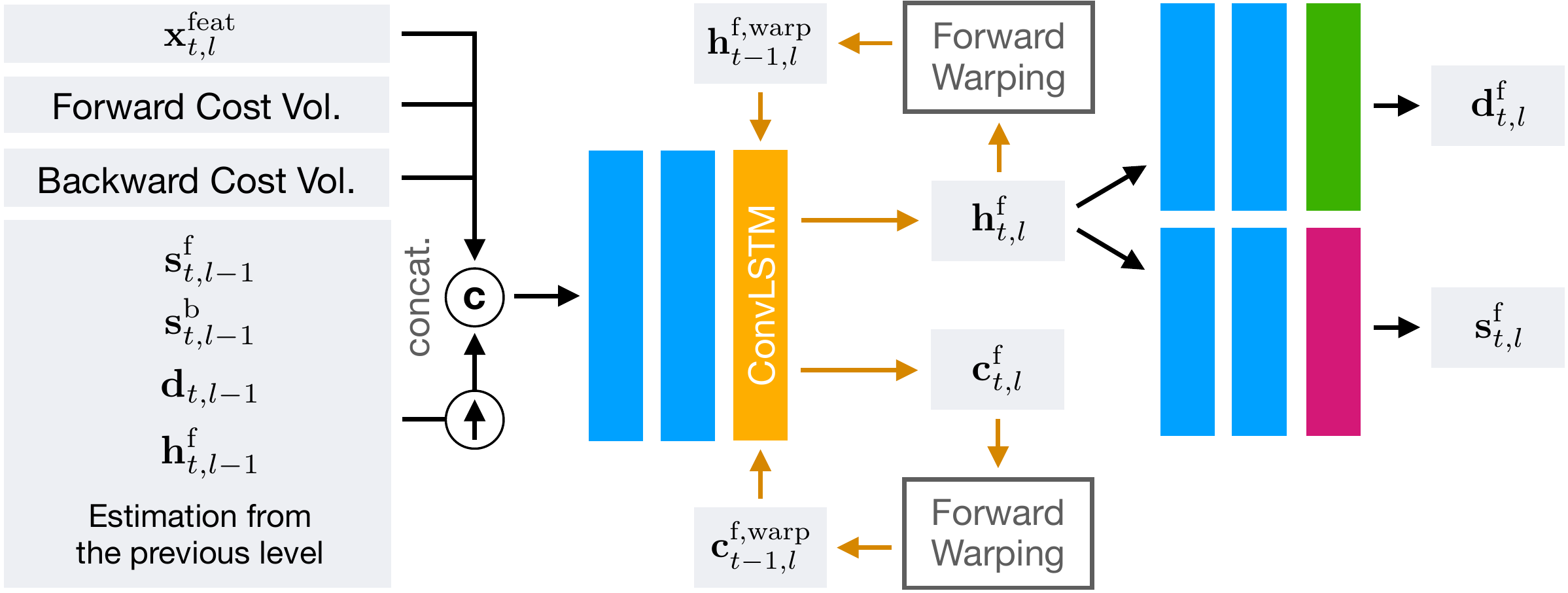}
\caption{\textbf{A detailed view of the decoder at pyramid level $l$}, based on a convolutional LSTM with forward warping.}
\label{fig:architecture_decoder}
\vspace{-0.5em}
\end{figure}

As an example, \cref{fig:forward_warping} shows the forward-warping result ($\mv{I}^\text{warp}_{t-1}$, \cref{fig:fw_warped}) of the previous frame $\mv{I}_{t-1}$ (\cref{fig:fw_source}), which matches the current frame $\mv{I}_{t}$ (\cref{fig:fw_target}) well.
When a pixel $\mv{p}_1$ moves to $\mv{p}_2$ in the next frame, the pixel $\mv{p}_2$ in $\mv{I}_{t}$ should attend to the previous state of the corresponding pixel in $\mv{I}_{t-1}$, \ie~$\mv{h}_{t-1, l}(\mv{p}_1)$.
To do so, we forward-warp the previous states using the estimated scene flow and disparity.
% which can let  = \mv{h}_{t-1, l}(\mv{p_1})$.
Furthermore, we use a validity mask $M$ to filter out states from mismatched pixels based on the affinity score of CNN feature vectors from corresponding pixels:
\begin{subequations}
\begin{eqnarray}
&\mv{h}^\text{warp}_{t-1, l} = f_\text{w}(\mv{h}_{t-1, l}) M(\mv{x}^\text{feat}_{t-1}, \mv{x}^\text{feat}_{t}) \\
&\mv{c}^\text{warp}_{t-1, l} = f_\text{w}(\mv{c}_{t-1, l}) M(\mv{x}^\text{feat}_{t-1}, \mv{x}^\text{feat}_{t})
\end{eqnarray}
with
\begin{equation}
M(\mv{x}^\text{feat}_{t-1}, \mv{x}^\text{feat}_{t}) = (\mathit{conv}_{1\times1}( f_\text{w}(\mv{x}^\text{feat}_{t-1}) \cdot \mv{x}^\text{feat}_{t}) > \text{0.5}),
\end{equation}
\end{subequations}
where $f_\text{w}$ is the forward-warping operation with (implicitly) given estimated scene flow and disparity.
To generate the per-pixel mask $M$, we forward-warp the previous (normalized) feature map ($\mv{x}^\text{feat}_{t-1}$), dot-product with the current (normalized) feature map ($\mv{x}^\text{feat}_{t}$) to calculate the affinity score, pass it through a $1\times1$ convolution layer, and apply a fixed threshold.
Here, the $1\times1$ convolution is used to learn to scale the affinity score before thresholding.

For forward warping, we adopt the softmax splatting strategy of Niklaus~\etal~\cite{Niklaus:2020:SSV}, which resolves conflicts between multiple pixels mapped into the same pixel location when forward-warping.
In our case, we utilize the estimated disparity as a cue to compare the depth orders, determine visible pixels, and preserve their hidden states.

\subsection{Self-supervised loss}
\label{sec:approach:loss}

\begin{figure}[t]
\centering
\includegraphics[width=\linewidth,trim=0 30 0 0,clip]{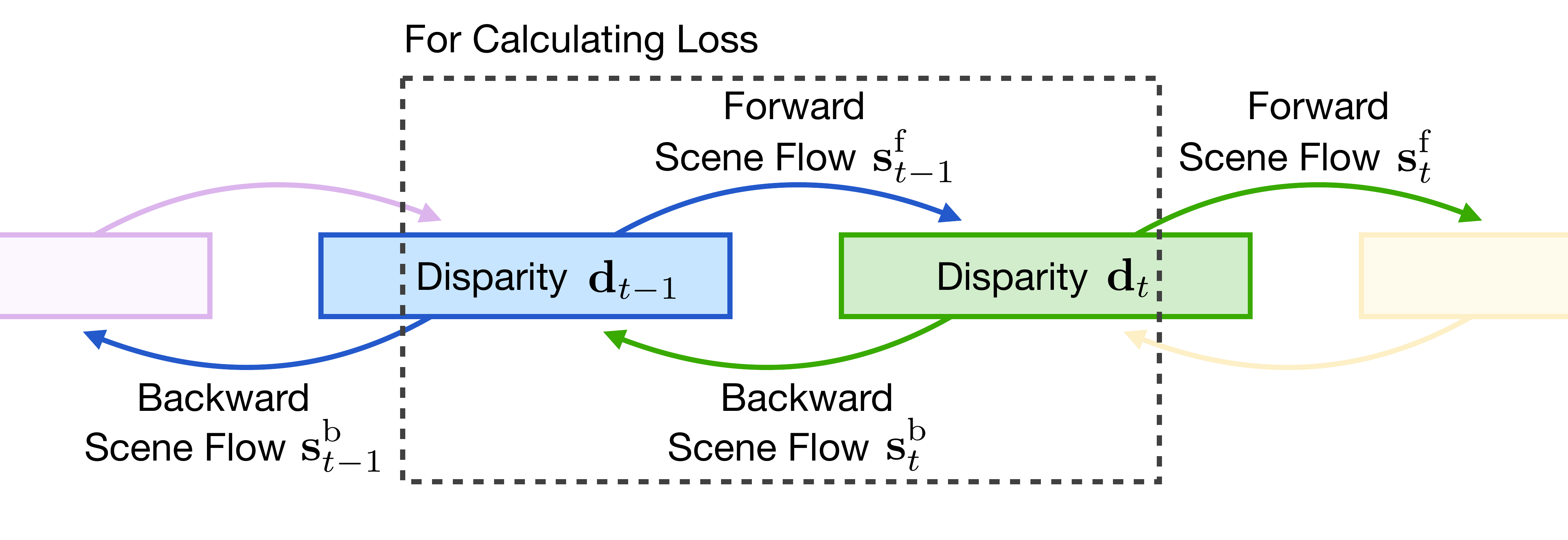}
\caption{\textbf{Loss scheme.} The self-supervised loss is applied between each pair of temporally neighboring estimates.}
\label{fig:loss_for_triplet}
%\vspace{-0.5em}
\end{figure}

Given the scene flow and disparity estimates over the multiple frames, we apply a self-supervised loss on each pair of temporally neighboring estimates that establish a bi-directional relationship.
This allows us to exploit occlusion cues.
As shown in \cref{fig:loss_for_triplet}, given the estimates for two time steps as \{$\mv{s}^\text{f}_{t-1}, \mv{s}^\text{b}_{t-1}, \mv{d}_{t-1}$\} and \{$\mv{s}^\text{f}_t, \mv{s}^\text{b}_t, \mv{d}_t$\}, we apply the proxy loss to \{$\mv{s}^\text{f}_{t-1}, \mv{d}_{t-1}, \mv{s}^\text{b}_t, \mv{d}_t$\}.
We adopt the self-supervised loss from Hur~\etal~\cite{Hur:2020:SSM}, which consists of a view synthesis loss and a 3D reconstruction loss, guiding the disparity and scene flow output to be consistent with the given input images.
The total self-supervised loss is a weighted sum of disparity loss $L_\text{d}$ and scene flow loss $L_\text{sf}$,
\begin{equation}
L_\text{total} = L_\text{d} + \lambda_\text{sf} L_\text{sf}.
\label{eq:total_loss}
\end{equation}
The main difference to \cite{Hur:2020:SSM} is that we newly propose an occlusion-aware census loss for penalizing the photometric difference.
We only introduce our novel contribution here and provide details on the losses from \cite{Hur:2020:SSM} with our modifications in \cref{supp:sec:loss}.

\begin{figure}
\centering
% \begin{minipage}{0.3\linewidth}
	\begin{subfigure}{0.29\linewidth}
		\includegraphics[width=\linewidth]{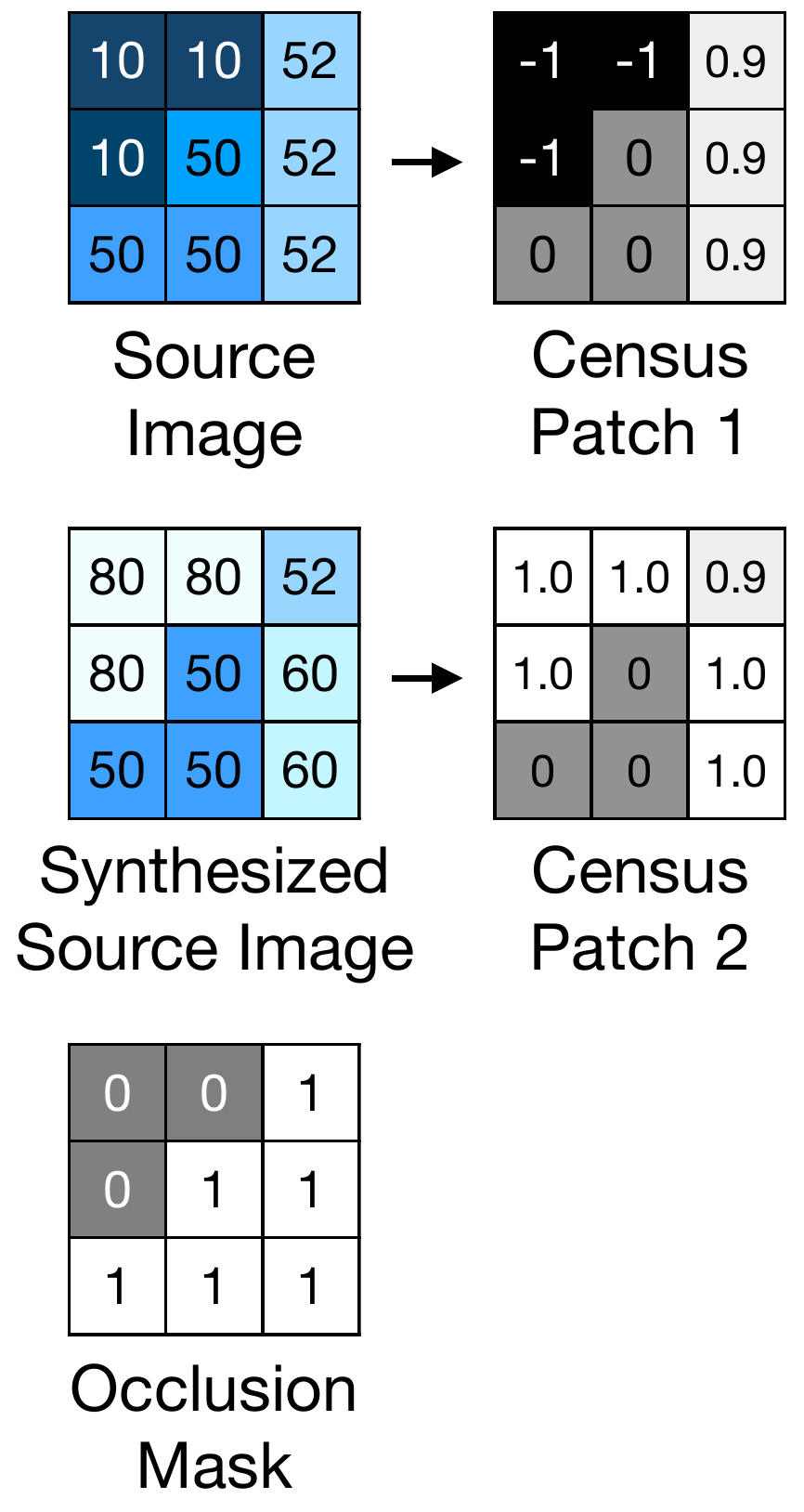}
		\caption{Census transform}
		\label{fig:census_img}
	\end{subfigure}
% \end{minipage}
\hspace{0.4em}
\begin{minipage}{0.67\linewidth}
	\centering
	\begin{subfigure}{0.8\linewidth}
		\includegraphics[width=\linewidth]{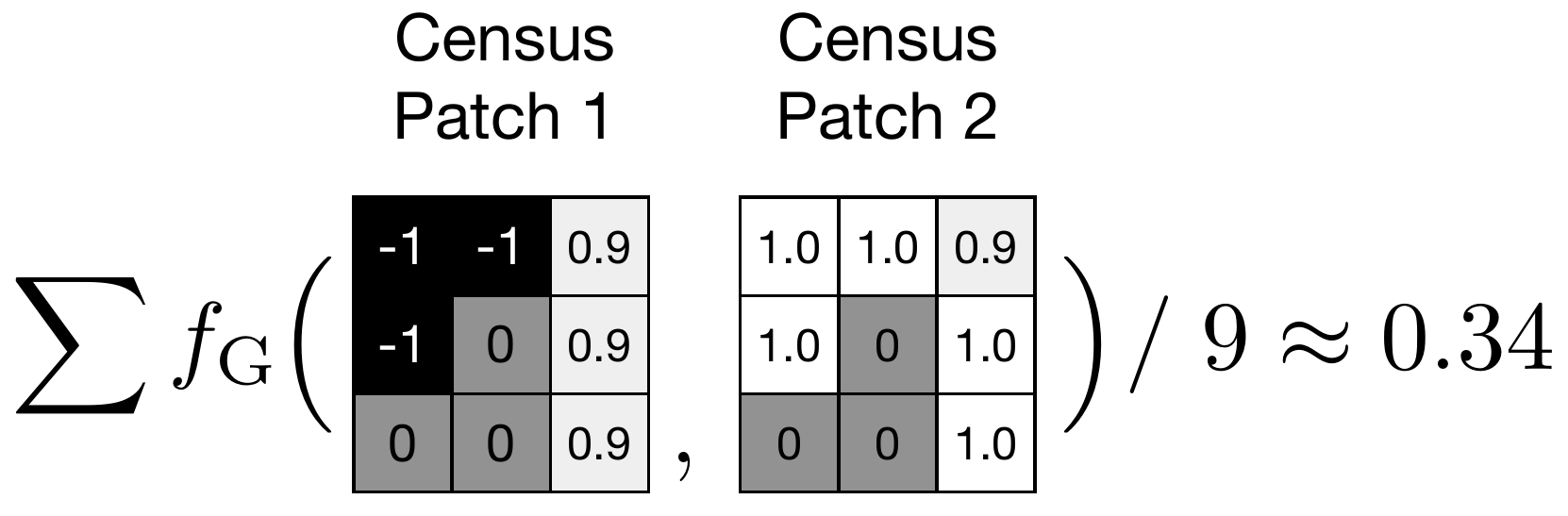}
		\caption{Standard approach}
		\label{fig:census_standard}
		\vspace{0.3em}
	\end{subfigure}
	\begin{subfigure}{\linewidth}
		\includegraphics[width=\linewidth]{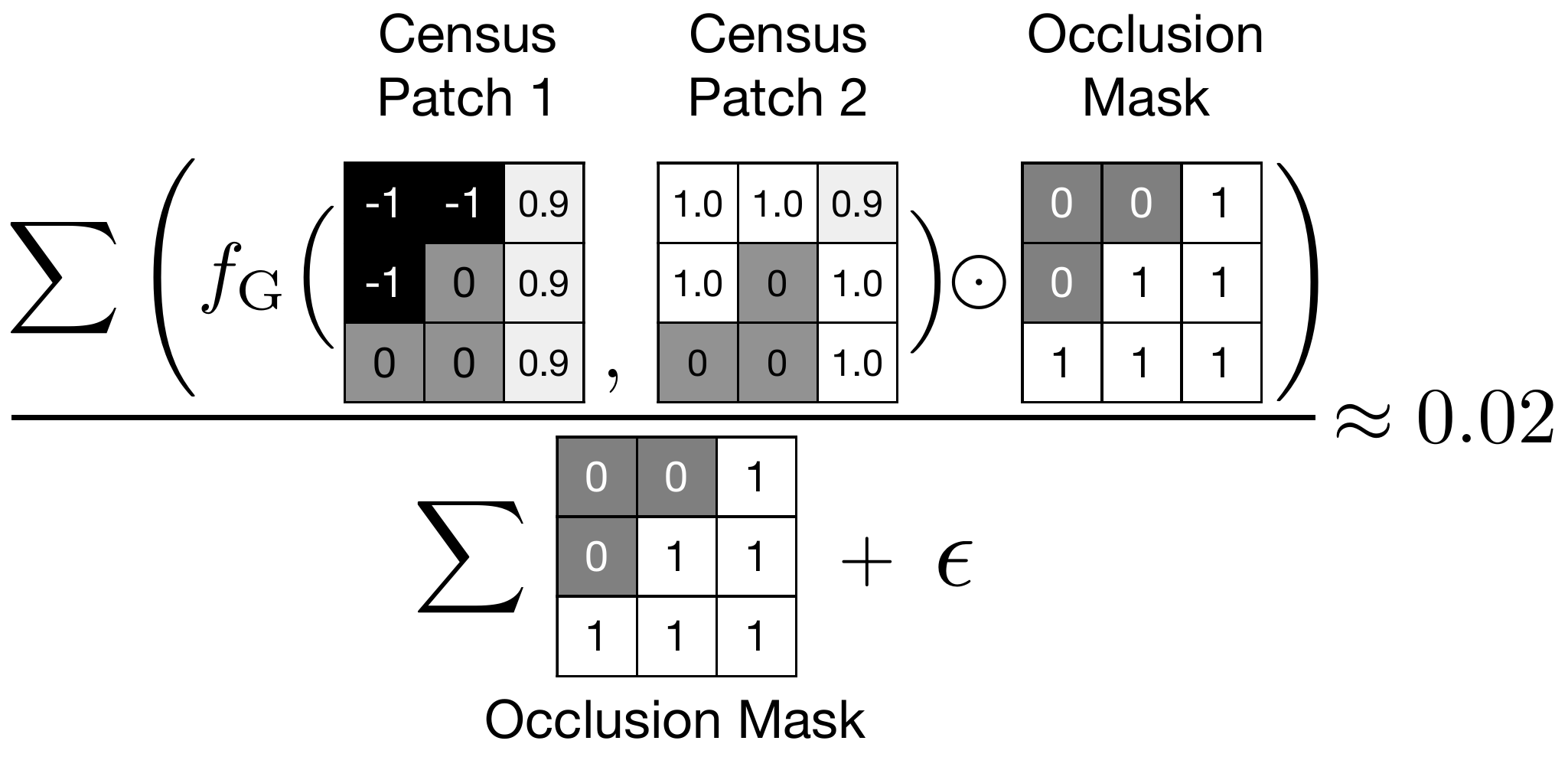}
		\caption{Our occlusion-aware approach}
		\label{fig:census_occ}
	\end{subfigure}
\end{minipage}
\caption{\textbf{Occlusion-aware census transform}: \emph{(a)} Computing the (continuous) census signature of two image patches and the corresponding occlusion mask, \emph{(b)} standard approach of computing the Hamming distance of the census signature divided by the number of pixels, and \emph{(c)} our occlusion-aware Hamming distance.}
\label{fig:occ_census}
\vspace{-0.5em}
\end{figure}

\myparagraph{Occlusion-aware census transform.}
Carefully designing the proxy loss function matters for the accuracy of self-supervised learning \cite{Jonschkowski:2020:WMU}.
For penalizing the photometric difference for the view-synthesis proxy task, the census transform \cite{Stein:2004:ECO,Zabih:1994:NPL} has demonstrated its robustness to illumination changes, \eg,~in outdoor scenes \cite{Jonschkowski:2020:WMU,Liu:2019:DLO,Meister:2018:ULO,Vogel:2013:EDC}.
The conventional (ternary) census transform computes the local census patch (\cref{fig:census_img}) and calculates the Hamming distance between them to evaluate the brightness difference (\cref{fig:census_standard}).
However, it is vulnerable to outlier pixels (\eg, occlusions) present in the patch, yielding a higher distance.

Taking into account the occlusion cue, we introduce an occlusion-aware census transform between images $\mv{I}$ and $\mv{\tilde{I}}$, which calculates the Hamming distance only for visible pixels:
\begin{subequations}
\begin{align}
&\rho_\text{census}\big(\mv{I}, \mv{\tilde{I}}, \mv{O}\big) = \frac{1}{\sum_\mv{p} \mv{O}(\mv{p}) } \cdot \label{eq:occ_census} \\\nonumber
& \sum_{\mv{p}} \frac{ \sum_{\mv{y} \in [-3, 3]^2} f_\text{G} \Big( T(\mv{I}, \mv{p}, \mv{y}), T(\mv{\tilde{I}}, \mv{p}, \mv{y}) \Big) \mv{O}(\mv{p} + \mv{y}) }{ \sum_{\mv{y} \in [-3, 3]^2} \mv{O}(\mv{p} + \mv{y}) + \epsilon }
\end{align}
with occlusion state $\mv{O}$ (with $\mv{O}(\mv{p})=1$ if visible) and
\begin{align}
T(\mv{I}, \mv{p}, \mv{y}) &= \frac{  \mv{I}(\mv{p} + \mv{y}) - \mv{I}(\mv{p})  }{  \sqrt{(\mv{I}(\mv{p} + \mv{y}) - \mv{I}(\mv{p}))^2 + \sigma_\text{T}^2}  }
\label{eq:ternary_value}\\
f_\text{G}(t_1, t_2) &= \frac{(t_1 - t_2)^2}{(t_1 - t_2)^2 + \sigma_\text{G}},
\label{eq:ternary_GM_ftn}
\end{align}
\end{subequations}
where $\sigma_\text{G} = 0.1$ and $\sigma_\text{T} = 0.9$.
To facilitate differentiability, $T(\mv{I}, \mv{p}, \mv{y})$ in \cref{eq:ternary_value} calculates a continuous approximation to the ternary value at pixel $\mv{p}$ with an offset $\mv{y}$ in image \mv{I}.
$f_G$ in \cref{eq:ternary_GM_ftn} is the Geman-McClure function \cite{Black:1996:ULP} that scores the difference of the two input ternary values.

As shown in \cref{fig:census_occ}, our occlusion-aware formulation can prevent from having a high score caused by occlusions in the census patch, thus providing a measure for the brightness difference that is more robust against outliers.

\subsection{Improving the training stability}
\label{sec:approach:stablility}

\begin{figure}[t]
\centering
\subcaptionbox{Detaching the gradient between scene flow loss and disparity decoder\label{fig:grad_detach_img}}{\includegraphics[width=0.9\linewidth]{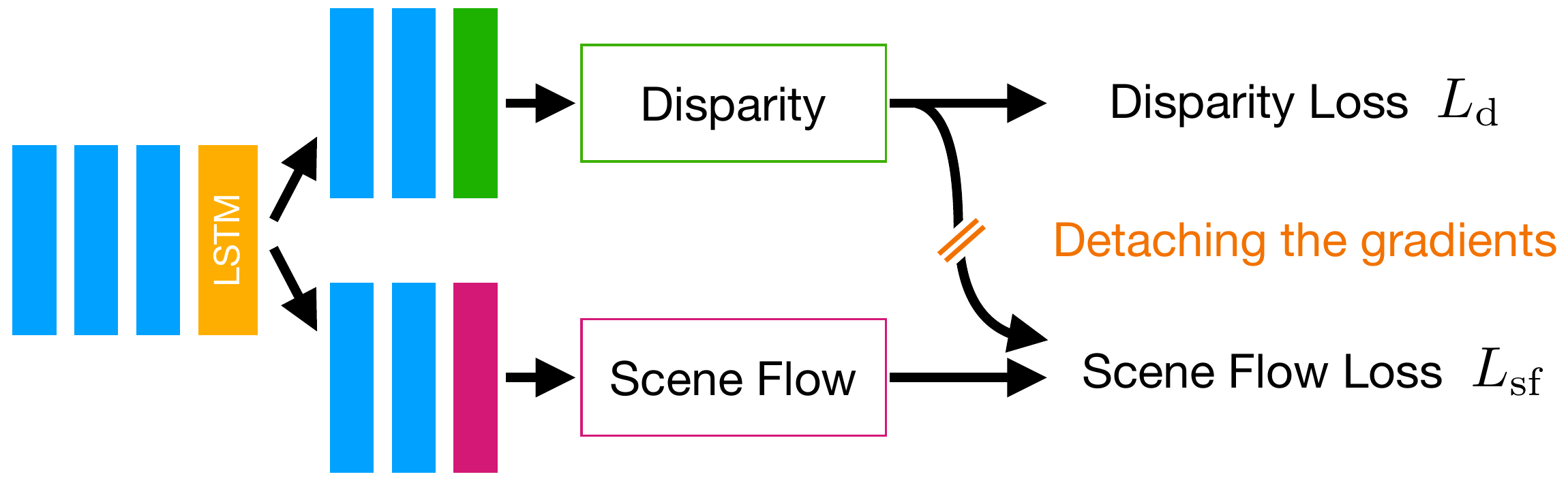}} \\ [0.5em]
\subcaptionbox{Training loss\label{fig:grad_detach_loss}}{\includegraphics[width=0.49\linewidth]{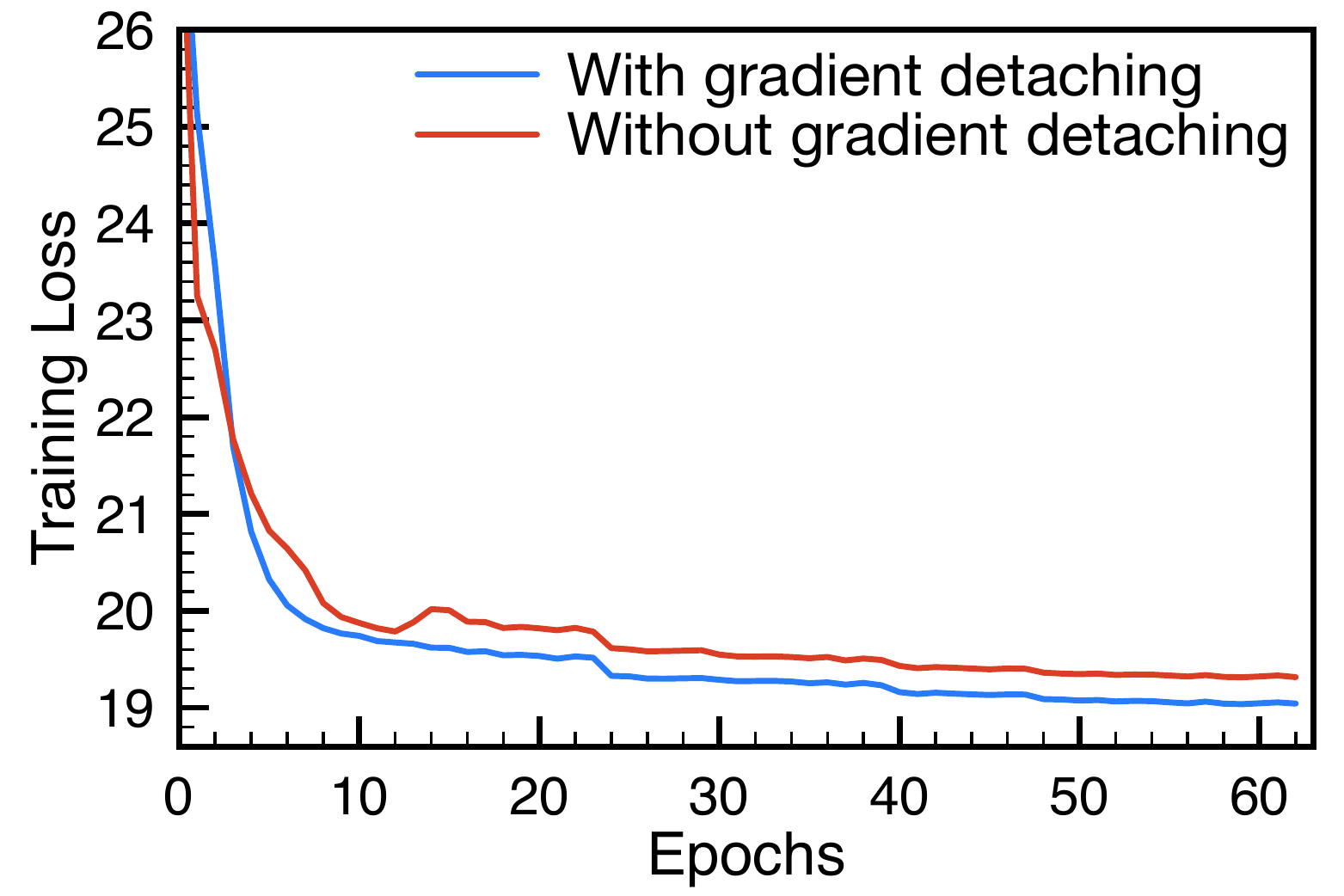}}
\subcaptionbox{Scene flow outlier rate\label{fig:grad_detach_error}}{\includegraphics[width=0.49\linewidth]{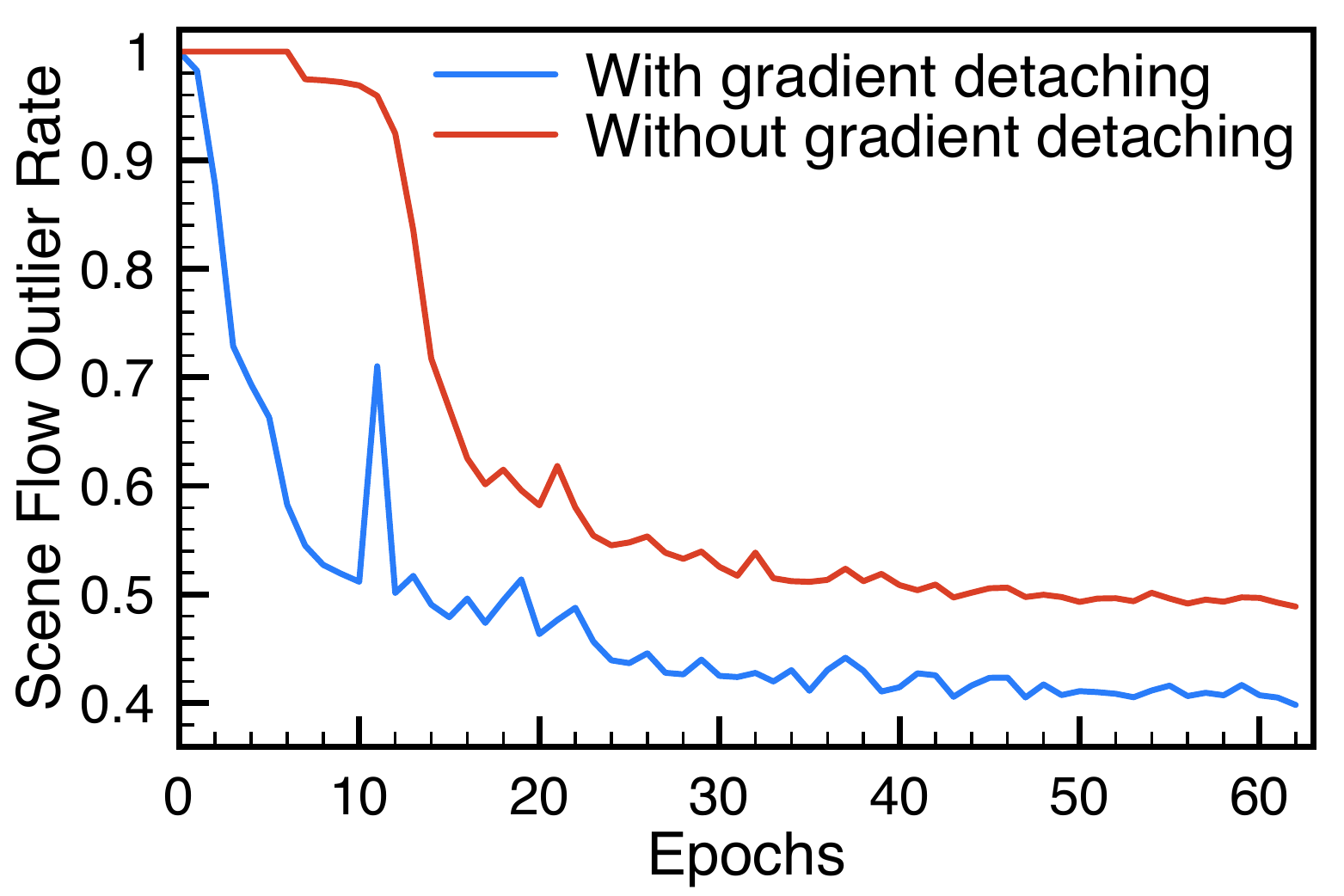}}
\caption{\textbf{Our gradient detaching strategy}: \emph{(a)} Detaching the gradient between the scene flow loss and the disparity decoder in the early stages of the training improves \emph{(b)} the training loss convergence and \emph{(c)} the scene flow accuracy significantly.}
\label{fig:grad_detach}
\vspace{-0.5em}
\end{figure}

As shortly discussed in \cref{sec:approach:architecture}, discarding the context network improves the training stability.
However, we found that integrating a ConvLSTM module \cite{Shi:2015:CLSTM} may still yield unstable training, resulting in trivial disparity predictions in the early stages of training.

To resolve the issue, we propose to detach the gradient from the scene flow loss ($L_\text{sf}$ in \cref{eq:total_loss}) to the disparity decoder in the early stages of training so that each split decoder focuses on its own task first.
We conjecture that the gradient back-propagated from the scene flow loss to the disparity decoder strongly affects the disparity estimate, yielding a trivial prediction in the end.
To prevent the scene flow from dominating, we detach the gradients, but only for the first 2 epochs of the training schedule as illustrated in \cref{fig:grad_detach_img}, and then continue to train in the normal setting.

\Cref{fig:grad_detach_loss,fig:grad_detach_error} demonstrate the effect of detaching the gradient in terms of the training loss and the scene flow outlier rate.
Without detachment, the model outputs a constant disparity map in the early stage of training, thus yields higher scene flow error rates.
In contrast, applying our gradient detaching strategy demonstrates faster and stable convergence, with much better accuracy (39.82\% \vs 49.69\%).

% !TEX root = ../arxiv.tex
\section{Experiments}
\label{sec:experiments}

\subsection{Implementation details}
\label{subsec:implementation_details}
For a fair comparison with the most closely related prior work \cite{Hur:2020:SSM}, we use the same dataset (\ie~KITTI raw \cite{Geiger:2013:VMR}) and the same training protocol, assuming a fixed stereo baseline.
We use the \emph{KITTI Split} \cite{Godard:2017:UMD} by splitting the \num{32} scenes total into \num{25} scenes for training and the remaining \num{7} for validation.
Unlike \cite{Hur:2020:SSM}, we divide the training/validation split at the level of entire scenes in order to exploit more continuous frames for our multi-frame setup and completely remove possible overlaps between the two splits.
Then, we evaluate our model on \emph{KITTI Scene Flow Training} \cite{Menze:2015:J3E,Menze:2018:OSF}, using the provided scene flow ground truth.
Note that \emph{KITTI Split} and \emph{KITTI Scene Flow Training} do not overlap.
After our self-supervised training on \emph{KITTI Split}, we optionally fine-tune our model on \emph{KITTI Scene Flow Training} in a semi-supervised manner and compare with previous state-of-the-art monocular scene flow methods \cite{Brickwedde:2019:MSF,Yang:2020:UOF}.

Given that we use the network of \cite{Hur:2020:SSM} as the basis, we use the same augmentation schemes and training configurations (\eg, learning rate, training schedule, optimizer, \etc), except for the following changes.
For training, we use one sequence of \num{5} temporally consecutive frames as a mini-batch.
To ensure training stability, we detach the gradient between the scene flow loss and the disparity decoder during the first \num{2} epochs, as discussed in \cref{sec:approach:stablility}.\footnote{
Code is available at \href{https://github.com/visinf/multi-mono-sf}{\nolinkurl{github.com/visinf/multi-mono-sf}}.
}\footnote{\label{footnote_supp2}See supplementary material for more details and analyses.}
%More details are given in the supplemental material.

\subsection{Ablation study}
\label{sec:exp:ablation}
We conduct a series of ablations to study the accuracy gain from our contributions over the two-frame baseline.
We use \emph{our} multi-frame train split of \emph{KITTI Split} and evaluate on \emph{KITTI Scene Flow Training} \cite{Menze:2015:J3E,Menze:2018:OSF} using the scene flow evaluation metric.
The metric reports the outlier rate (in \%, lower is better) among pixels with ground truth; a pixel is regarded an outlier if exceeding a threshold of 3 pixels or 5\% w.r.t.~the ground-truth disparity or motion.
After evaluating the outlier rate of the disparity (D1-all), disparity change (D2-all), and optical flow (Fl-all), the scene flow outlier rate (SF-all) is obtained by checking if a pixel is an outlier on either of them.

\myparagraph{Advanced two-frame baseline.}
In \cref{table:ablation_baseline}, we first conduct an ablation study of our advanced baseline described in \cref{sec:approach:architecture}.
We first train the original implementation of Hur~\etal~\cite{Hur:2020:SSM} on \emph{our} train split.
Interestingly, the accuracy significantly drops by \num{29.4}\% (relative change) compared to training on their data split.
%Note that is is not a reproducibility issue as we can reproduce their numbers on their split,\footnote{The split of \cite{Hur:2020:SSM} is unsuitable for multi-frame methods.} but from using our train split with less diverse scenes.
This difference comes down to our train split containing less diverse scenes, which suggests that \cite{Hur:2020:SSM} may be somewhat sensitive to the choice of training data.
Simply removing the context network already significantly improves the accuracy up to \num{17.1}\% by improving the training stability, almost closing the gap.
As discussed in \cref{sec:approach:architecture}, we follow \cite{Jonschkowski:2020:WMU} and use one less pyramid level and feature normalization, which further improves the accuracy, even slightly beyond that of the original baseline trained on the split of \cite{Hur:2020:SSM} (covering more diverse scenes).

\begin{table}[t]
\centering
\footnotesize
\setlength\tabcolsep{3pt}
% \rotatebox[origin=c]{90}{Occ}
\begin{tabular*}{\columnwidth}{@{\extracolsep{\fill}}l@{\hskip 1em}S[table-format=2.2]S[table-format=2.2]S[table-format=2.2]S[table-format=2.2]@{}}
	\toprule
	Baseline type & {D$1$-all} & {D$2$-all} & {Fl-all} & {SF-all} \\
	\midrule
	Hur~\etal~\cite{Hur:2020:SSM} (on original split) & \bfseries 31.25 & 34.86 & 23.49 & 47.05 \\
	\midrule
	Hur~\etal~\cite{Hur:2020:SSM} (on our train split) & 36.70 & 45.50 & 24.65 & 60.88 \\
	\cite{Hur:2020:SSM} $-$ Context Net & 34.24 & 37.32 & 25.06 & 50.49 \\
	\bfseries \cite{Hur:2020:SSM} $-$ Context Net $+$ \cite{Jonschkowski:2020:WMU} & 33.56 & \bfseries 34.49 & \bfseries 22.92 & \bfseries 46.70 \\
	\bottomrule
\end{tabular*}
\caption{The accuracy of the original baseline of \cite{Hur:2020:SSM} on our train split \emph{(second row)} is significantly lower than on the original split \emph{(first row)}. Removing the context network \emph{(third row)} and exploiting the findings from \cite{Jonschkowski:2020:WMU} \emph(\emph{fourth row}, \textbf{our advanced two-frame baseline}\emph) improves both the accuracy and the training stability.}
\label{table:ablation_baseline}
\end{table}

\begin{table}[t]
\centering
\footnotesize
\setlength\tabcolsep{3pt}
% \rotatebox[origin=c]{90}{Occ}
\begin{tabular*}{\columnwidth}{@{\extracolsep{\fill}}cc@{\hskip 4em}S[table-format=2.2]S[table-format=2.2]S[table-format=2.2]S[table-format=2.2]@{}}
	\toprule
	\makecell{Multi-frame \\ extension} & \makecell{Occ-aware \\ census loss}& {D$1$-all} & {D$2$-all} & {Fl-all} & {SF-all} \\
	\midrule
	\multicolumn{2}{l}{\emph{(Our advanced baseline)}} & 33.56 & 34.49 & 22.92 & 46.70 \\
	\cmark &        					& 28.68 & 32.58 & 21.11 & 42.37 \\
	       & \cmark 					& 30.35 & 31.92 & 21.86 & 43.87 \\
	\bfseries \cmark & \bfseries \cmark & \bfseries 27.33 & \bfseries 30.44 & \bfseries 18.92 & \bfseries 39.82 \\
	\bottomrule
\end{tabular*}
\caption{\textbf{Ablation study of our main contributions}: Our key contributions (multi-frame extension, occlusion-aware census loss) consistently improve the accuracy over our advanced baseline.}
\label{table:ablation_overall}
\vspace{-0.5em}
\end{table}

\myparagraph{Major contributions.}
In \cref{table:ablation_overall}, we validate our major contributions, \ie~multi-frame estimation (\cref{sec:approach:multiframe}) and the occlusion-aware census transform (\cref{sec:approach:loss}), compared to our advanced two-frame baseline.
Both contributions yield a relative accuracy improvement of \num{9.3}\% (multi-frame extension) and \num{6.1}\% (occlusion-aware census transform) over the baseline.
Overall, our final model achieves \num{14.7}\% more accurate results than our baseline.
Note that our advanced baseline already significantly outperforms the original implementation of \cite{Hur:2020:SSM} on our train split by \num{23.3}\%.

\myparagraph{Multi-frame extension.}
We further ablate the technical details of our multi-frame estimation in \cref{table:ablation_multi-frame}.
Inputting bidirectional cost volumes to the decoder (\ie~three-frame estimation) only marginally improves the accuracy.
Adding a ConvLSTM that temporally propagates the hidden states (without using warping) only shows limited improvement as well.
Using backward warping in the LSTM for temporal propagation with backward flow, interestingly, degrades the accuracy, possibly due to using less accurate backward flow from the previous level to warp the hidden states.
%, which possibly mismatched, and use it for the current level.
%possibly due to the network having difficulty deciding which previous hidden states to attend to before accurately estimating the backward flow.

However, when propagating the hidden states with forward warping with the estimated scene flow from the previous frame, we observe a significant relative accuracy improvement of up to \num{9.3}\% compared to two-frame estimation.
Notably, this is the only setting in which there is a clear gain from going to more than two frames.
This highlights the importance of propagating the hidden states and choosing a suitable warping method inside the ConvLSTM.

\begin{table}[t]
\centering
\footnotesize
\setlength\tabcolsep{3pt}
% \rotatebox[origin=c]{90}{Occ}
\begin{tabular*}{\columnwidth}{@{\extracolsep{\fill}}ccc@{\hskip 3em}S[table-format=2.2]S[table-format=2.2]S[table-format=2.2]S[table-format=2.2]@{}}
	\toprule
	Frames & LSTM & Warping & {D$1$-all} & {D$2$-all} & {Fl-all} & {SF-all} \\
	\midrule
	2 & -- & --  & 33.56 & 34.49 & 22.92 & 46.70 \\
	3 & -- & --  & 32.87 & 34.70 & 22.75 & 46.15 \\
	3 & \cmark & --  & \bfseries 28.43 & 33.75 & 25.22 & 45.43 \\
	3 & \cmark & Backward & 30.41 & 35.63 & 25.79 & 46.90 \\
	\bfseries 3 & \bfseries \cmark & \bfseries Forward & 28.68 & \bfseries 32.58 & \bfseries 21.11 & \bfseries 42.37 \\
	\bottomrule
\end{tabular*}
\caption{\textbf{Ablation study of our multi-frame estimation} (\cref{sec:approach:multiframe}): Using a ConvLSTM with forward-warping the hidden states improves the accuracy up to \num{9.3}\% (relative improvement).}
\label{table:ablation_multi-frame}
\end{table}

\begin{table}[t]
\centering
\footnotesize
\setlength\tabcolsep{2pt}
\begin{tabular*}{\columnwidth}{@{\extracolsep{\fill}}l@{\hskip 1.6em}S[table-format=2.2]S[table-format=2.2]S[table-format=2.2]S[table-format=2.2]@{}}
	\toprule
	Loss type & {D$1$-all} & {D$2$-all} & {Fl-all} & {SF-all} \\
	\midrule
	Brightness difference + SSIM				& 28.68 & 32.58 & 21.11 & 42.37 \\
	Standard census \cite{Jonschkowski:2020:WMU}& 28.89 & 31.69 & 20.55 & 41.71 \\
	\bfseries Occlusion-aware census (ours) 	& \bfseries 27.33 & \bfseries 30.44 & \bfseries 18.92 & \bfseries 39.82 \\
	\bottomrule
\end{tabular*}
\caption{\textbf{Ablation study of the occlusion-aware census transform}: Our occlusion-aware census loss provides a more discriminative proxy by taking occlusion cues into account.}
\label{table:ablation_occ_census}
\vspace{-0.5em}
\end{table}

\myparagraph{Occlusion-aware census transform.}
Lastly in \cref{table:ablation_occ_census}, we compare our occlusion-aware census transform to the standard census transform and the widely used basic photometric loss consisting of brightness difference and SSIM \cite{Godard:2019:DIS,Godard:2017:UMD}.
We conduct this experiment on top of our multi-frame architecture.
Our occlusion-aware census further improves the scene flow accuracy by \num{6.0}\% (relative improvement) over the basic photometric loss and by \num{4.5}\% over the standard census transform \cite{Jonschkowski:2020:WMU}.

\subsection{Monocular scene flow}
\cref{table:eval_monosf_kitti_train} compares our method with state-of-the-art monocular scene flow methods on \emph{KITTI Scene Flow Training}.
Our multi-frame architecture achieves the best scene flow accuracy among monocular methods \cite{Hur:2020:SSM,Luo:2019:EPC,Yang:2018:EPC} based on purely self-supervised learning.
Our contributions yield a relative improvement over the direct baseline of \cite{Hur:2020:SSM} of \num{15.4}\%.
Also note that our method closes a substantial part of the gap to the semi-supervised method of \cite{Brickwedde:2019:MSF}, but is significantly faster.
Despite using multiple frames and having better accuracy, the runtime of our approach per time step is actually notably shorter (0.063\si{\s}) than the direct baseline \cite{Hur:2020:SSM} (0.09\si{\s}), benefiting from removing the context network and using one less pyramid level.

\begin{table}[t]
\centering
\footnotesize
\begin{tabularx}{\columnwidth}{@{}XS[table-format=2.2]@{\hskip 0.7em}S[table-format=2.2]@{\hskip 0.7em}S[table-format=2.2]@{\hskip 0.6em}S[table-format=2.2]@{\hskip 0.9em}S[table-format=2.3,table-space-text-post = \si{\s}]@{}}
	\toprule
	Method    &   {D$1$-all}   &  {D$2$-all}  & {Fl-all}   & {SF-all} & {Runtime} \\
	\midrule
	EPC \cite{Yang:2018:EPC}   & 26.81 & 60.97 & 25.74 & {(\gt 60.97)} & 0.05\si{\s}\\
	EPC++ \cite{Luo:2019:EPC}  & \bfseries 23.84 & 60.32 & 19.64 & {(\gt 60.32)} & 0.05\si{\s} \\
	Self-Mono-SF \cite{Hur:2020:SSM}   & 31.25 & 34.86 & 23.49 & 47.05 & 0.09\si{\s}\\
	\bfseries Multi-Mono-SF (ours) 	   & 27.33 & \bfseries 30.44 & \bfseries 18.92 & \bfseries 39.82 & 0.063\si{\s}\\
	\midrule
	Mono-SF \cite{Brickwedde:2019:MSF}$^{\S}$ & 16.72 & 18.97 & 11.85 & 21.60 & 41\si{\s}\\
	% Mono expansion \cite{Yang:2020:UOF} & & & & & 0.25\si{\s} \\
	% Self-Mono-SF-ft \cite{Hur:2020:SSM} & (2.89) & (3.91) & (6.19) & (7.53) & 0.09\si{\s}\\
	\bottomrule
\end{tabularx}
\caption{\textbf{Evaluation on KITTI 2015 Scene Flow Training} \cite{Menze:2015:J3E,Menze:2018:OSF}: Our method achieves state-of-the-art accuracy among un-/self-supervised methods. $^{\S}$semi-supervised method.}
\label{table:eval_monosf_kitti_train}
\end{table}

\begin{table}[t]
\centering
\footnotesize
\begin{tabularx}{\columnwidth}{@{}XS[table-format=2.2]@{\hskip 0.6em}S[table-format=2.2]@{\hskip 0.6em}S[table-format=2.2]@{\hskip 0.6em}S[table-format=2.2]@{\hskip 0.9em}S[table-format=2.3,table-space-text-post = \si{\s}]@{}}
	\toprule
	Method & {D$1$-all} & {D$2$-all} & {Fl-all} & {SF-all} & {Runtime} \\
	\midrule
	Mono-SF \cite{Brickwedde:2019:MSF} 	& \bfseries 16.32 & \bfseries 19.59 & 12.77 & \bfseries 23.08 & 41\si{\s} \\
	Mono expansion \cite{Yang:2020:UOF} & 25.36 & 28.34 & \bfseries 6.30 & 30.96 & 0.25\si{\s} \\
	Self-Mono-SF-ft \cite{Hur:2020:SSM} & 22.16 & 25.24 & 15.91 & 33.88 & 0.09\si{\s}\\
	\bfseries Multi-Mono-SF-ft (ours) & 22.71 & 26.51 & 13.37 & 33.09 & \bfseries 0.063\si{\s}\\
	\midrule
	Self-Mono-SF \cite{Hur:2020:SSM}   	& 34.02 & 36.34 & 23.54 & 49.54 & 0.09\si{\s}\\
	\bfseries Multi-Mono-SF (ours)  & \bfseries 30.78 & \bfseries 34.41 & \bfseries 19.54 & \bfseries 44.04 & \bfseries 0.063\si{\s}\\
	\bottomrule
\end{tabularx}
\caption{\textbf{KITTI 2015 Scene Flow Test} \cite{Menze:2015:J3E,Menze:2018:OSF}: Our method consistently outperforms \cite{Hur:2020:SSM} and moves closer to the accuracy of semi-supervised methods with significantly faster runtime.}
\label{table:eval_sf_kitti_test}
\end{table}

\begin{figure}[t!]
\centering
\scriptsize
\setlength\tabcolsep{0.1pt}
\renewcommand{\arraystretch}{0.2}
\begin{tabular}{>{\centering\arraybackslash}m{.33\linewidth} >{\centering\arraybackslash}m{.33\linewidth} >{\centering\arraybackslash}m{.33\linewidth}}
	\includegraphics[width=\linewidth]{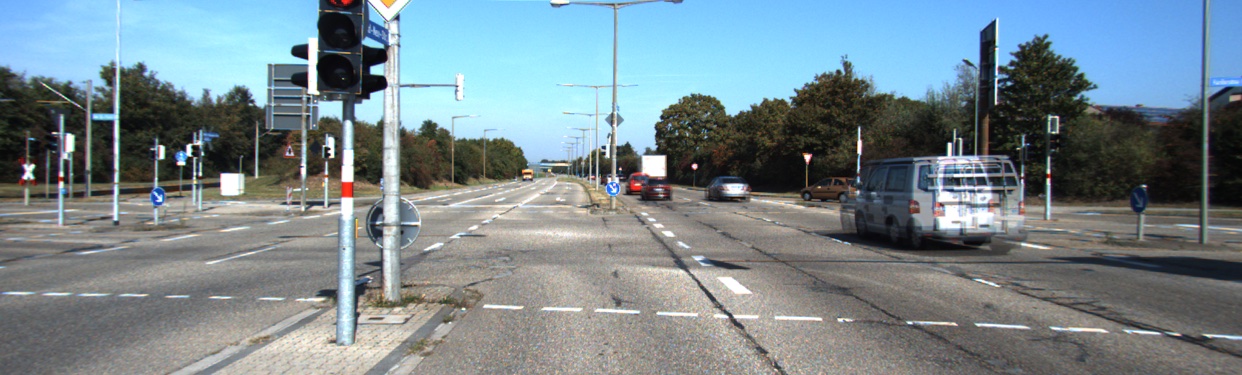} &
	\includegraphics[width=\linewidth]{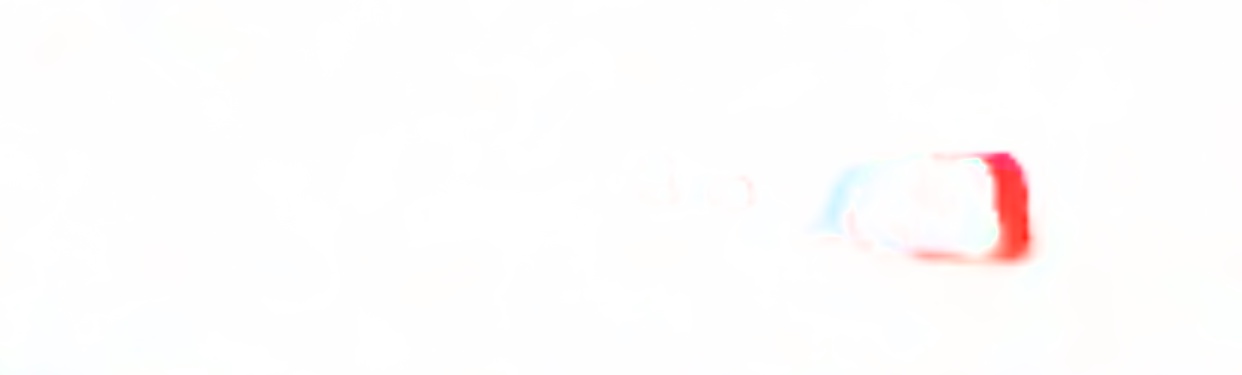} &
	\includegraphics[width=\linewidth]{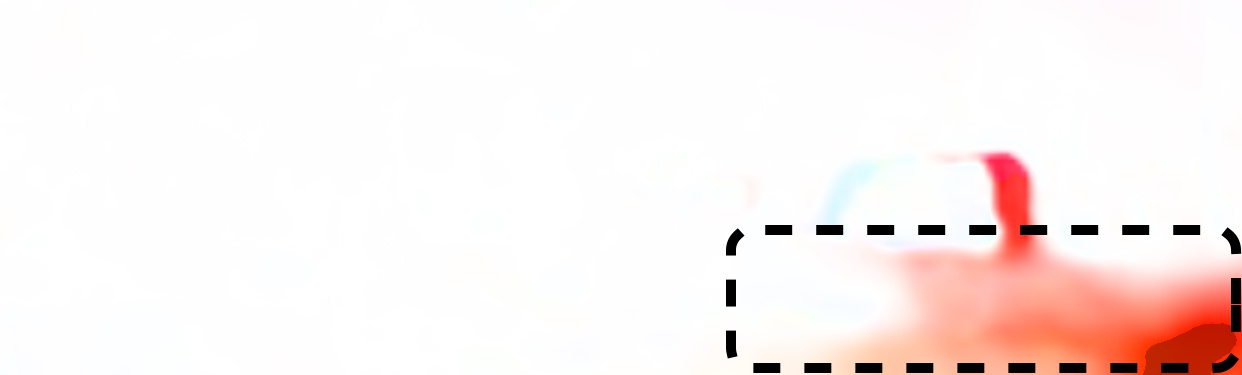} \\
	\includegraphics[width=\linewidth]{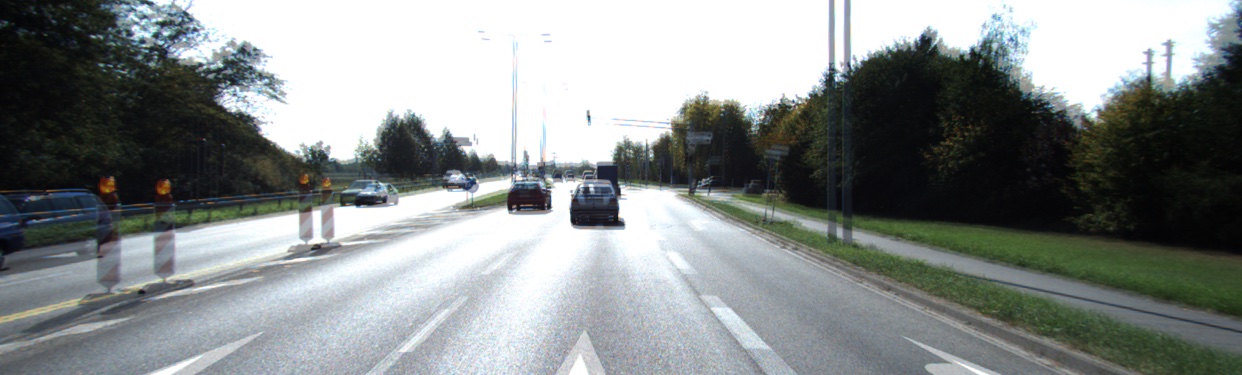} &
	\includegraphics[width=\linewidth]{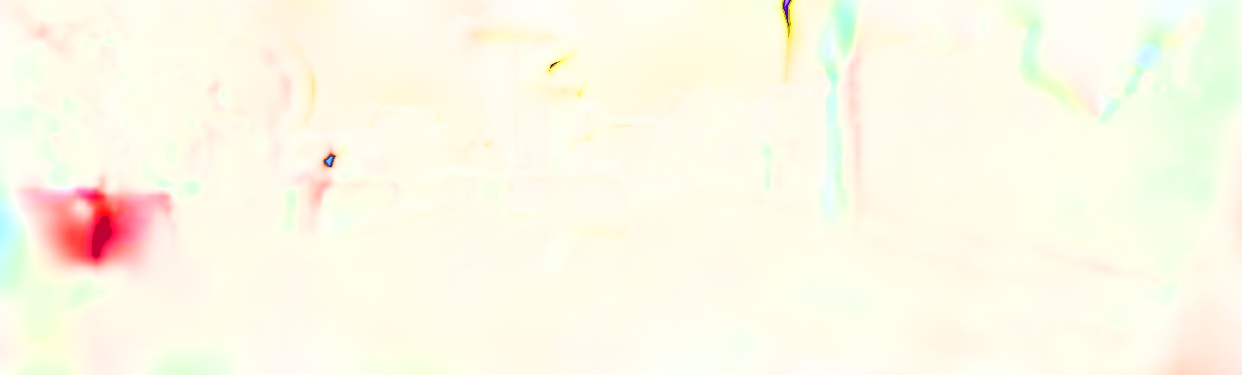} &
	\includegraphics[width=\linewidth]{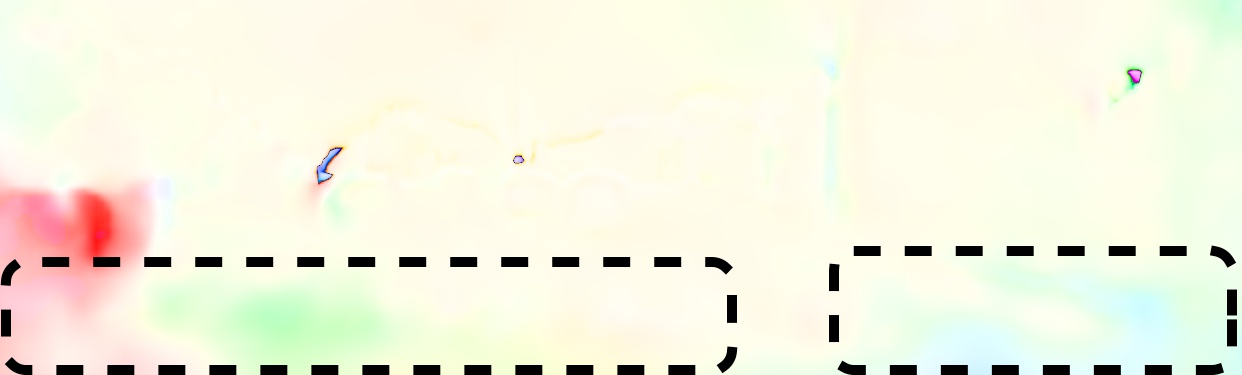} \\
	\includegraphics[width=\linewidth]{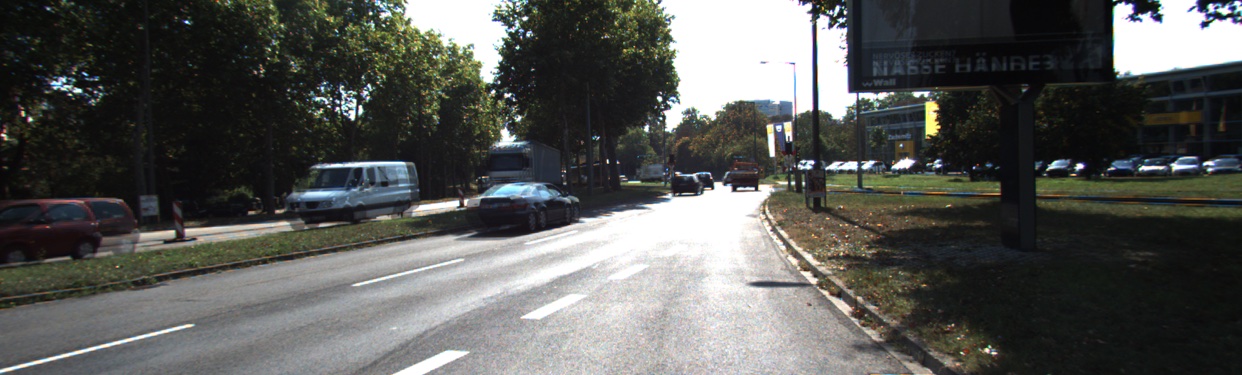} &
	\includegraphics[width=\linewidth]{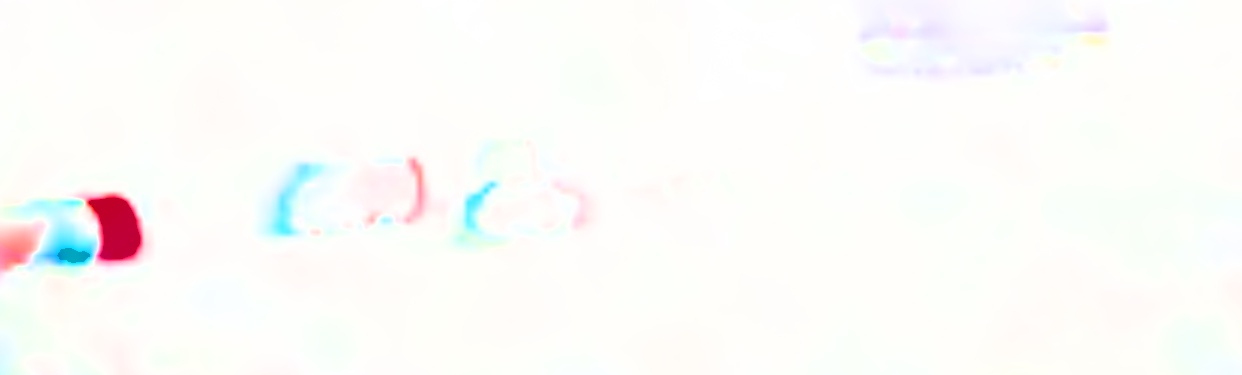} &
	\includegraphics[width=\linewidth]{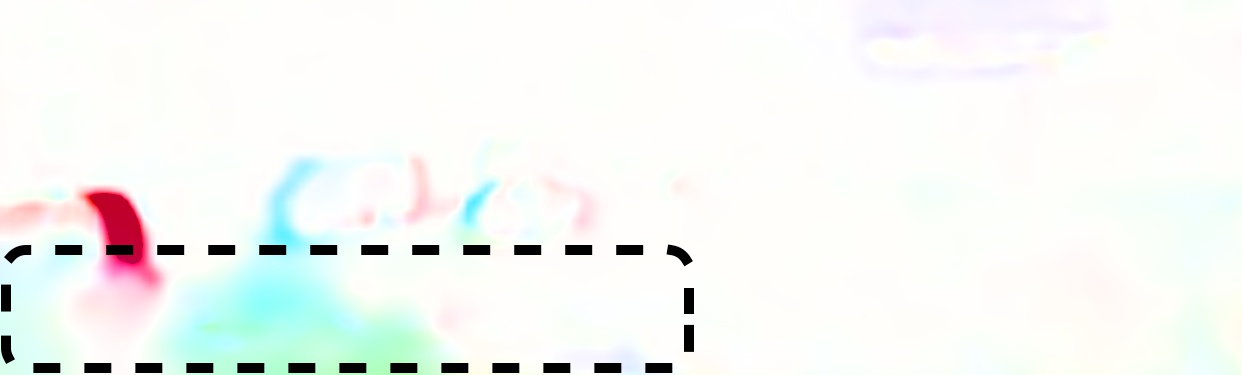} \\ [2em] \\
	(a) Overlayed input images & (b) Ours & (c) Direct baseline \cite{Hur:2020:SSM} \\
\end{tabular}
\caption{\textbf{Qualitative comparison of temporal consistency}: Each scene shows \emph{(a)} overlayed input images and scene flow difference maps of \emph{(b)} our method, and \emph{(c)} the direct baseline of \cite{Hur:2020:SSM}, visualized using an optical flow color coding. The \emph{dashed regions} highlight our more temporally consistent estimates.}
\label{fig:temporal_consistency}
\vspace{-0.5em}
\end{figure}

We further evaluate our model on the KITTI Scene Flow benchmark and compare with monocular scene flow approaches based on self- or semi-supervised learning in \cref{table:eval_sf_kitti_test}.
Our model consistently outperforms \cite{Hur:2020:SSM}.
Optionally fine-tuning (-ft) with 200 annotated pairs, our approach reduces the gap to semi-supervised methods that use a large amount of 3D LiDAR data \cite{Brickwedde:2019:MSF} or multiple synthetic datasets for optical flow \cite{Yang:2020:UOF}.
Yet, our model offers significantly faster runtime (\eg, $650\times$ faster than \cite{Brickwedde:2019:MSF} and $4\times$ faster than \cite{Yang:2020:UOF}).
Our model can thus exploit available labeled datasets for accuracy gains while keeping the same runtime.

\begin{figure*}[hbt!]
\centering
\scriptsize
\setlength\tabcolsep{0.1pt}
\renewcommand{\arraystretch}{0.2}
\begin{tabular}{>{\centering\arraybackslash}m{.20\textwidth} >{\centering\arraybackslash}m{.20\textwidth} >{\centering\arraybackslash}m{.20\textwidth} >{\centering\arraybackslash}m{.20\textwidth} >{\centering\arraybackslash}m{.20\textwidth}}
	\multicolumn{5}{c}{  \footnotesize \textcolor{customgray}{$\blacksquare$} Both methods correct \quad \quad \textcolor{customblue}{$\blacksquare$} Ours is correct, \cite{Hur:2020:SSM} is not \quad \quad \textcolor{customred}{$\blacksquare$} \cite{Hur:2020:SSM} is correct, ours is not \quad \quad \textcolor{customyellow}{$\blacksquare$} Both failed}\\ [0.5em]
	\includegraphics[width=\linewidth]{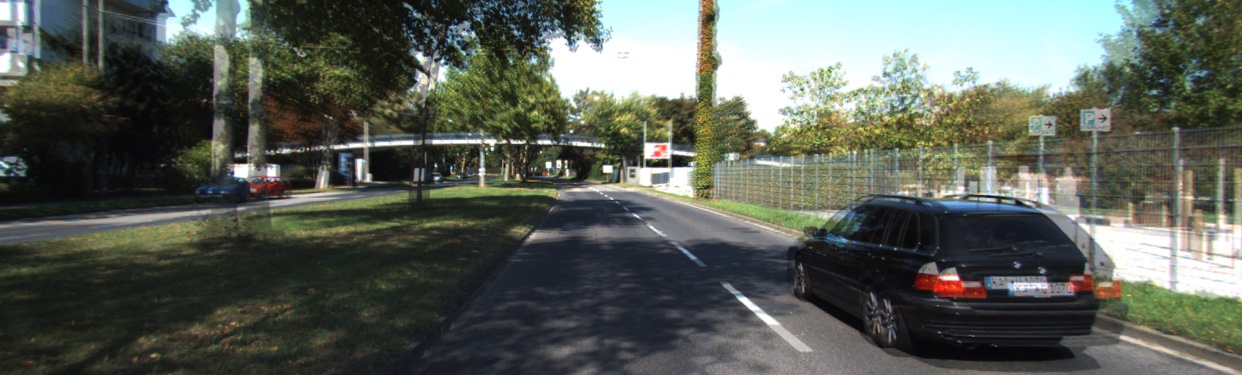} &
	\includegraphics[width=\linewidth]{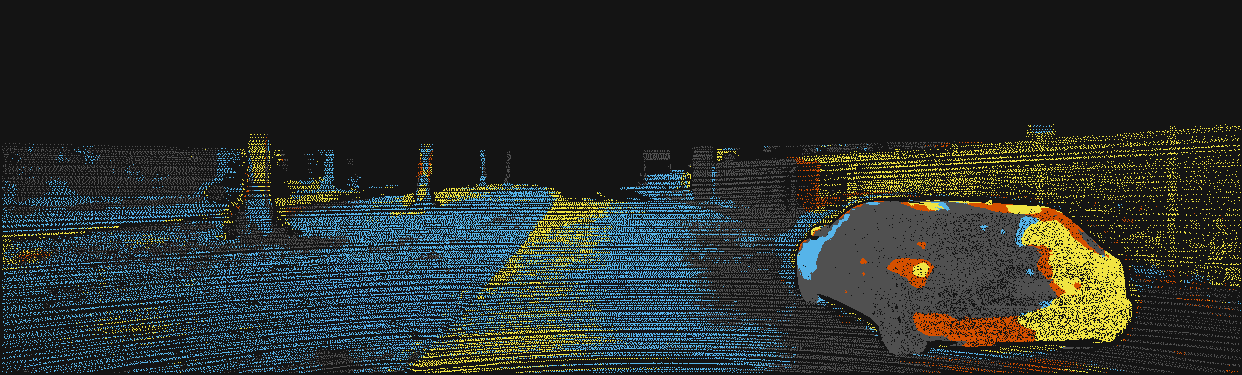} &
	\includegraphics[width=\linewidth]{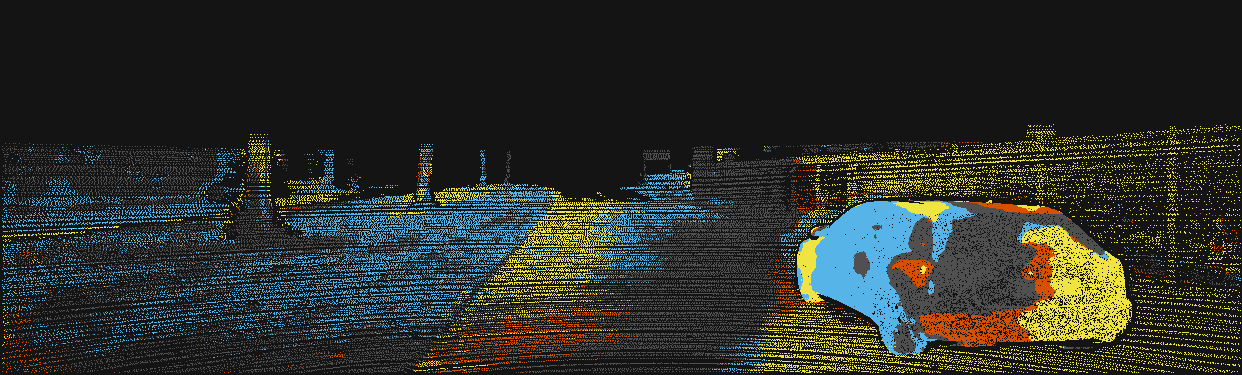} &
	\includegraphics[width=\linewidth]{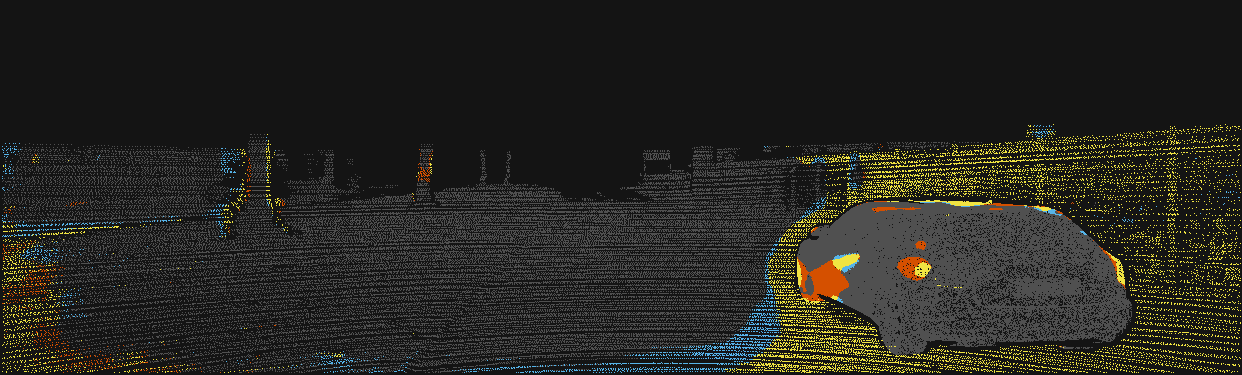} &
	\includegraphics[width=\linewidth]{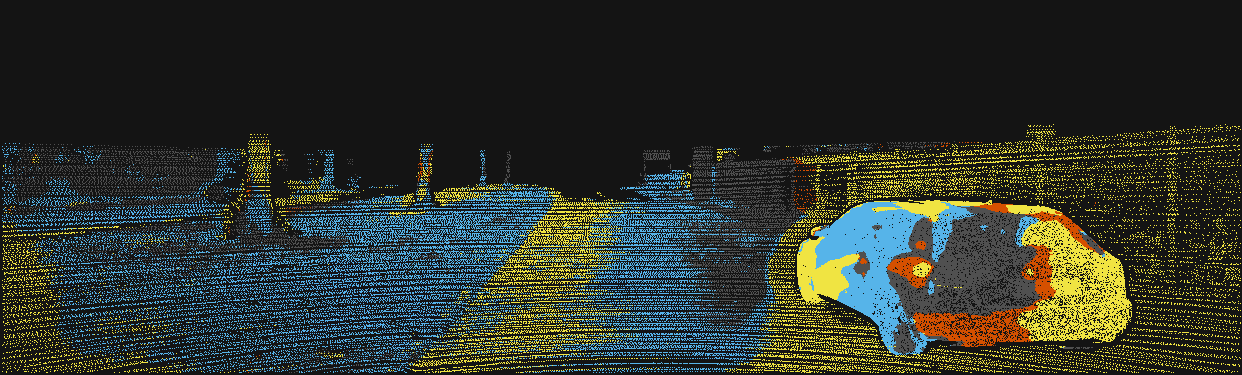} \\
	\includegraphics[width=\linewidth]{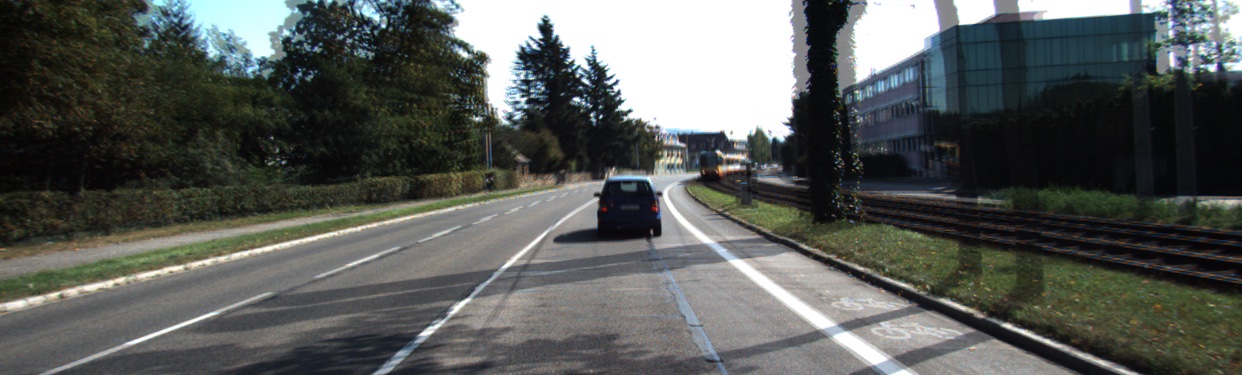} &
	\includegraphics[width=\linewidth]{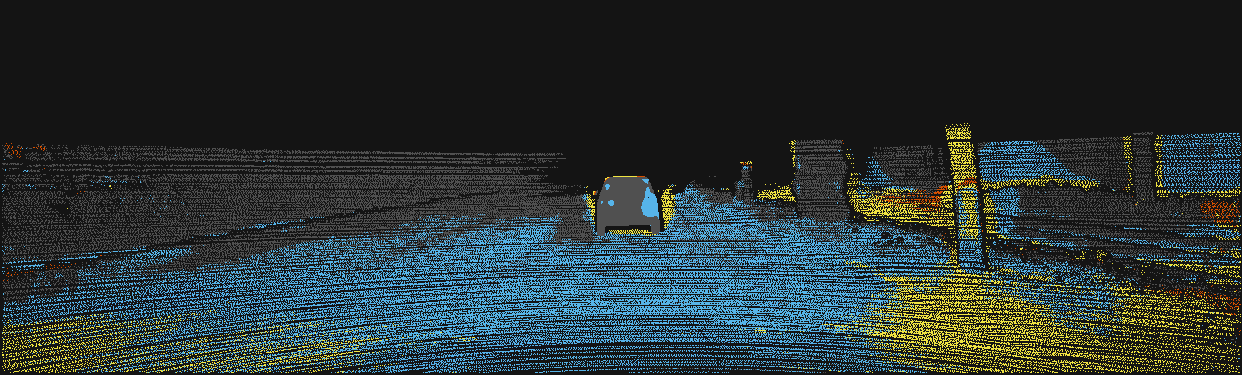} &
	\includegraphics[width=\linewidth]{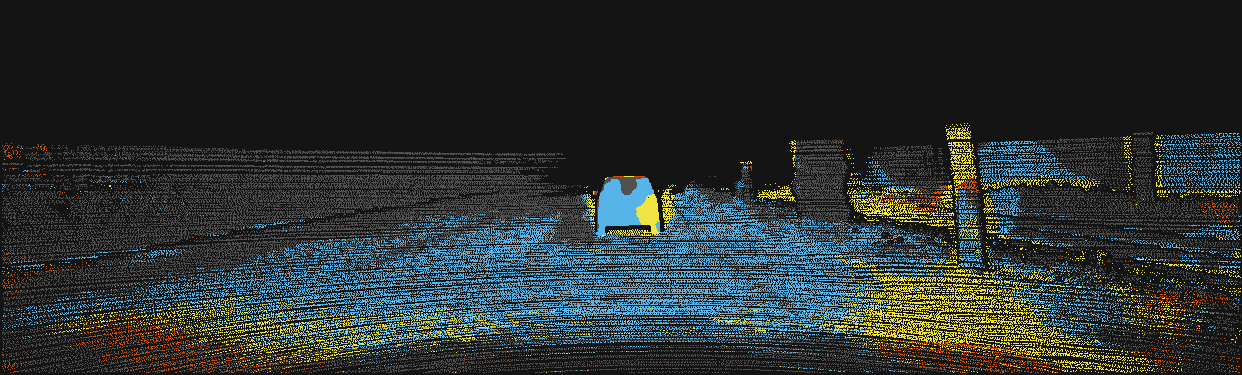} &
	\includegraphics[width=\linewidth]{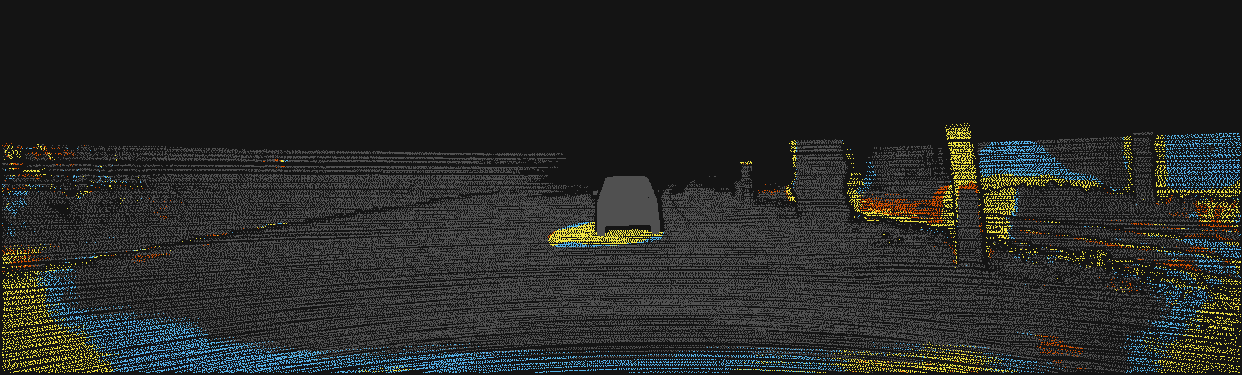} &
	\includegraphics[width=\linewidth]{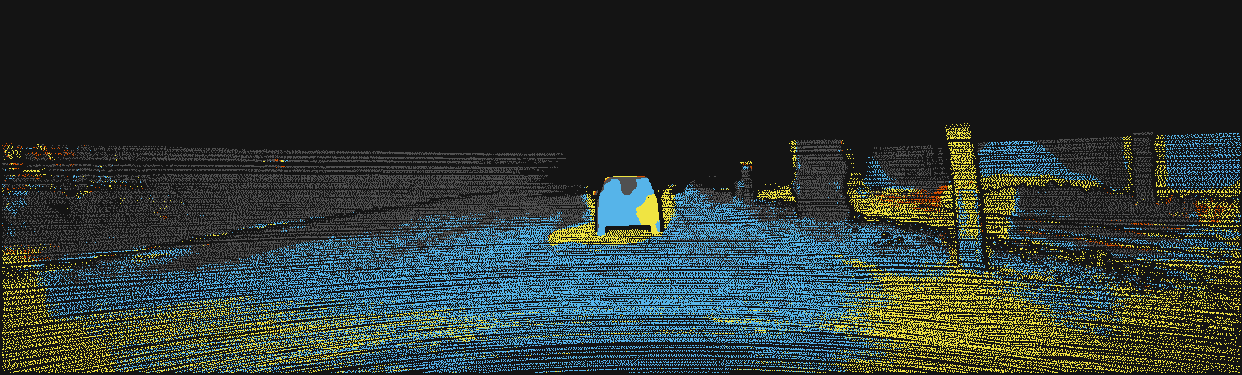} \\
	\includegraphics[width=\linewidth]{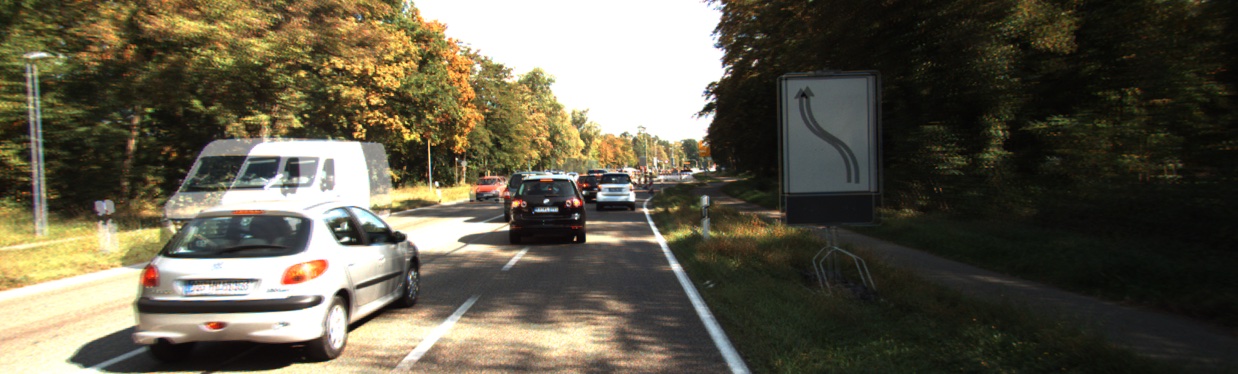} &
	\includegraphics[width=\linewidth]{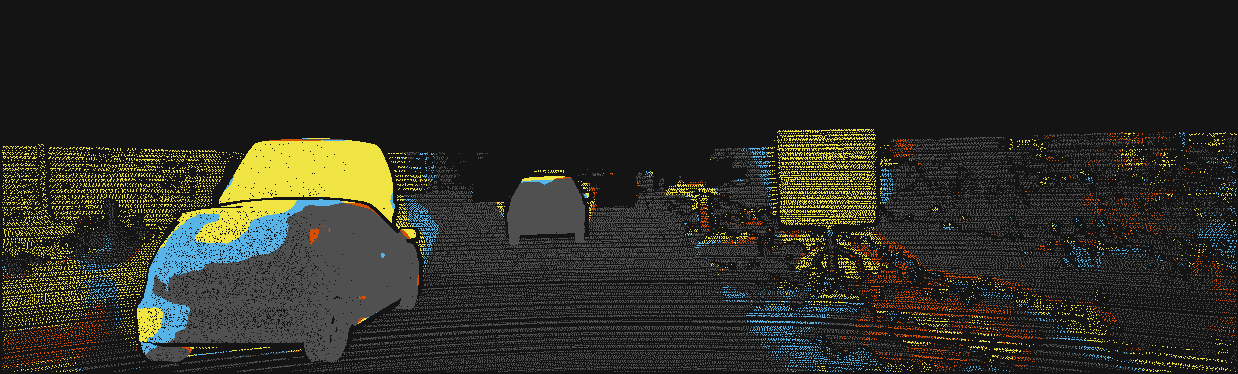} &
	\includegraphics[width=\linewidth]{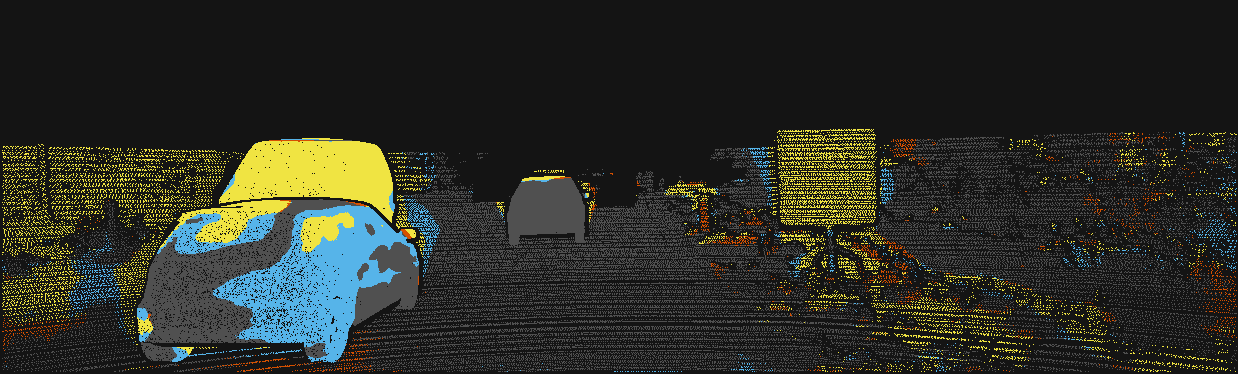} &
	\includegraphics[width=\linewidth]{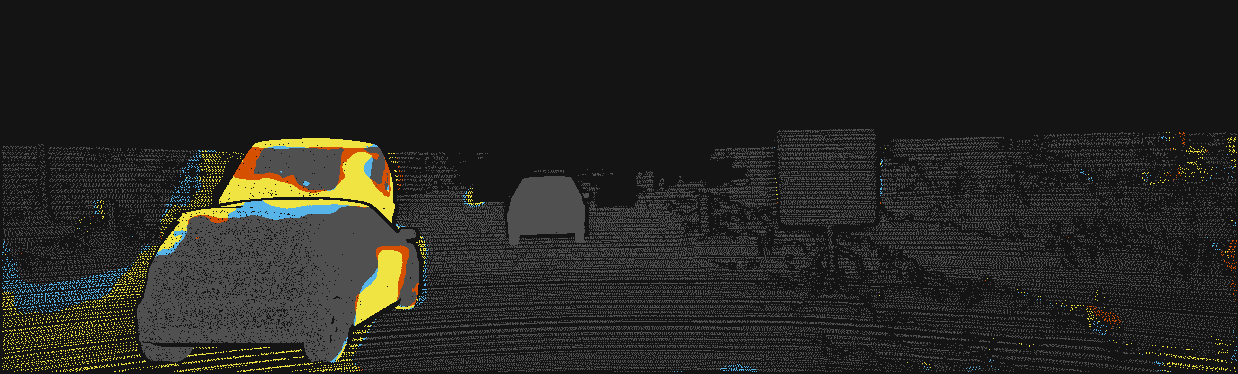} &
	\includegraphics[width=\linewidth]{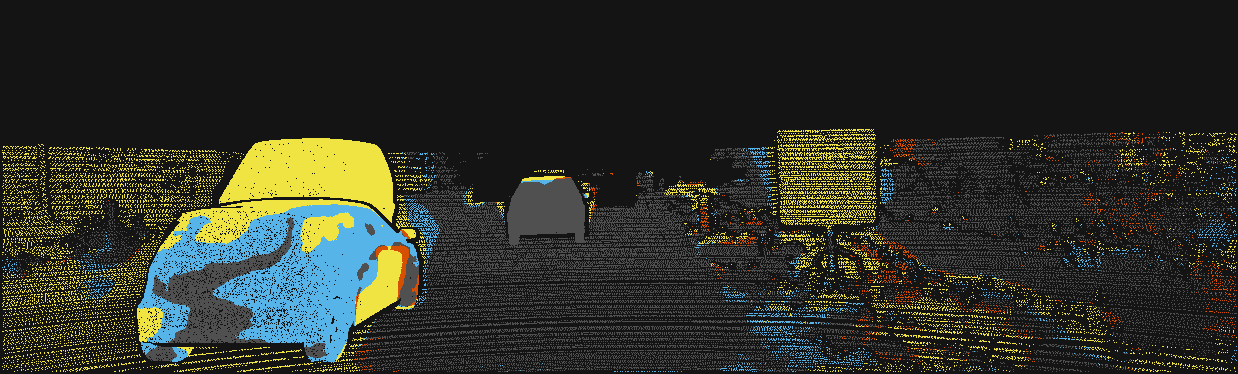} \\
	 [2.25em] \\
	(a) Overlayed input images & (b) Disparity error map & (c) Disparity change error map & (d) Optical flow error map & (e) Scene flow error map \\[0.5em]
\end{tabular}
\caption{\textbf{Qualitative comparison with the direct baseline of \cite{Hur:2020:SSM}}: Each scene shows \emph{(a)} overlayed input images and error maps for \emph{(b)}~disparity, \emph{(c)} disparity change, \emph{(d)} optical flow, and \emph{(e)} scene flow. Please refer to the color code above the figure.}
\label{fig:qualitative_comp}
%\vspace{-0.5em}
\end{figure*}

\subsection{Temporal consistency}
\label{subsec:temporal_consistency}
We evaluate the temporal consistency of our model, comparing to the direct baseline of \cite{Hur:2020:SSM}.
Lacking multi-frame metrics, we subtract two temporally consecutive scene flow estimates, project to optical flow, and visualize the result in \cref{fig:temporal_consistency}.
This shows how scene flow changes over time at the same pixel location.
Our model produces visibly more temporally consistent scene flow, especially near out-of-bound regions and foreground objects.\textsuperscript{\ref{footnote_supp2}}

\subsection{Qualitative comparison}
\label{subsec:qual_comp}
In \cref{fig:qualitative_comp}, we qualitatively compare our model with the direct baseline of \cite{Hur:2020:SSM} by visualizing where the accuracy gain mainly originates from.
Our method outputs more accurate scene flow especially on planar road surfaces (\nth{1} and \nth{2} row), out-of-bound regions (\nth{2} row), and foreground objects (\nth{1} and \nth{3} row).
Especially the accuracy gain on foreground objects and planar road surfaces originates from more accurate estimates on the disparity and disparity change map.\textsuperscript{\ref{footnote_supp2}}

\subsection{Generalization to other datasets}
\label{subsec:generalization}

\begin{figure}[]
\centering
\scriptsize
\setlength\tabcolsep{0.1pt}
\renewcommand{\arraystretch}{0.2}
\begin{tabular}{>{\centering\arraybackslash}m{.16\textwidth} >{\centering\arraybackslash}m{.16\textwidth} >{\centering\arraybackslash}m{.16\textwidth}}
	\includegraphics[width=\linewidth]{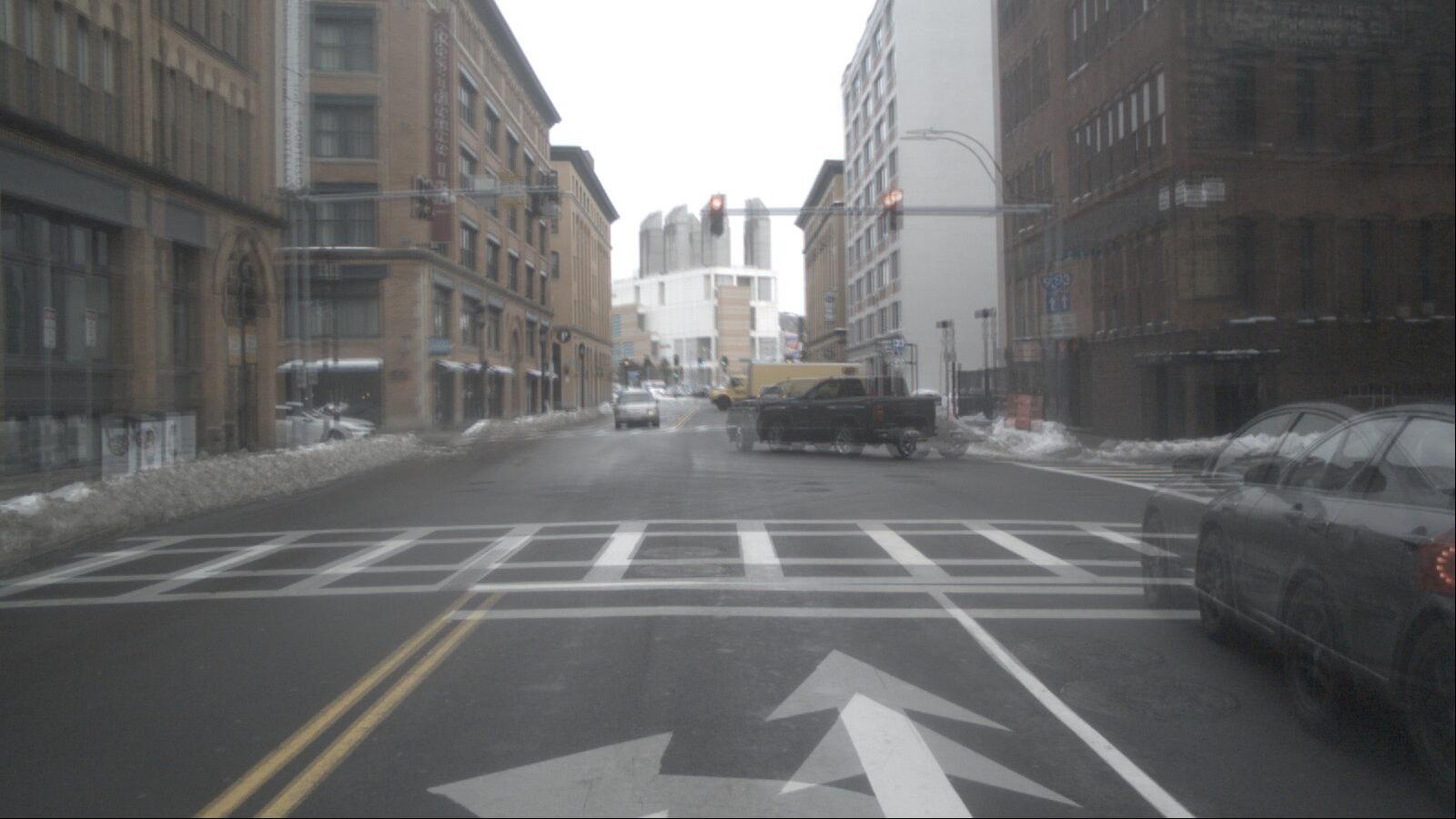} &
	\includegraphics[width=\linewidth]{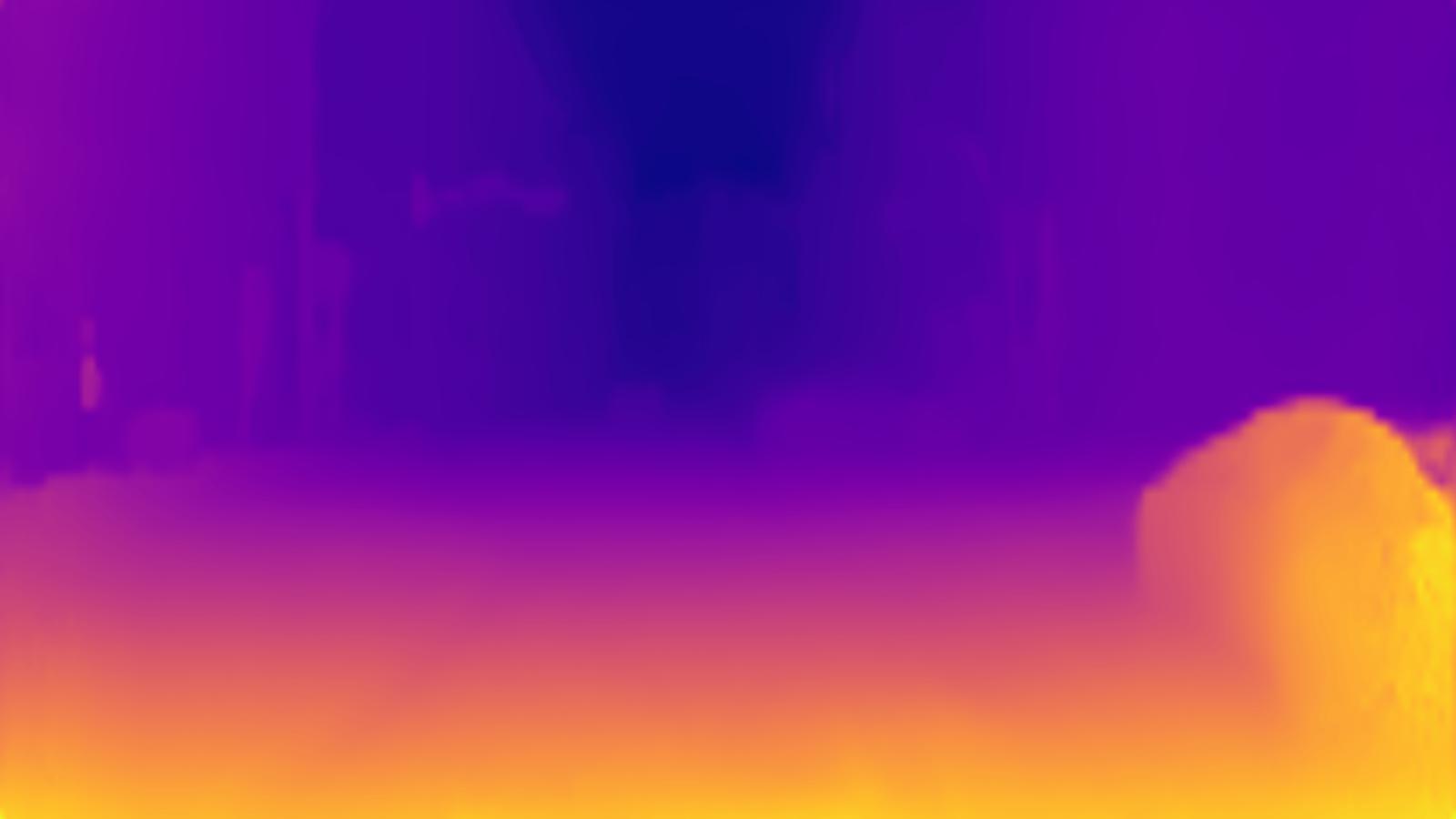} &
	\includegraphics[width=\linewidth]{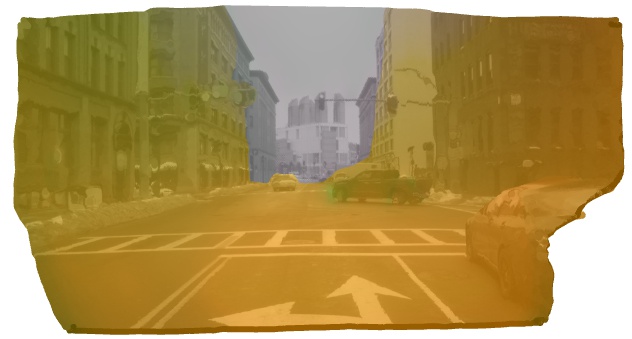} \\
	\includegraphics[width=\linewidth]{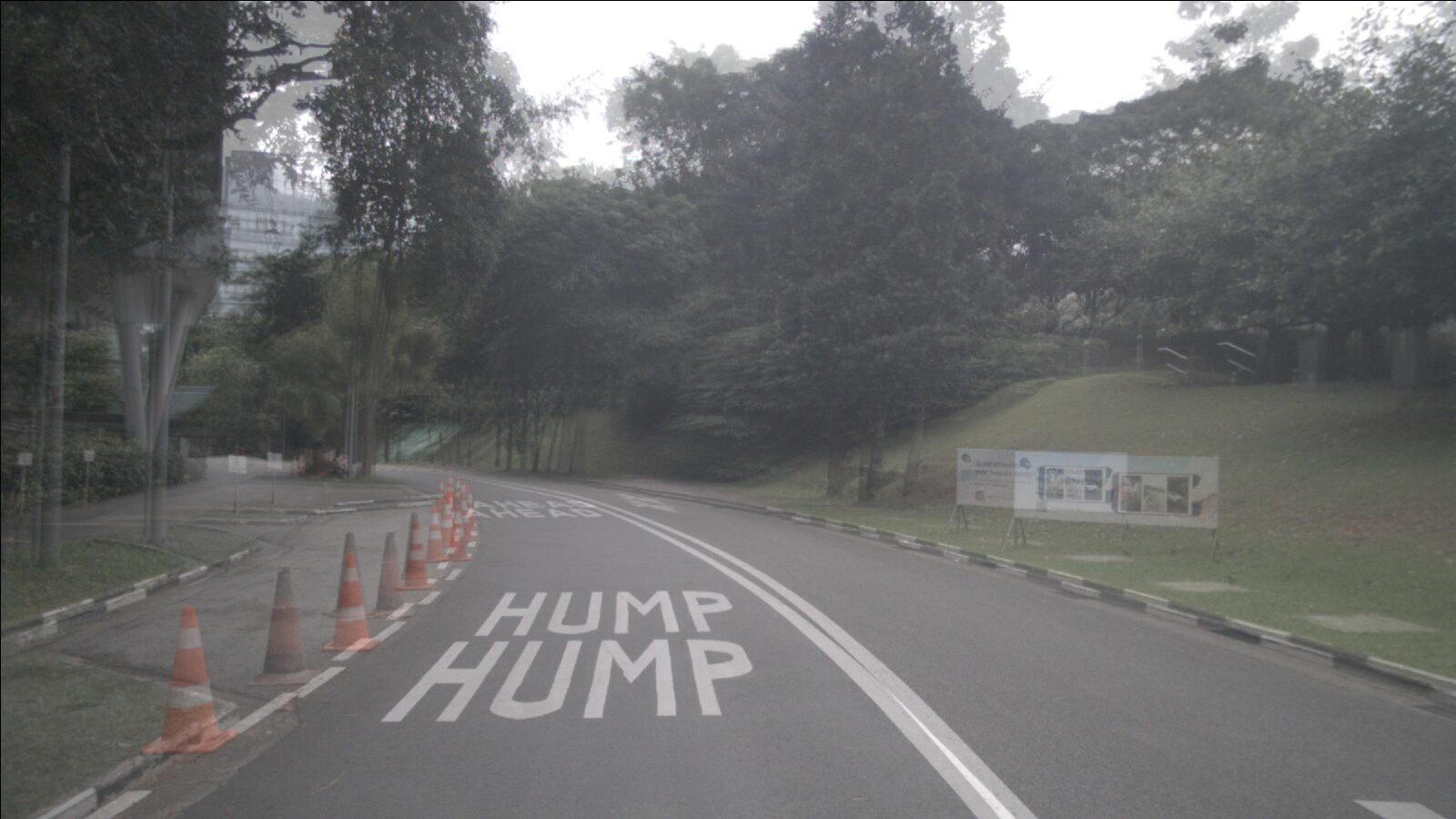} &
	\includegraphics[width=\linewidth]{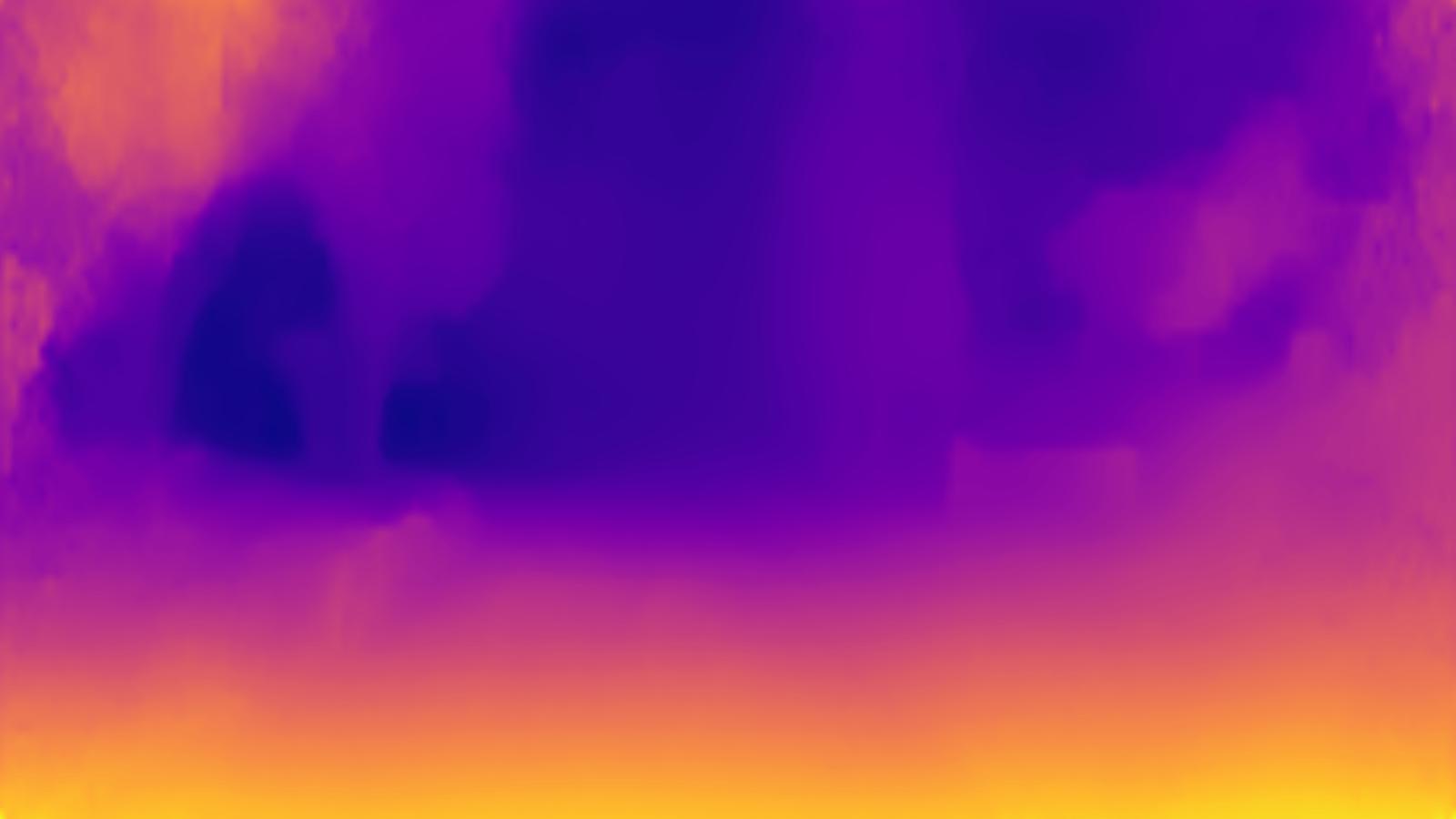} &
	\includegraphics[width=0.95\linewidth]{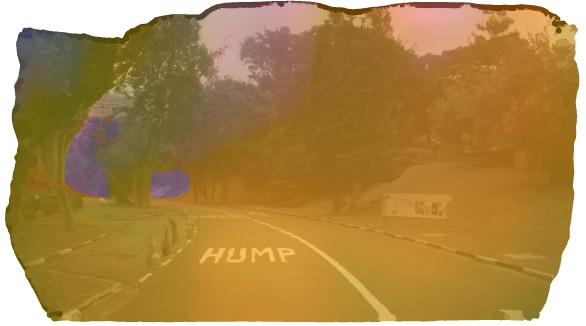} \\ [1em] \\

	\includegraphics[width=\linewidth]{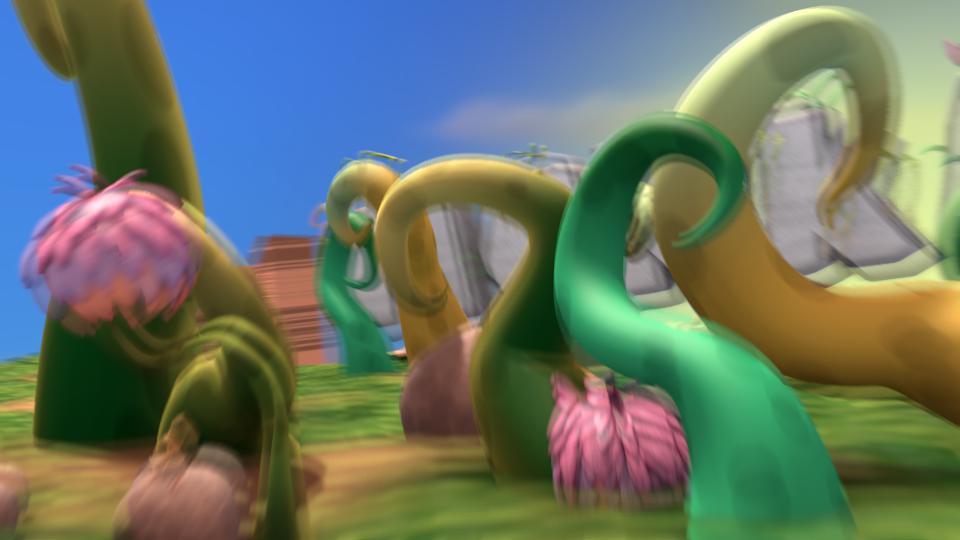} &
	\includegraphics[width=\linewidth]{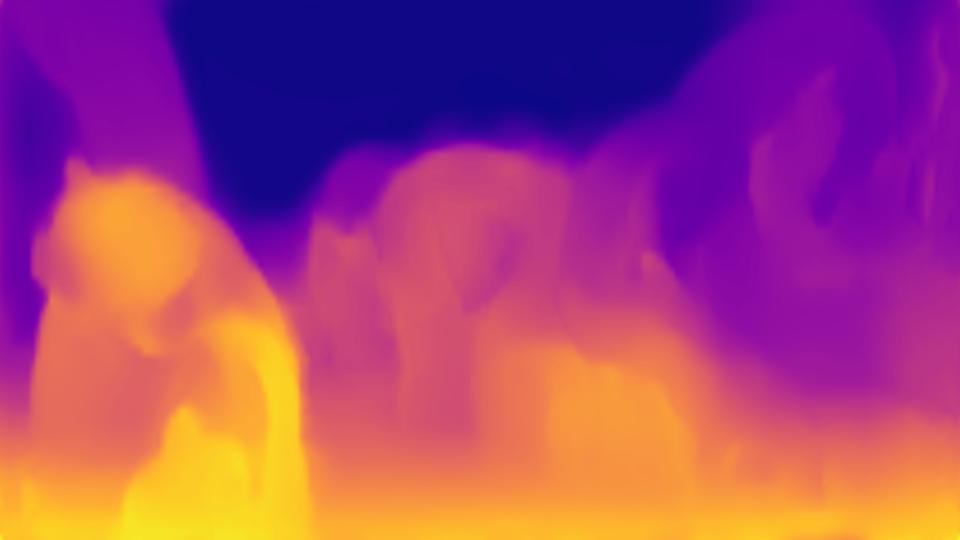} &
	\includegraphics[width=0.9\linewidth]{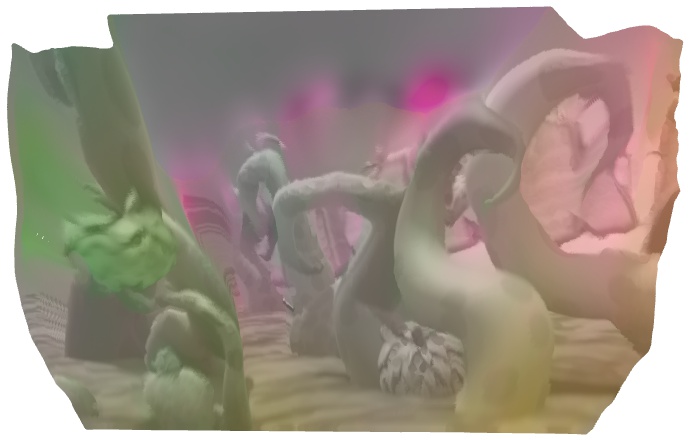} \\
	\includegraphics[width=\linewidth]{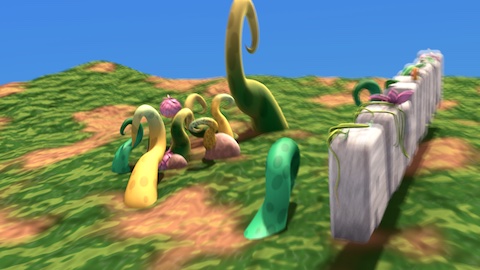} &
	\includegraphics[width=\linewidth]{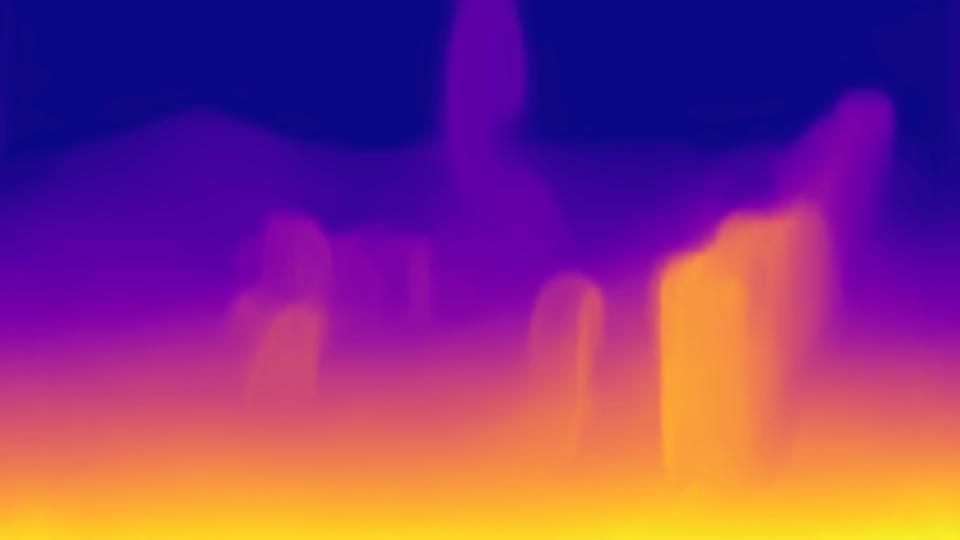} &
	\includegraphics[width=\linewidth]{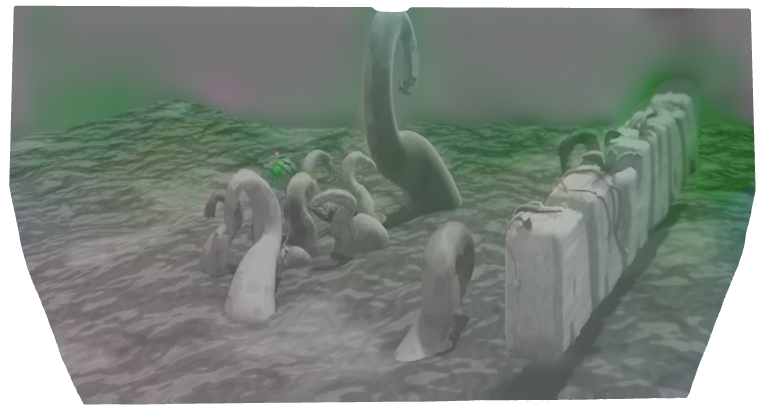} \\ [1em] \\

	\includegraphics[width=\linewidth]{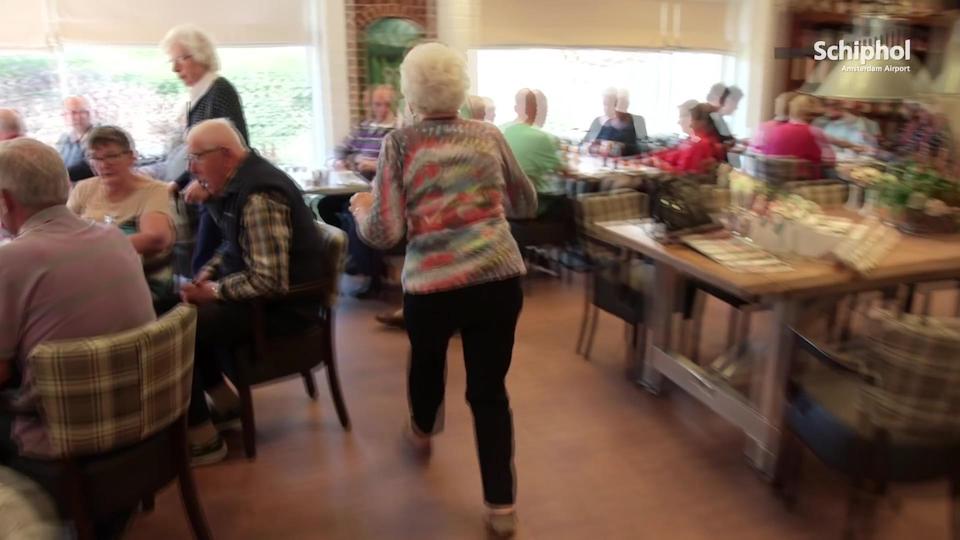} &
	\includegraphics[width=\linewidth]{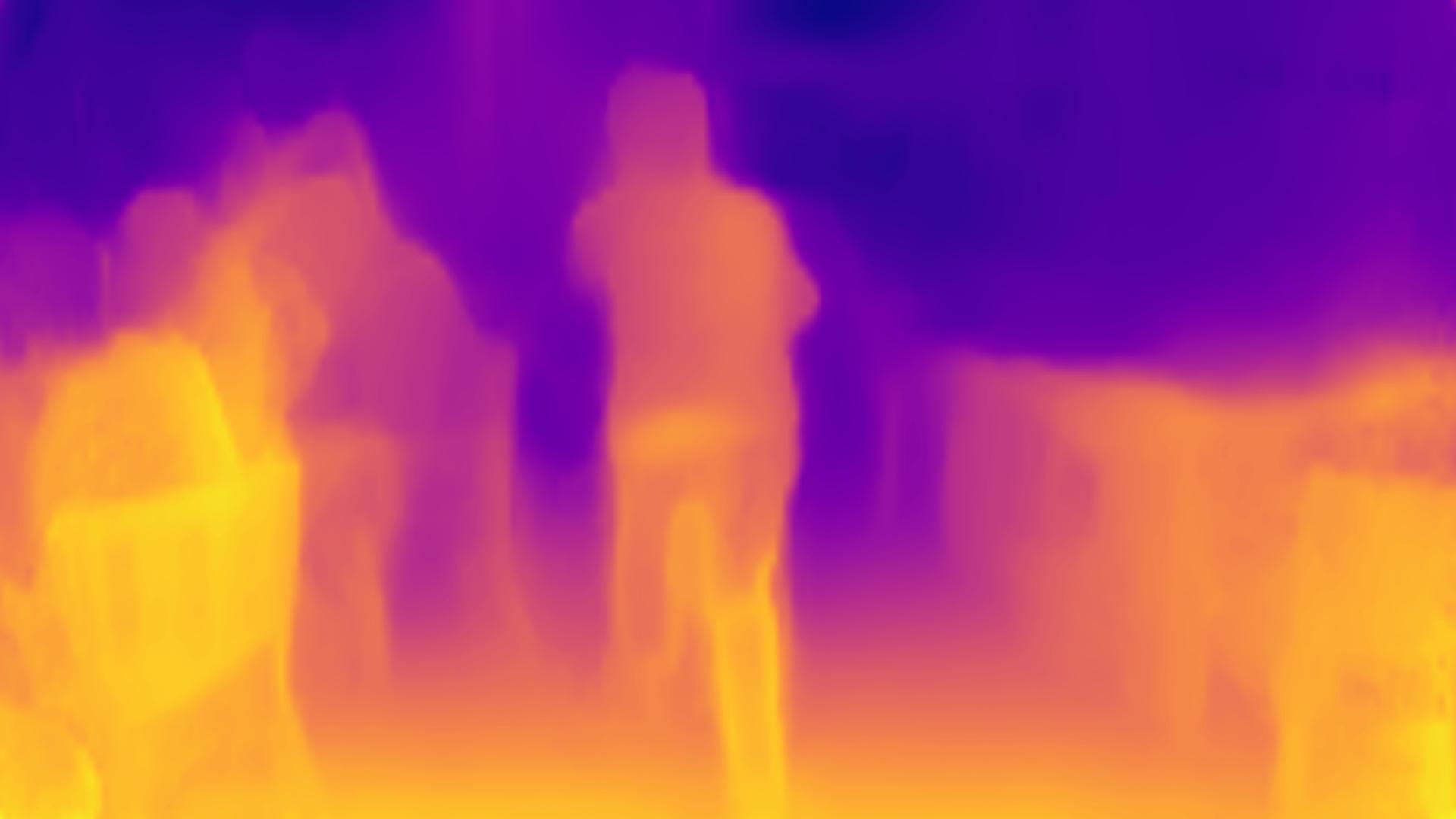} &
	\includegraphics[width=0.9\linewidth]{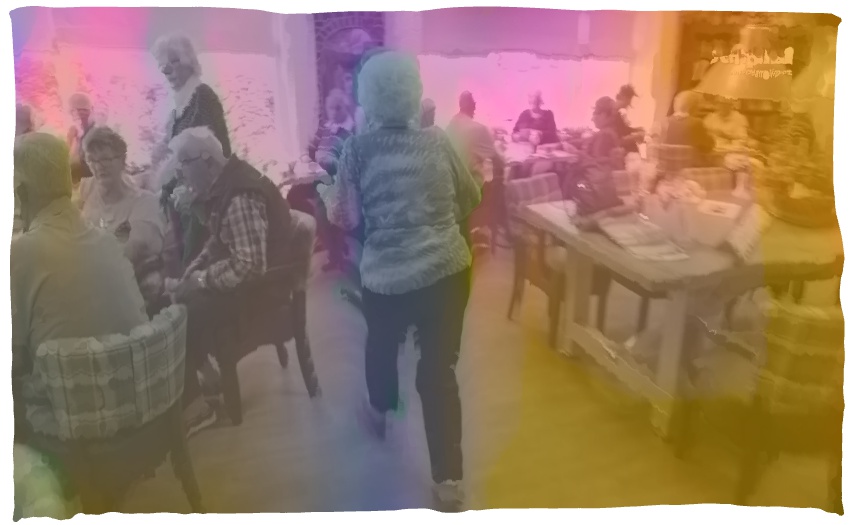} \\
	\includegraphics[width=\linewidth]{figures/supp_generalization/davis/bike-packing_00012_overlay.jpg} &
	\includegraphics[width=\linewidth]{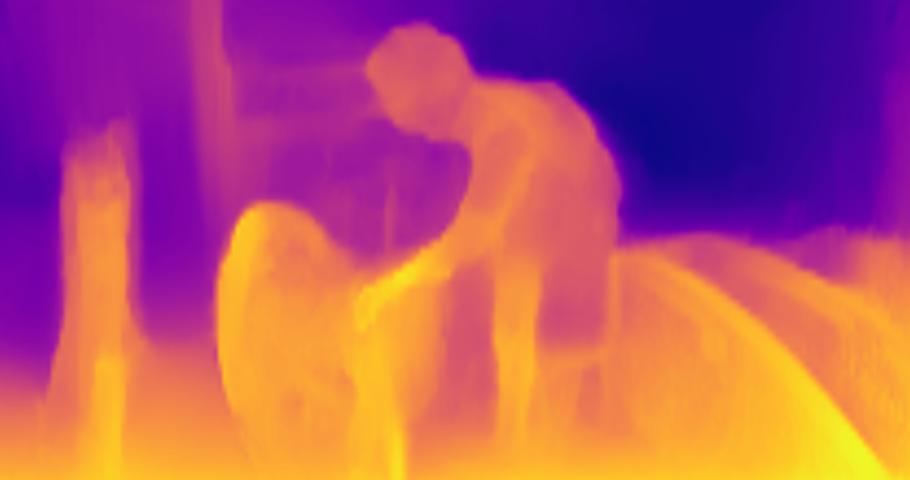} &
	\includegraphics[width=0.95\linewidth]{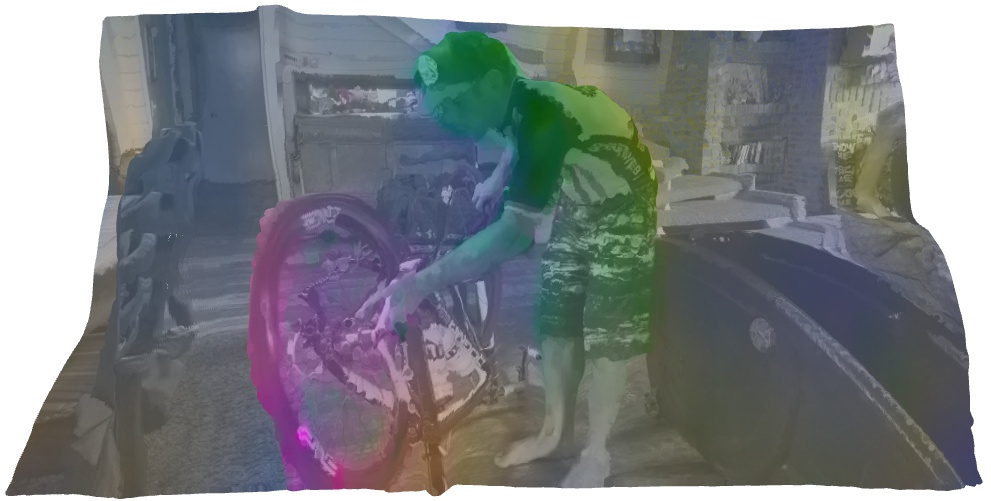} \\

	\\ \\ (a) Overlayed input images & (b) Depth map & (c) Scene flow visualization \\[0.5em]
\end{tabular}
\caption{\textbf{Generalization to nuScenes (top), Monkaa (middle), and DAVIS (bottom) datasets}: Each scene shows \emph{(a)} overlayed input images, \emph{(b)} depth, and \emph{(c)} a 3D scene flow visualization.}
\label{fig:generalization}
\vspace{-0.5em}
\end{figure}

We test the generalization of our model trained on KITTI \cite{Geiger:2013:VMR} to other datasets, such as the nuScenes \cite{Caesar:2020:NSA}, Monkaa \cite{Mayer:2016:ALD}, and DAVIS \cite{Perazzi:2016:ABD} datasets.
\cref{fig:generalization} provides visual examples.
Our method demonstrates good generalization to the real-world nuScenes dataset, but shows visually less accurate results on the synthetic domain (\ie~Monkaa), as can be expected.
Interestingly on DAVIS, our method demonstrates reasonable performance on completely unseen domains (\eg, indoor, ego-centric).\textsuperscript{\ref{footnote_supp2}}

% !TEX root = ../arxiv.tex
\section{Conclusion}

We proposed a multi-frame monocular scene flow network based on self-supervised learning.
Starting from the recent self-supervised two-frame network of \cite{Hur:2020:SSM}, we first pointed out limitations of the decoder design and introduced an advanced two-frame baseline, which is stable to train and already improves the accuracy.
Our multi-frame model then exploits the temporal coherency of 3D scene flow using overlapping triplets of input frames and temporally propagating previous estimates via a convolutional LSTM.
Using forward warping in the ConvLSTM turned out to be crucial for accurate temporal propagation.
An occlusion-aware census loss and a gradient detaching strategy further boost the accuracy and training practicality.
Our model achieves state-of-the-art scene flow accuracy among self-supervised methods, while even yielding faster runtime, and reduces the accuracy gap to less efficient semi-supervised methods.

Future work should consider extending our self-supervised approach to challenging, uncontrolled capture setups with variable training baselines for even broader applicability.
Also, while we do not explicitly model camera ego-motion mainly for the simplicity of the pipeline, additionally exploiting ego-motion can yield further benefits, \eg, for driving scenarios where rigid motion dominates.

{\small
\myparagraph{Acknowledgements.} This project has received funding from the European Research Council (ERC) under the European Union’s Horizon 2020 research and innovation programme (grant agreement No.~866008).
}

{\small
\balance
\bibliographystyle{bib/ieee_fullname}

}

%% Supplemental
\title{Self-Supervised Multi-Frame Monocular Scene Flow \\ {\large -- Supplementary Material --}}
\author{Junhwa Hur$^1$ \qquad\qquad Stefan Roth$^{1,2}$\\
$\ ^1$Department of Computer Science, TU Darmstadt \hspace{1cm} $\ ^2$ hessian.AI}
\maketitle
\appendix
\nobalance
%% Supplemental body
% !TEX root = ../arxiv.tex

In this supplementary material, we introduce our 3D scene flow color coding scheme and provide details on our refined backbone architecture, self-supervised loss function, implementation, and computational cost.
Then, we demonstrate additional results on temporal consistency, qualitative comparisons with the direct two-frame baseline (Self-Mono-SF \cite{Hur:2020:SSM}), and more experimental results for the generalization to other datasets.
Lastly, we provide preliminary results of our model trained on a vast amount of unlabeled web videos in a self-supervised manner.
We discuss the results as well as a current limitation of our method.
% For reproducibility, we attach our source code (\textbf{code.zip}) with pre-trained models included.
% The code and the pre-trained model will be made publicly available upon acceptance.

% !TEX root = ../arxiv.tex
\section{Scene Flow Color Coding}
\label{supp:sec:color_coding}

\begin{figure}[t]
\centering
\includegraphics[width=0.9\linewidth]{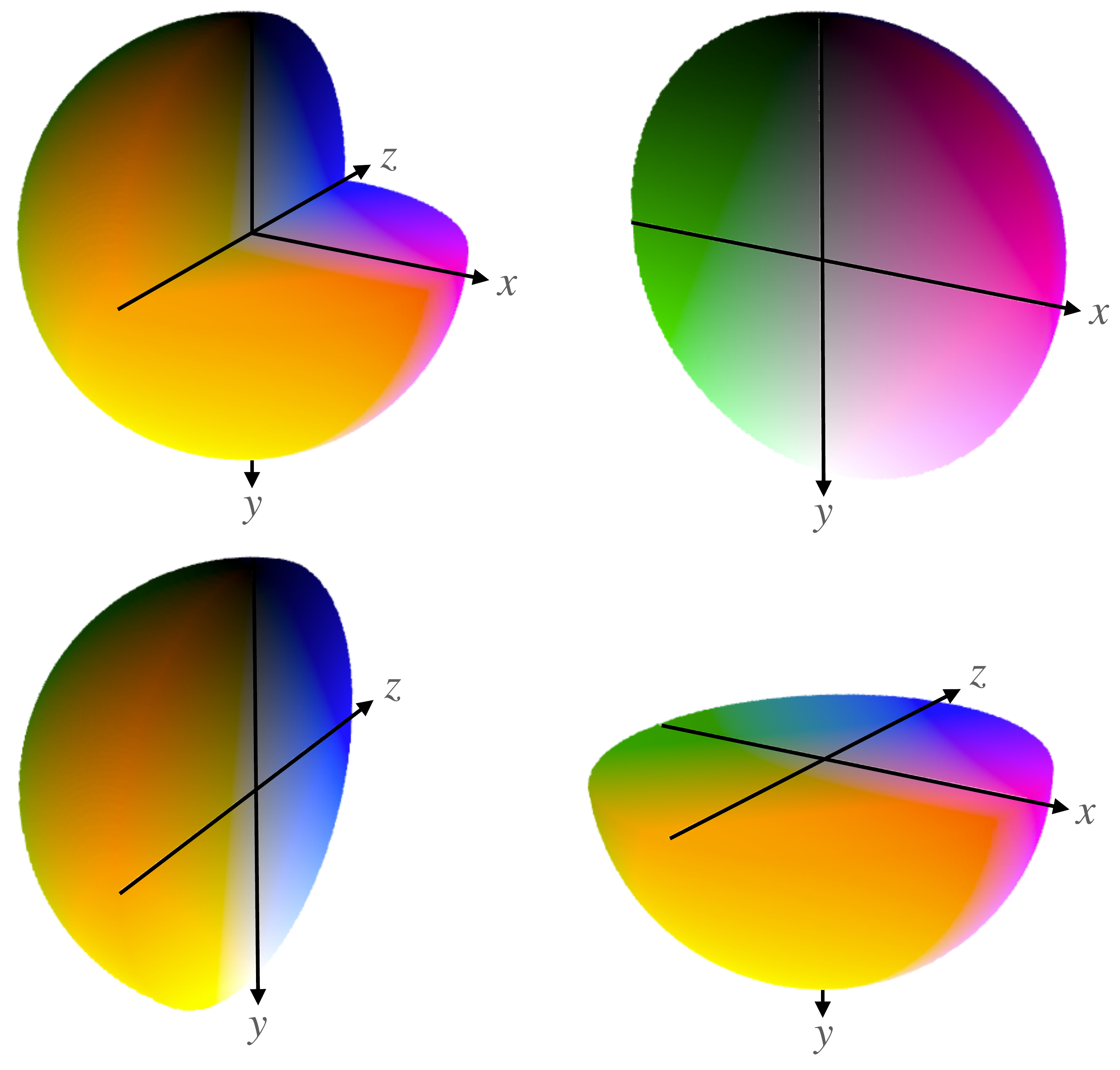}
% \subcaptionbox{A detailed view of the decoder at pyramid level $l$\label{fig:architecture_decoder}}{\includegraphics[width=0.94\linewidth]{figures/architecture_decoder.pdf}}
\caption{\textbf{3D scene flow color coding scheme using the CIE-LAB color space:} Each figure shows a sliced sphere along each plane for ease of visualization.}
\label{fig:color_coding}
%\vspace{-0.5em}
\end{figure}

For visualizing 3D scene flow in 2D image coordinates, we use the CIE-LAB color space, as visualized in \cref{fig:color_coding}.

% !TEX root = ../arxiv.tex
\section{Refined Backbone Architecture}
\label{supp:sec:architecture}

We provide a more in-depth analysis of our refined backbone two-frame architecture introduced in \cref{sec:approach:architecture} of the main paper.
We first present a simple empirical study of key findings from \cite{Jonschkowski:2020:WMU} and discuss which key factors can be carried over to monocular scene flow estimation.
Afterward, we demonstrate an accuracy analysis and the improved training stability of our refined architecture by discarding the context network and splitting the decoder.

\myparagraph{Empirical study on key findings from \cite{Jonschkowski:2020:WMU}.}
Jonschkowski~\etal~\cite{Jonschkowski:2020:WMU} provide a systematical analysis of the key design factors for highly accurate self-supervised optical flow.
We conduct an empirical study on which of their key findings are beneficial in the context of monocular 3D scene flow.
We report results on \emph{KITTI Scene Flow Training} \cite{Menze:2015:J3E,Menze:2018:OSF} using the scene flow metrics (\cf \cref{sec:exp:ablation} in the main paper).

\begin{table}[t]
\centering
\footnotesize
\setlength\tabcolsep{2pt}
\begin{tabular*}{\columnwidth}{@{\extracolsep{\fill}}l@{\hskip 1em}cccc@{}}
	\toprule
	Model & {D$1$-all} & {D$2$-all} & {Fl-all} & {SF-all} \\
	\midrule
	\cite{Hur:2020:SSM} $-$ Cont.~Net. & 34.24 & 37.32 & 25.06 & 50.49 \\
	\cite{Hur:2020:SSM} $-$ Cont.~Net. $+$ CV.~Norm. (\textbf{Baseline}) & \bfseries 31.91 & \bfseries 35.31 & \bfseries 24.80 & \bfseries 48.29 \\
	\midrule
	\multicolumn{5}{@{}l@{}}{Applying each row on top of the baseline above:} \\
	\midrule
	Census loss 						& 32.52 & \textcolor{blue}{34.54} & \textcolor{blue}{21.79} & \textcolor{blue}{45.61} \\
	Using one less pyramid level 		& 33.68 & \textcolor{blue}{35.03} & \textcolor{blue}{23.98} & \textcolor{blue}{47.77} \\
	Data distillation 					& 34.62 & 35.74 & \textcolor{blue}{24.09} & \textcolor{blue}{48.11} \\
	Using $640 \times 640$ resolution	& 33.12 & \textcolor{blue}{34.59} & \textcolor{blue}{22.43} & \textcolor{blue}{48.28} \\
	Level dropout & \multicolumn{4}{c}{(\emph{not converged})} \\
	\bottomrule
\end{tabular*}
\caption{\textbf{An empirical study of the key findings from \cite{Jonschkowski:2020:WMU}}: We take \cite{Hur:2020:SSM} after discarding the context network (Cont.~Net.) and applying the cost volume normalization (CV.~Norm.) as the baseline network.
Then we apply each key factor to the baseline and compare the scene flow accuracy.
Numbers colored in \textcolor{blue}{blue} outperform the baseline accuracy.}
\label{supp:table:empirical_study}
\vspace{-0.5em}
\end{table}

\begin{figure*}[t]
\centering
\subcaptionbox{Training loss, original train split \cite{Hur:2020:SSM} \label{supp:fig:loss_orig}}{\includegraphics[width=0.4\linewidth]{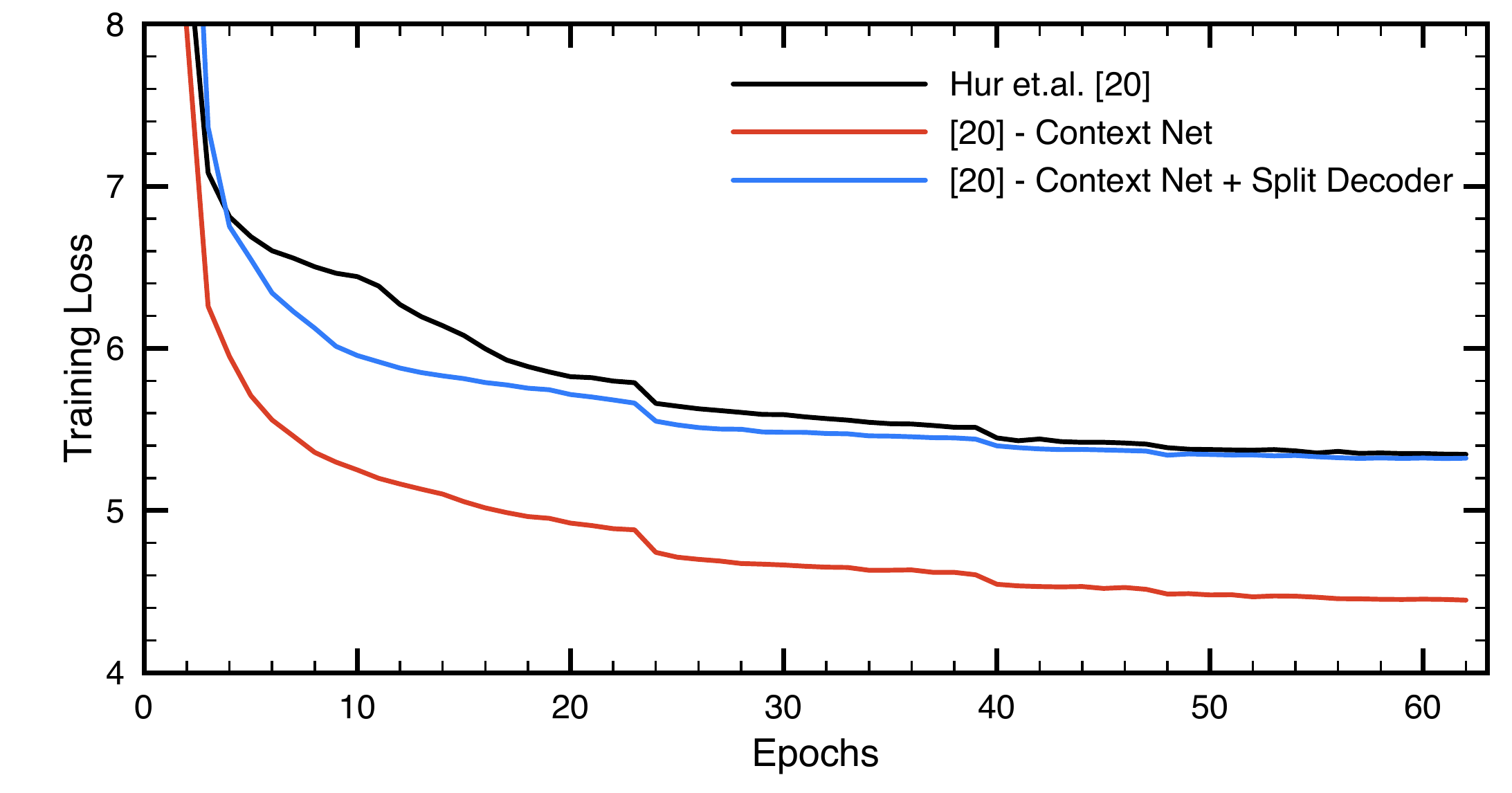}}
\hspace{1cm}
\subcaptionbox{Scene flow outlier rate, original train split \cite{Hur:2020:SSM} \label{supp:fig:accuracy_orig}}{\includegraphics[width=0.4\linewidth]{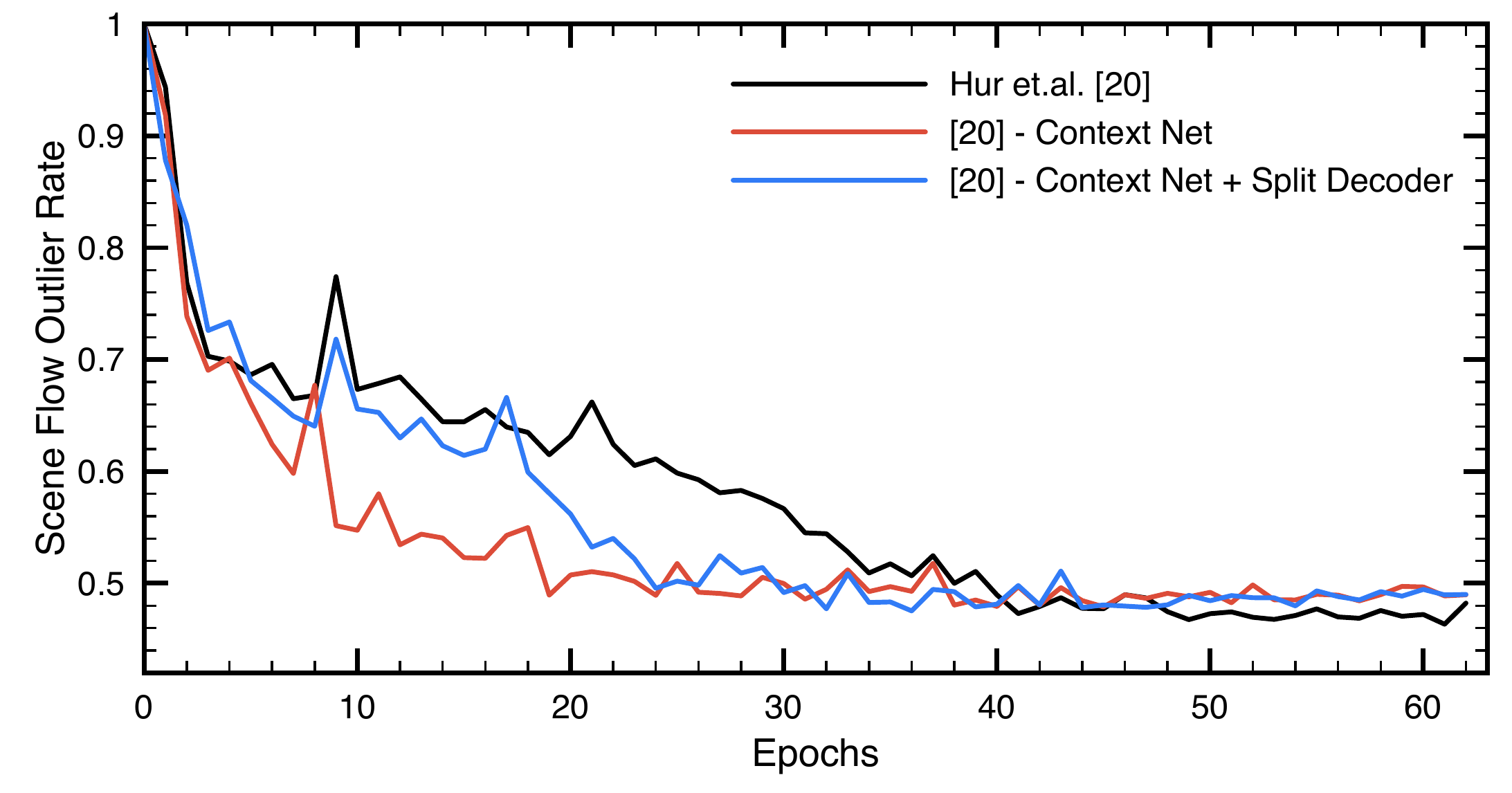}} \\
\subcaptionbox{Training loss, our train split \label{supp:fig:loss_our}}{\includegraphics[width=0.4\linewidth]{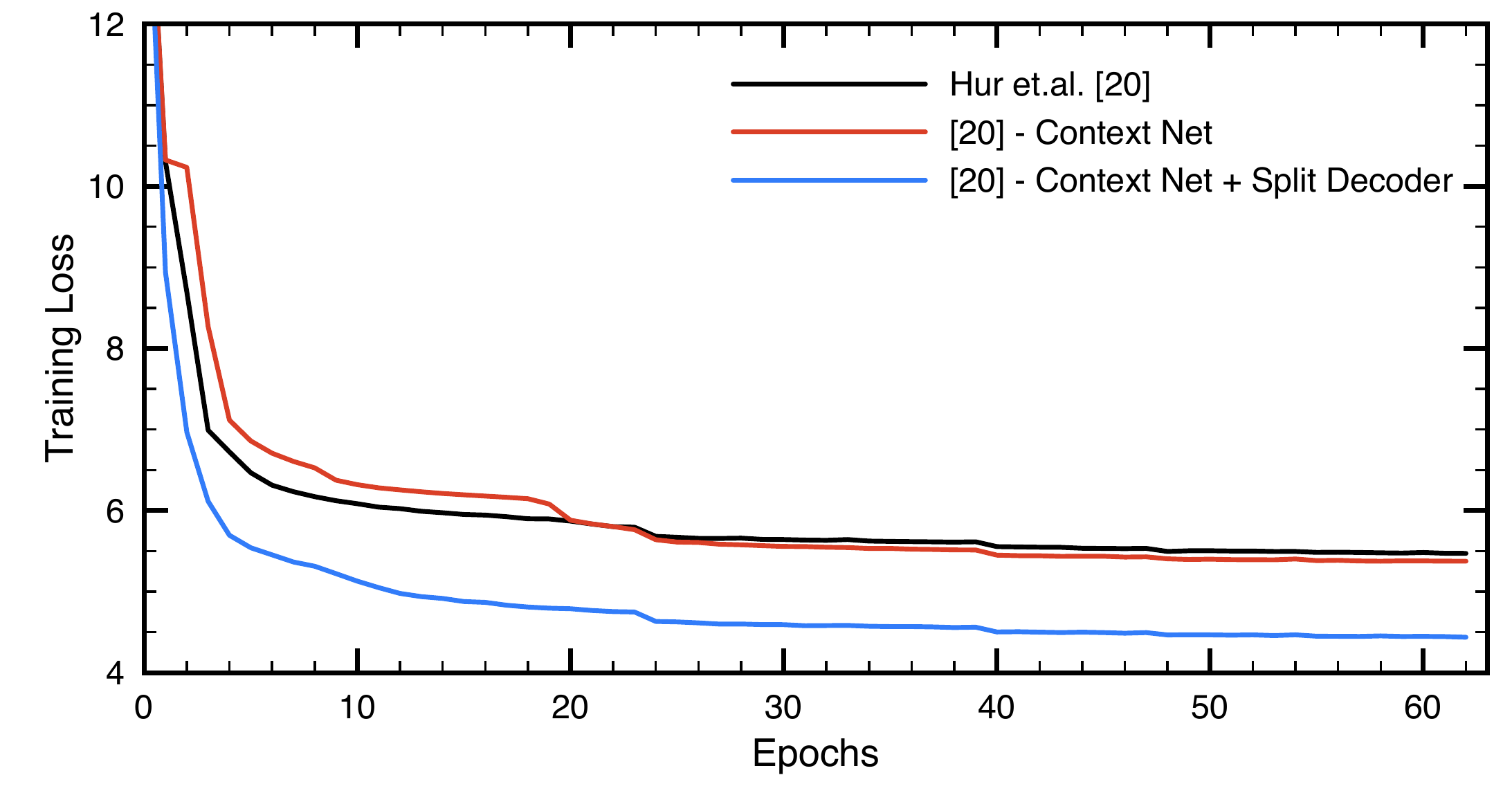}}
\hspace{1cm}
\subcaptionbox{Scene flow outlier rate, our train split \label{supp:fig:accuracy_our}}{\includegraphics[width=0.4\linewidth]{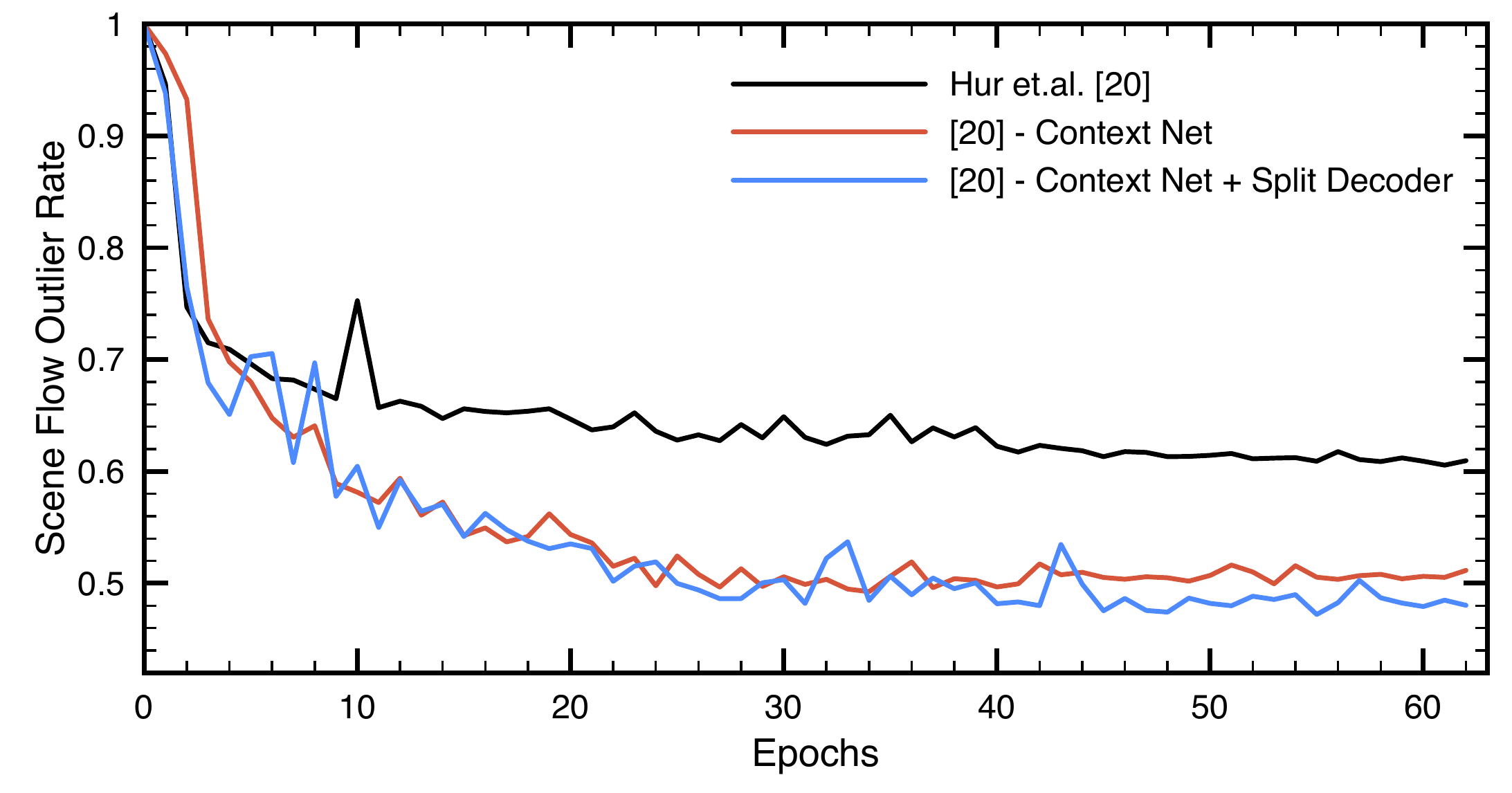}}

\caption{\textbf{An analysis of training stability}: We compare the training loss and the accuracy of three models (\ie, direct baseline \cite{Hur:2020:SSM}, \cite{Hur:2020:SSM} without context network, and \cite{Hur:2020:SSM} without context network and with split decoder), training on two different splits (\ie~the original training split from \cite{Hur:2020:SSM} and our multi-frame train split). Discarding the context network offers more stable, faster convergence of the accuracy on both splits. Further applying the split decoder improves the accuracy.}
\label{supp:fig:training_stability}
\vspace{-0.5em}
\end{figure*}

\Cref{supp:table:empirical_study} provides our empirical study on adopting each key factor on top of our baseline and reports the scene flow accuracy.
We follow the training setup from \cite{Hur:2020:SSM}.
We first apply cost volume normalization (CV.~Norm.) on the model from \cite{Hur:2020:SSM} without the context network (Cont.~Net.).\footnote{We use the model without the context network for more stable training, see main paper.}
Cost volume normalization clearly improves the accuracy on all metrics, up to \num{4.4}\% (relative improvement) in terms of the scene flow accuracy.
We choose this model as the baseline and conduct further empirical study on top of it.

We find the census loss and using one less pyramid level to be beneficial for the monocular scene flow setup as well.
On the other hand, using data distillation or a $640 \times 640$ pixel resolution only brings marginal improvements, but requires a much longer training time; thus we decide not to apply them here.
We observe that using a square resolution improves the optical flow accuracy but hurts the disparity accuracy; this offsets the improvement in the end.
When using level dropout, which randomly skips some pyramid levels when training, training unfortunately did not converge.

In \cref{supp:table:empirical_study_occ}, we also compare different occlusion estimation techniques, specifically disocclusion detection and the forward-backward consistency check as described in \cite{Jonschkowski:2020:WMU}.
We use our final model with multi-frame estimation for the study.
Unlike the conclusion from \cite{Jonschkowski:2020:WMU}, we find that using disocclusion information for occlusion detection produces more accurate scene flow than using a forward-backward consistency check.

\begin{table}[t]
\centering
\footnotesize
\setlength\tabcolsep{2pt}
\begin{tabular*}{\columnwidth}{@{\extracolsep{\fill}}l@{\hskip 1em}S[table-format=2.2]S[table-format=2.2]S[table-format=2.2]S[table-format=2.2]@{}}
	\toprule
	Model & {D$1$-all} & {D$2$-all} & {Fl-all} & {SF-all} \\
	\midrule
	Forward-backward consistency & 27.99 & 30.51 & 20.35 & 40.80 \\
	Using disocclusion & \bfseries 27.33 & \bfseries 30.44 & \bfseries 18.92 & \bfseries 39.82 \\
	\bottomrule
\end{tabular*}
\caption{\textbf{Comparison of different occlusion estimation strategies}: Using disocclusion produces more accurate scene flow estimates than using a forward-backward consistency check.}
\label{supp:table:empirical_study_occ}
\vspace{-0.5em}
\end{table}

\myparagraph{Improved training stability.}
As discussed in \cref{sec:approach:architecture} of the main paper, discarding the context network and splitting the decoder improve the training stability with faster convergence.
\cref{supp:fig:training_stability} plots the training loss and the scene flow outlier rate of the direct baseline \cite{Hur:2020:SSM}, the baseline without the context network, and additionally applying our split decoder design.
We demonstrate the results on using two different train splits, the original split from \cite{Hur:2020:SSM} and our multi-frame train split (see \cref{subsec:implementation_details}).

We make the following main observations:
\begin{enumerate*}[label=(\roman*), font=\itshape]
  \item The direct baseline \cite{Hur:2020:SSM} shows a significant accuracy drop when using our train split.
  \item Discarding the context network resolves the issue and offers more stable, faster convergence regarding the accuracy on both train splits.
  \item After applying our split decoder design, the model further improves the accuracy, showing more stable and faster convergence regarding the training loss on both train splits.
\end{enumerate*}

Note that a lower training loss does not always directly translate to better accuracy, since the model optimizes the self-supervised proxy loss.
For this reason we also plot the scene flow outlier rate (on KITTI Scene Flow Training).

% Discarding the context network offers more stable, faster convergence of the accuracy in the early stage of training.
% Afterwards, our split-decoder design offers a lower validation outlier rate (\num{47.53}\% \vs \num{48.97}\%) found at an earlier epoch (\nth{32} \vs \nth{46}) comparing to the baseline without the context network.
% Near the end of the training schedule, the baseline \cite{Hur:2020:SSM} achieves marginally better accuracy (\num{47.05}\% \vs \num{47.53}\%).
% However, the split decoder model offers the benefit of faster convergence and does not have an issue of the accuracy drop on using our \emph{multi-frame train split} with less diverse scenes (\cf~\cref{table:ablation_baseline} in the main paper).
% \todo{This does not really support the conclusions from the main paper. If your original baseline gets better in the end (after 60 epochs), why not use it?}

% !TEX root = ../arxiv.tex
\section{Self-Supervised Loss}
\label{supp:sec:loss}

We provide further details on the self-supervised proxy loss introduced in \cref{sec:approach:loss} in the main paper.
The weighting constant $\lambda_\text{sf}$ in \cref{eq:total_loss} in the main paper is calculated at every iteration step to make the scene flow loss $L_\text{sf}$ equal to the disparity loss $L_\text{d}$, which previous work \cite{Hur:2020:SSM} empirically found to be better than using a fixed constant.

\myparagraph{Disparity loss.}
Following Godard \etal~\cite{Godard:2019:DIS,Godard:2017:UMD}, we use the right view of a stereo image pair for the guidance of disparity estimation at training time; the second view is not used at test time.
The disparity loss consists of a photometric loss $L_\text{d,ph}$ and a smoothness loss $L_\text{d,sm}$ with regularization constant $\lambda_\text{d,sm}=0.1$,
\begin{subequations}
\begin{align}
L_\text{d} &= L_\text{d,ph} + \lambda_\text{d,sm} L_\text{d,sm}
\label{eq:disp}\\
\shortintertext{with}
L_\text{d,ph} &= \rho_\text{census}\big(\mv{I}_{t}, \mv{\tilde{I}}^{\text{disp}}_{t}, \mv{O}^\text{disp}_t\big).
\label{eq:disp_photometric_diff}
\end{align}
% \begin{equation}
% L_\text{d\_sm} = \frac{1}{N} \sum_\mv{p} \sum_{i \in \{x, y\}} {\big\lvert} \nabla^2_i d^\text{l}_{t}(\mv{p}) {\big\rvert} \cdot e^{-\beta {\lVert} \nabla_i \mv{I}^\text{l}_{t}(\mv{p}) {\rVert}_1 }.
% \label{eq:disp_smooth_loss}
% \end{equation}
\end{subequations}
The photometric loss $L_\text{d,ph}$ in \cref{eq:disp_photometric_diff} penalizes the photometric difference between the left view $\mv{I}_{t}$ and the synthesized left view $\mv{\tilde{I}}^{\text{disp}}_{t}$, obtained from the output disparity $\mv{d}_t$ and the given right view $\mv{I}^\text{r}_{t}$ via backward warping \cite{Jaderberg:2015:STN}.
To calculate the photometric difference, we use our new occlusion-aware census loss $\rho_\text{census}$ in \cref{eq:occ_census} in the main paper.
As in \cite{Hur:2020:SSM}, we obtain the disparity occlusion mask $\mv{O}^\text{disp}_t$ from forward-warping the right disparity map by inputting the right view $\mv{I}^\text{r}_{t}$ into the network.

We use an edge-aware \nth{2}-order smoothness term \cite{Hur:2020:SSM,Jonschkowski:2020:WMU} to define the disparity smoothness $L_\text{d,sm}$ in \cref{eq:disp}.
\begin{equation}
L_\text{d,sm} = \frac{1}{N} \sum_\mv{p} \sum_{i \in \{x, y\}} {\big\lvert} \nabla^2_i \mv{d}_{t}(\mv{p}) {\big\rvert} \cdot e^{-\beta {\lVert} \nabla_i \mv{I}_{t}(\mv{p}) {\rVert}_1 },
\label{eq:disp_smooth_loss}
\end{equation}
with $\beta=150$, divided by the number of pixels $N$ \cite{Jonschkowski:2020:WMU}.

\myparagraph{Scene flow loss.}
The scene flow loss consists of three terms \cite{Hur:2020:SSM}: a photometric loss $L_\text{sf,ph}$, a 3D point reconstruction loss $L_\text{sf,pt}$, and a scene flow smoothness loss $L_\text{sf,sm}$,
\begin{subequations}
\begin{align}
L_\text{sf} &= L_\text{sf,ph} + \lambda_\text{sf,pt} L_\text{sf,pt} + \lambda_\text{sf,sm} L_\text{sf,sm},
\label{eq:sf_loss}\\
\shortintertext{with}
L_\text{sf,ph} &= \rho_\text{census}\big(\mv{I}_{t}, \mv{\tilde{I}}^{\text{sf}}_{t}, \mv{O}^\text{sf}_t\big),
\label{eq:sf_photometric_diff}
\end{align}
% \begin{equation}
% L_\text{sf\_pt} = \frac{\sum_\mv{p} \big(1 - O^\text{l,sf}_t(\mv{p}) \big) \cdot {\big\lVert} \mv{P}'_t - \mv{P}'_{t+1} {\big\rVert}_2 }{\sum_{\mv{q}} \big(1 - O^\text{l,sf}_t(\mv{q})\big)}
% \label{eq:sf_3d_dist}
% \end{equation}
% \begin{equation}
% {L_\text{sf\_sm} = \frac{1}{N} \sum_\mv{p} \sum_{i \in \{x, y\}} {\big\lvert} \nabla^2_i \mv{s}^\text{l}_{\text{fw}}(\mv{p}) {\big\rvert} \cdot e^{-\beta {\lVert} \nabla_i \mv{I}^\text{l}_{t}(\mv{p}) {\rVert}_1 } },
% \label{eq:sf_3d_sm}
% \end{equation}
\end{subequations}
and regularization weights $\lambda_\text{sf,pt}=0.2$, $\lambda_\text{sf,sm}=1000$.

The scene flow photometric loss in \cref{eq:sf_photometric_diff} penalizes the photometric difference between the reference image $\mv{I}_{t}$ and the synthesized reference image $\mv{\tilde{I}}^{\text{sf}}_{t}$, obtained from the camera intrinsics $\mm{K}$, estimated disparity $\mv{d}_t$, and the scene flow $\mv{s}^\text{f}_t$ (\cf Fig.~4a in \cite{Hur:2020:SSM}).
Here we also apply our novel occlusion-aware census transform $\rho_\text{census}$ from \cref{eq:occ_census}.
The scene flow occlusion mask $\mv{O}^\text{sf}_t$ is obtained by the disocclusion from the backward scene flow $\mv{s}^\text{b}_{t+1}$.

The 3D reconstruction loss $L_\text{sf,pt}$ in \cref{eq:sf_loss} penalizes the Euclidean distance between the corresponding 3D points, $\mv{P}'_t$ and $\mv{P}'_{t+1}$, however only for visible pixels:
\begin{subequations}
\begin{equation}
L_\text{sf,pt} = \frac{1}{\sum_{\mv{q}} \mv{O}^\text{sf}_t(\mv{q})} \sum_\mv{p} \frac{\mv{O}^\text{sf}_t(\mv{p}) \cdot {\big\lVert} \mv{P}'_t - \mv{P}'_{t+1} {\big\rVert}_2 }{ {\big\lVert} \mv{P}_t {\big\rVert}_2  }
\label{eq:sf_3d_dist}
\end{equation}
with
\begin{align}
\mv{P}'_t &= {\hat{\mv{d}}}_t(\mv{p}) \cdot \mm{K}^{-1} \mv{p} + \mv{s}^\text{f}_t(\mv{p})\\
\mv{P}'_{t+1} &= {\hat{\mv{d}}}_{t+1}(\mv{p'}) \cdot \mm{K}^{-1} \mv{p'}, \\
\shortintertext{and}
\mv{p'} &= \mm{K} \Big( {\hat{\mv{d}}}_t(\mv{p}) \cdot \mm{K}^{-1} \mv{p} + \mv{s}^\text{f}_t(\mv{p}) \Big),
\end{align}
\end{subequations}
where $\mv{p'}$ is the corresponding pixel of $\mv{p}$ given the scene flow and disparity estimate.
$\hat{\mv{d}}_t$ and $\hat{\mv{d}}_{t+1}$ are the depth maps at time $t$ and $t+1$ respectively.
The depth $\hat{\mv{d}}$ is trivally converted from the disparity estimates given the camera focal length $f_{\text{focal}}$ and the baseline of the stereo rig $b$, specifically $\hat{\mv{d}} = f_\text{focal}\cdot b/\mv{d}$.
Here, we assume that the camera focal length and the stereo baseline is given so that the network outputs disparity (or depth) on a certain, fixed scale.

The loss is normalized by the 3D distance of each point $\mv{P}_{t}$ to the camera to penalize the relative distance to camera.
%prevent possible large values occured by points located far, from dominating the overall loss.

The same edge-aware smoothness loss is applied to 3D scene flow, yielding $L_\text{sf,sm}$ in \cref{eq:sf_loss}, also normalized by its 3D distance to camera:
\begin{equation}
L_\text{sf,sm} = \frac{1}{N} \sum_\mv{p} \frac {\sum_{i \in \{x, y\}} {\big\lvert} \nabla^2_i \mv{s}^\text{f}_{t}(\mv{p}) {\big\rvert} \cdot e^{-\beta {\lVert} \nabla_i \mv{I}_{t}(\mv{p}) {\rVert}_1 } } {  {\big\lVert} \mv{P}_t {\big\rVert}_2  },
\label{eq:sf_3d_sm}
\end{equation}
with $\beta=150$ and $N$ being the number of pixels.

% The 3D reconstruction loss in \cref{eq:sf_3d_dist} penalizes the Euclidean distance between the 3D points of two corresponding pixels, \ie the translated 3D point $\mv{P}'$ of pixel $\mv{p}$ in the reference view and the matched 3D point $\mv{P}'_{t+1}$ in the target view (\cf Fig. (4b) in \cite{Hur:2020:SSM}).
% Likewise, the loss is only applied on non-occluded pixels.

% !TEX root = ../arxiv.tex
\section{Implementation Details}

\begin{figure*}[t!]
\centering
\scriptsize
\setlength\tabcolsep{0.1pt}
\renewcommand{\arraystretch}{0.2}
\begin{tabular}{>{\centering\arraybackslash}m{.33\linewidth} >{\centering\arraybackslash}m{.33\linewidth} >{\centering\arraybackslash}m{.33\linewidth}}
	\includegraphics[width=\linewidth]{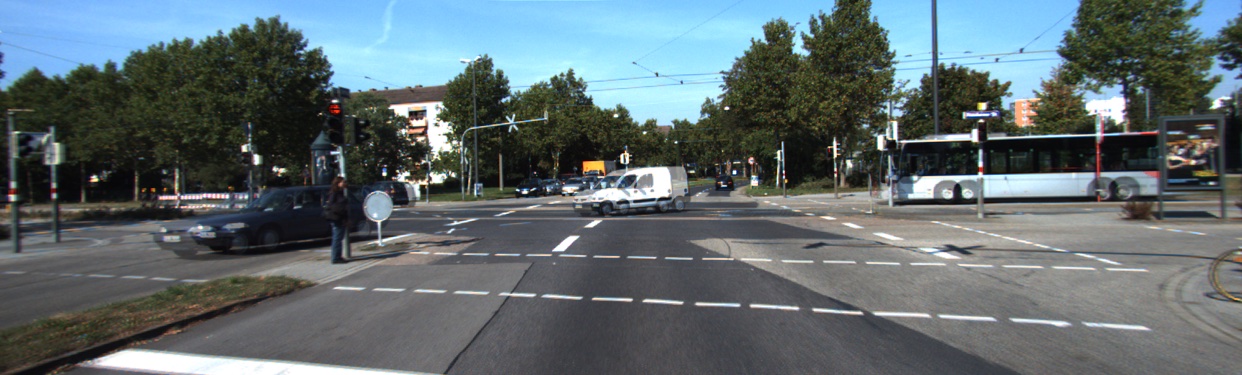} &
	\includegraphics[width=\linewidth]{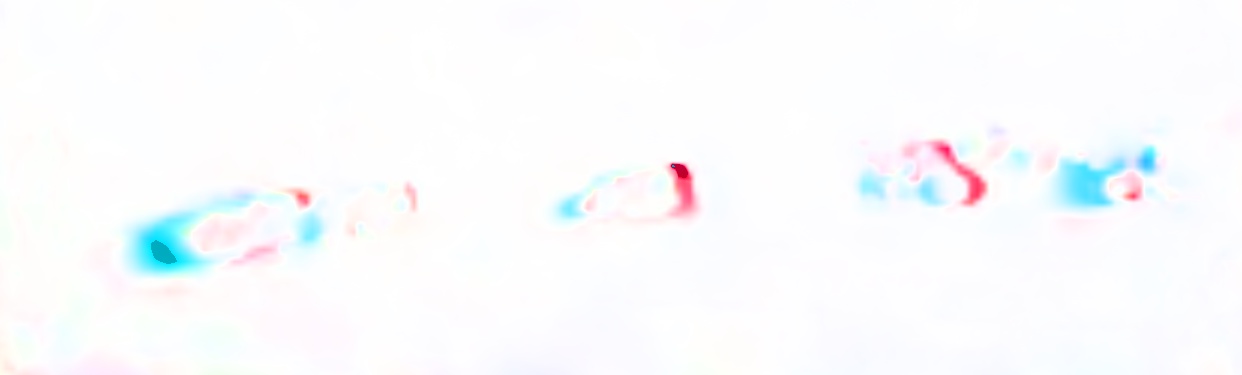} &
	\includegraphics[width=\linewidth]{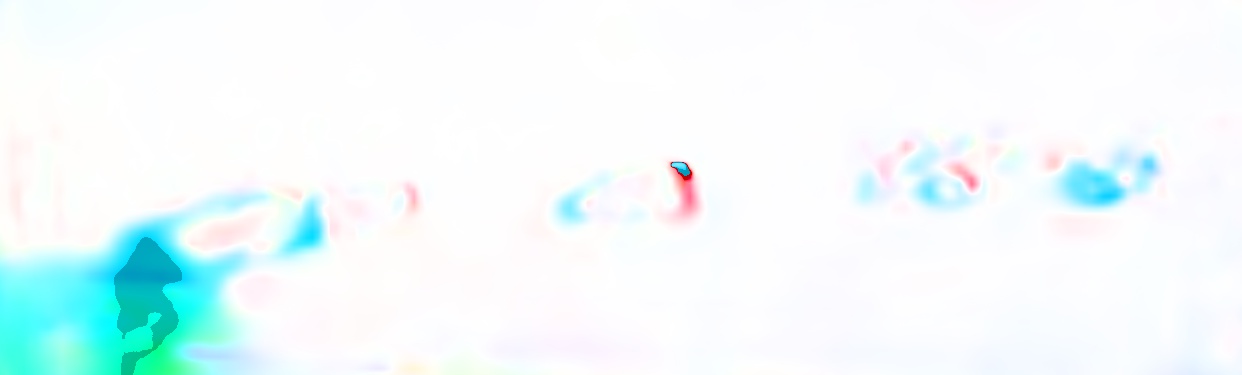} \\
	\includegraphics[width=\linewidth]{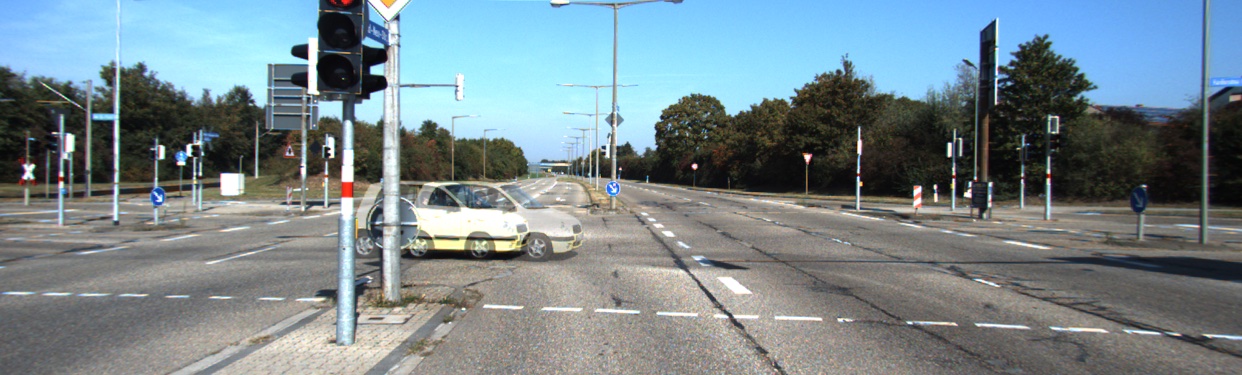} &
	\includegraphics[width=\linewidth]{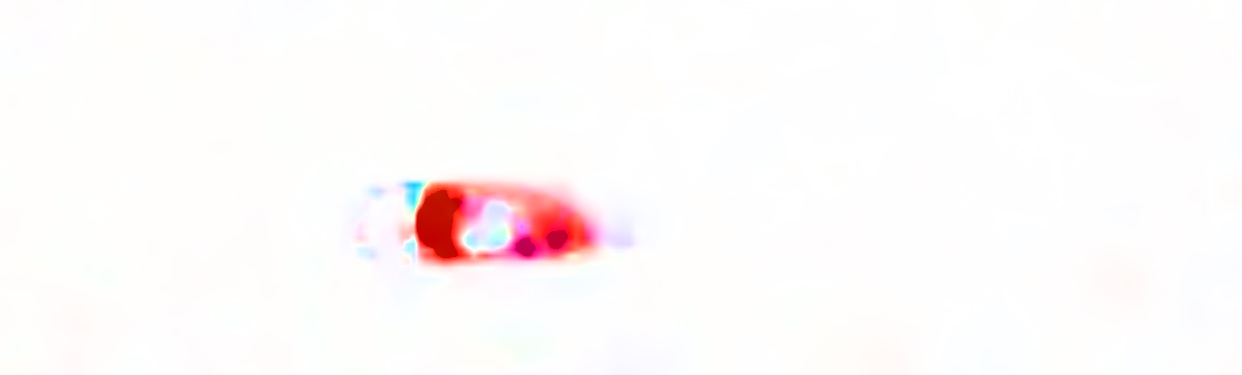} &
	\includegraphics[width=\linewidth]{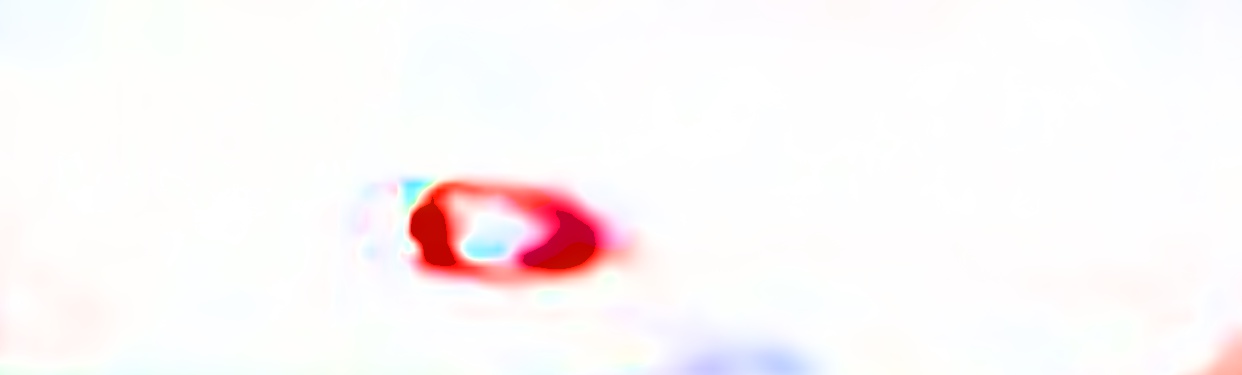} \\
	\includegraphics[width=\linewidth]{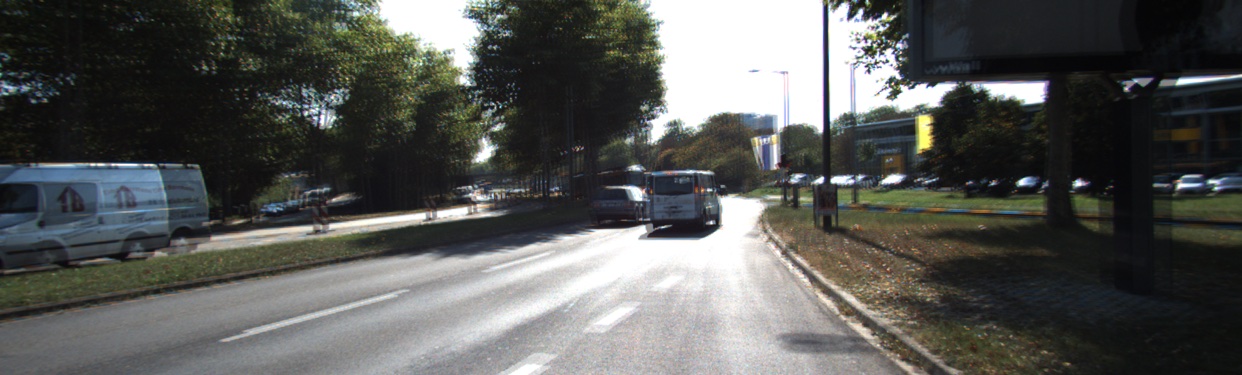} &
	\includegraphics[width=\linewidth]{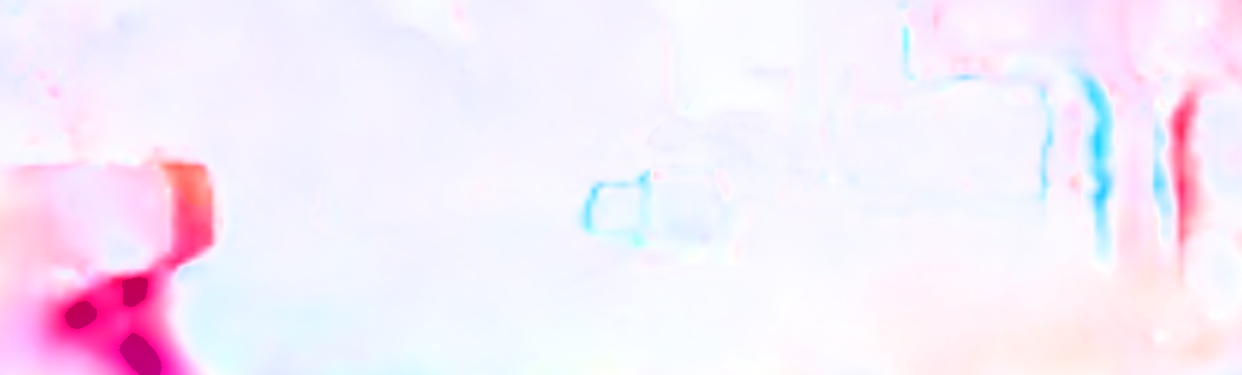} &
	\includegraphics[width=\linewidth]{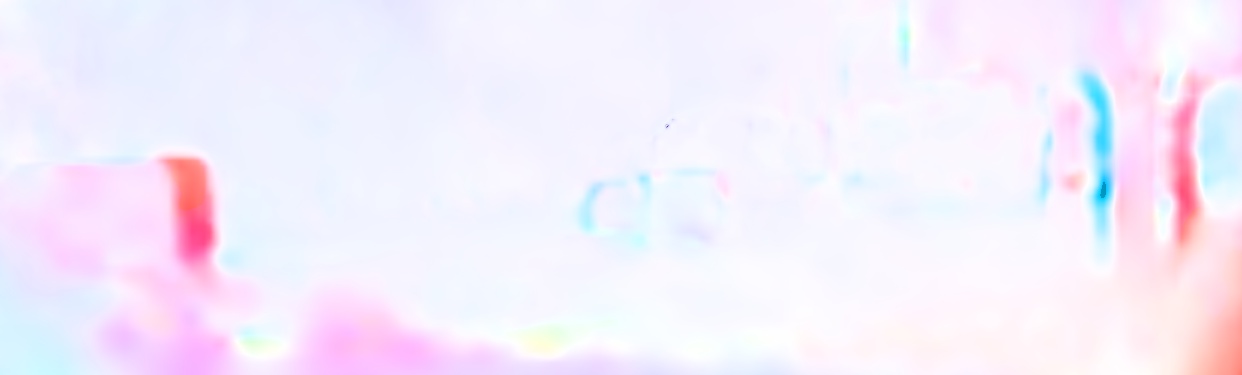} \\ [2em] \\
	(a) Overlayed input images & (b) Ours & (c) Direct baseline \cite{Hur:2020:SSM} \\
\end{tabular}
\caption{\textbf{Qualitative comparison of temporal consistency}: Each scene shows \emph{(a)} overlayed input images and scene flow difference maps of \emph{(b)} our method, and \emph{(c)} the direct two-frame baseline of \cite{Hur:2020:SSM}, visualized using optical flow color coding. Our method provides more temporally consistent estimates near moving objects and out-of-bound regions.}
\label{supp:fig:temporal_consistency}
%\vspace{-0.5em}
\end{figure*}

\begin{table}[b]
\centering
\footnotesize
\vspace{-0.5em}
\begin{tabular*}{\columnwidth}{ @{\extracolsep{\fill}}lc@{}}
	\toprule
	Augmentation type & Sampling range \\
	\midrule
	Random scaling & $[0.93, 1.0]$  \\
	Random cropping & $[-3.5\%, 3.5\%]$ \\
	Gamma adjustment & $[0.8, 1.2]$ \\
	Brightness change multiplication factor & $[0.5, 2.0]$ \\
	Color change multiplication factor & $[0.8, 1.2]$ \\
	\bottomrule
\end{tabular*}
\caption{\textbf{Augmentation parameters}: The augmentation parameters are uniformly sampled from the given sampling ranges.}
\label{supp:table:augmentation}
\end{table}

As briefly discussed in \cref{subsec:implementation_details} of the main paper, we use the augmentation scheme and training configuration suggested by \cite{Hur:2020:SSM}.
The geometric augmentation consists of horizontal flips (with \num{50}\% probability), random scaling, cropping, and resizing into $256 \times 832$ pixels.
Then, a photometric augmentation is applied with \num{50}\% probability, consisting of gamma adjustment, random brightness and color changes.
The augmentation parameters are uniformly sampled from the ranges given in \cref{supp:table:augmentation}.
We use the same augmentation parameters for all consecutive frames included in the same mini-batch.

For training, we use the Adam optimizer \cite{Kingma:2015:AMS} with $\beta_1 = 0.9$ and $\beta_2 = 0.999$.
We do not apply weight decay because we found that it harms the accuracy.
The learning rate schedule from \cite{Hur:2020:SSM} is used.
That is, for self-supervised training for \num{400}k iterations, the initial learning rate starts at $2\times10^{-4}$, being halved at \num{150}k, \num{250}k, \num{300}k, and \num{350}k iteration steps.
Afterwards, for semi-supervised fine-tuning for \num{45}k iterations, the learning rate starts from $4\times10^{-5}$, being halved at \num{10}k, \num{20}k, \num{30}k, \num{35}k, and \num{40}k iteration steps.

% !TEX root = ../arxiv.tex
\section{Computational Cost and Training Time}

Our model takes 153.1\si{\giga} FLOPS of computation per frame pair, with a model size of 7.537\si{\mega} parameters.
It requires only one GPU to train, consumes only 4.89\si{\giga} GPU memory, and trains for 4.5 days (on a single NVIDIA GTX 1080 Ti GPU).

% !TEX root = ../arxiv.tex

\section{Temporal Consistency}

We provide an additional analysis of the temporal consistency, continuing from \cref{subsec:temporal_consistency} in the main paper.
\cref{supp:fig:temporal_consistency} visualizes additional comparisons of the scene flow difference map, comparing to the direct two-frame baseline \cite{Hur:2020:SSM}.
Our model produces visibly more temporally consistent scene flow, especially near moving objects and out-of-bound regions.

In \cref{supp:table:eval_tc}, we also quantitatively evaluate the temporal consistency on \emph{KITTI Scene Flow Training} by calculating the average Euclidean distance of two corresponding scene flow vectors between the two temporally consecutive estimates.
The corresponding scene flow is found using the provided ground truth labels.
While this is not an ideal way for measuring temporal consistency, it shows how much each corresponding scene flow vector changes over time, assuming constant velocity.
Comparing to the direct two-frame baseline \cite{Hur:2020:SSM}, our method gives lower AEPE between two temporally corresponding scene flow vectors, which indicates more temporally consistent estimates.

\begin{table}[]
\centering
\footnotesize
\begin{tabularx}{\columnwidth}{@{}XS[table-format=0.4]S[table-format=0.4]@{}}
	\toprule
	Method & {Self-Mono-SF \cite{Hur:2020:SSM}} & {\bfseries Multi-Mono-SF (ours) } \\
	\midrule
	AEPE & 0.0969 & \bfseries 0.0911 \\
	\bottomrule
\end{tabularx}
\caption{\textbf{Temporal consistency evaluation}: Our method shows lower average end-point error (AEPE) between two temporally consecutive estimates.}
\label{supp:table:eval_tc}
%\vspace{-0.5em}
\end{table}

% !TEX root = ../arxiv.tex
\section{Qualitative Comparison}

\begin{figure*}[t]
\centering
\scriptsize
\setlength\tabcolsep{0.1pt}
\renewcommand{\arraystretch}{0.2}
\begin{tabular}{>{\centering\arraybackslash}m{.20\textwidth} >{\centering\arraybackslash}m{.20\textwidth} >{\centering\arraybackslash}m{.20\textwidth} >{\centering\arraybackslash}m{.20\textwidth} >{\centering\arraybackslash}m{.20\textwidth}}
	\multicolumn{5}{c}{  \small \textcolor{customgray}{$\blacksquare$} Both methods correct \quad \quad \textcolor{customblue}{$\blacksquare$} Ours is correct, \cite{Hur:2020:SSM} is not \quad \quad \textcolor{customred}{$\blacksquare$} \cite{Hur:2020:SSM} is correct, ours is not \quad \quad \textcolor{customyellow}{$\blacksquare$} Both failed}\\ [0.5em]
	\includegraphics[width=\linewidth]{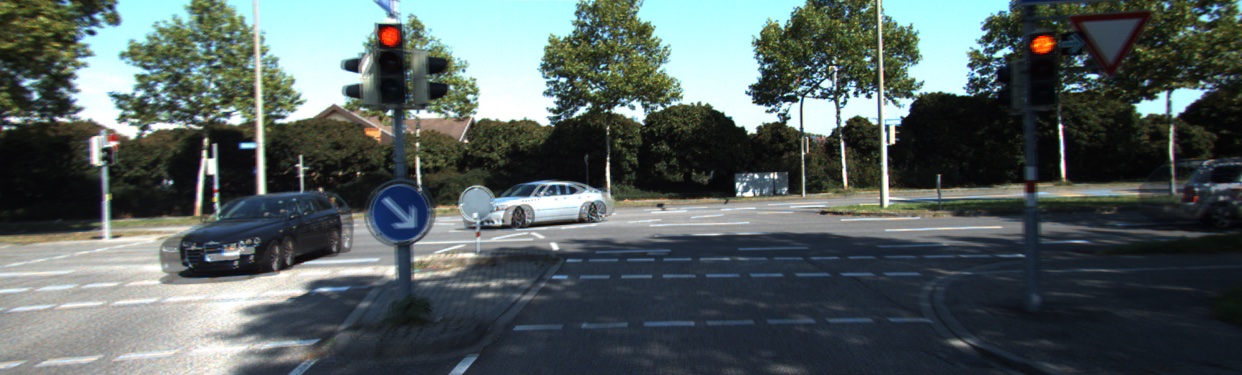} &
	\includegraphics[width=\linewidth]{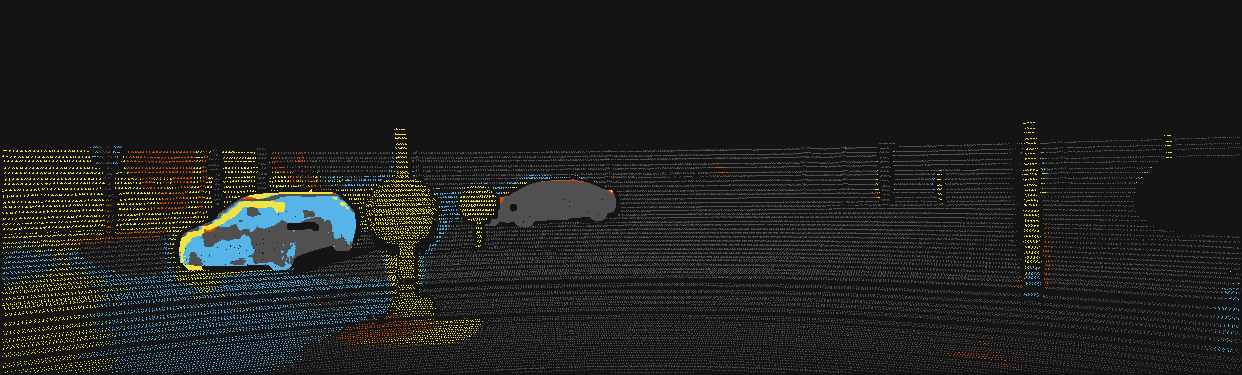} &
	\includegraphics[width=\linewidth]{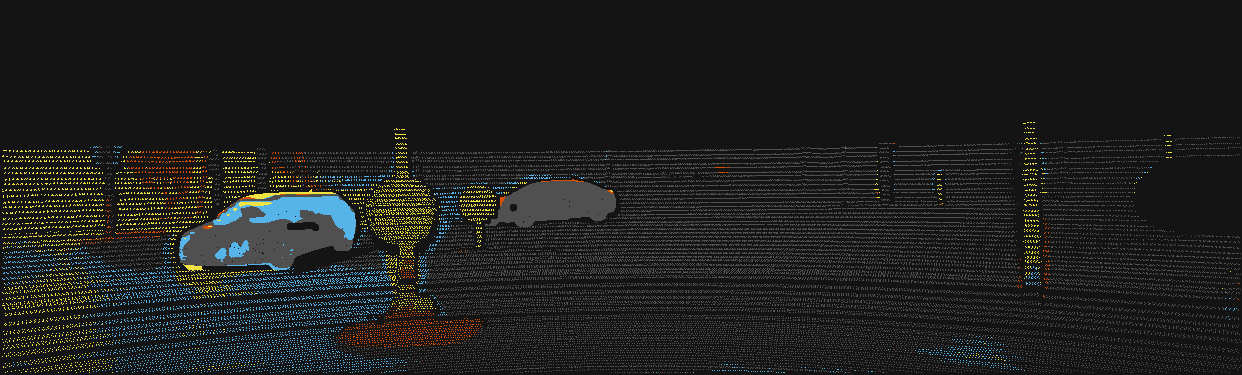} &
	\includegraphics[width=\linewidth]{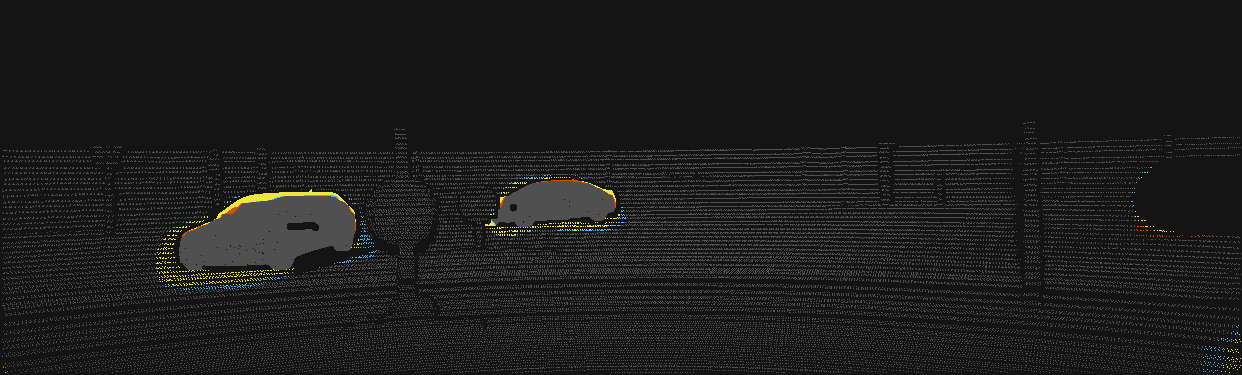} &
	\includegraphics[width=\linewidth]{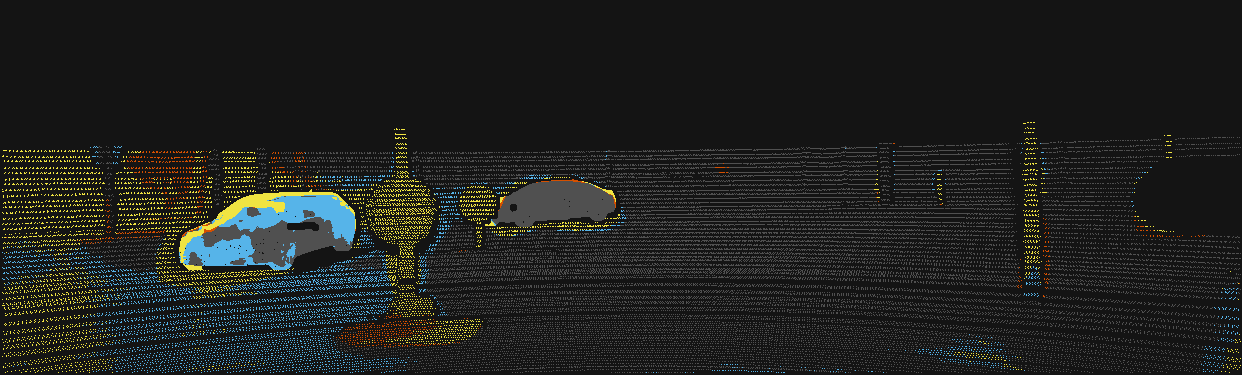} \\
	\includegraphics[width=\linewidth]{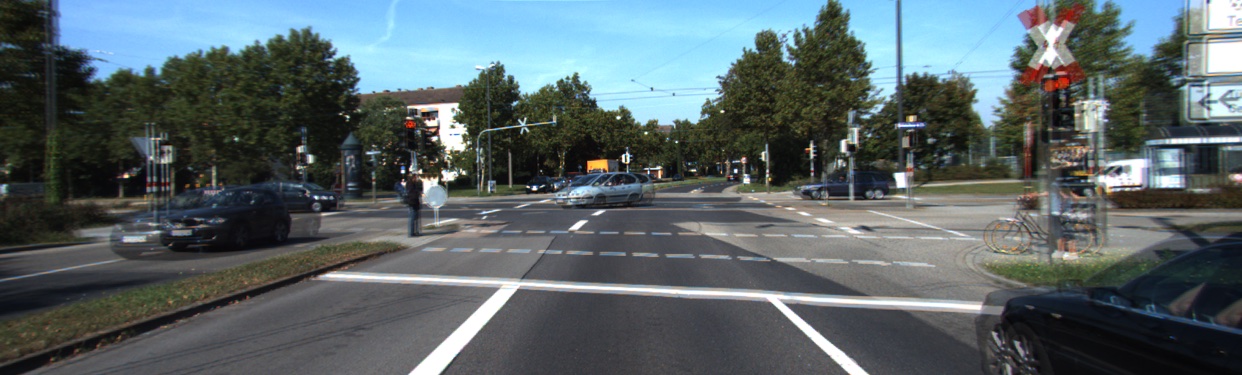} &
	\includegraphics[width=\linewidth]{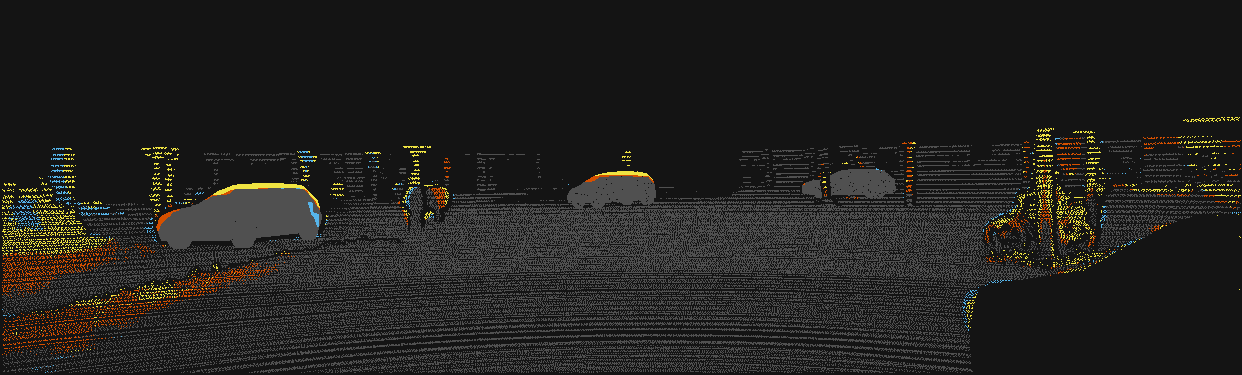} &
	\includegraphics[width=\linewidth]{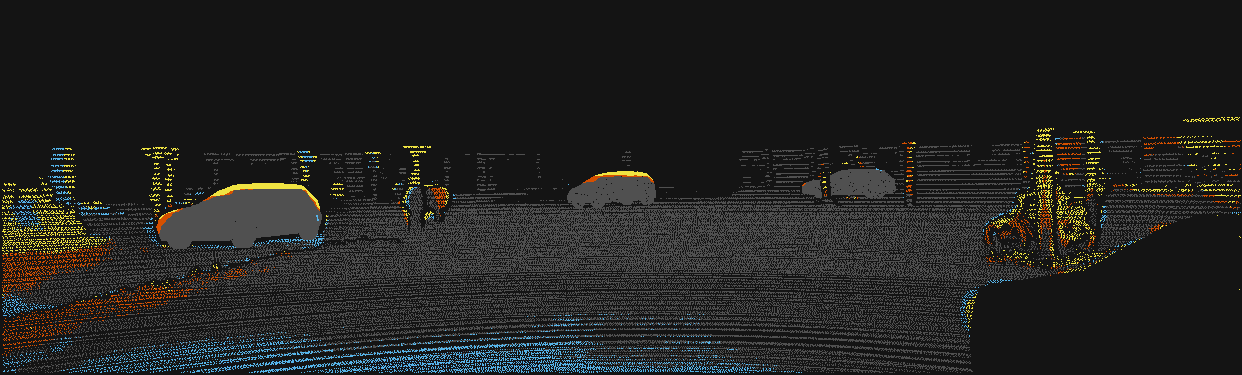} &
	\includegraphics[width=\linewidth]{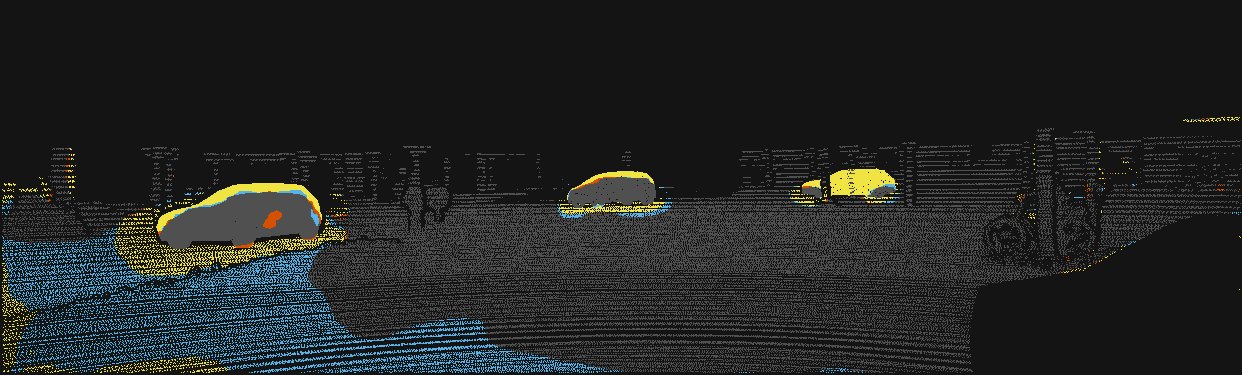} &
	\includegraphics[width=\linewidth]{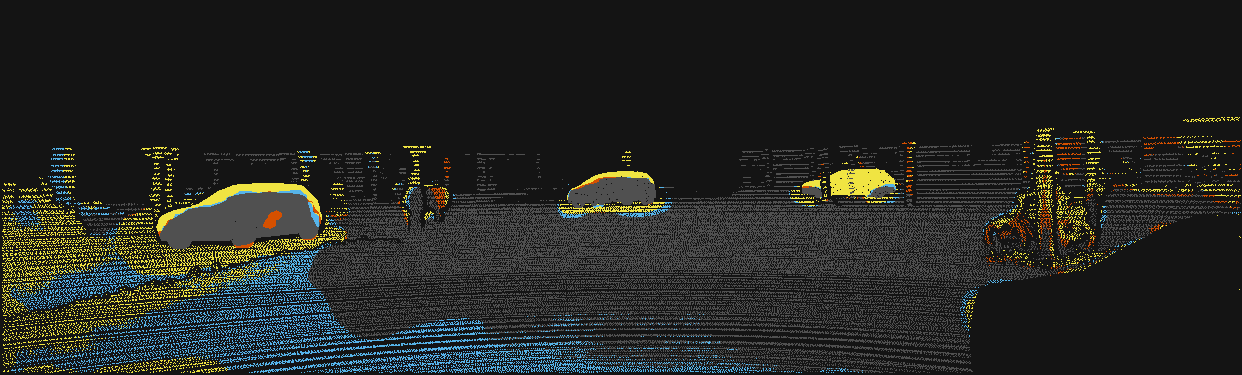} \\
	\includegraphics[width=\linewidth]{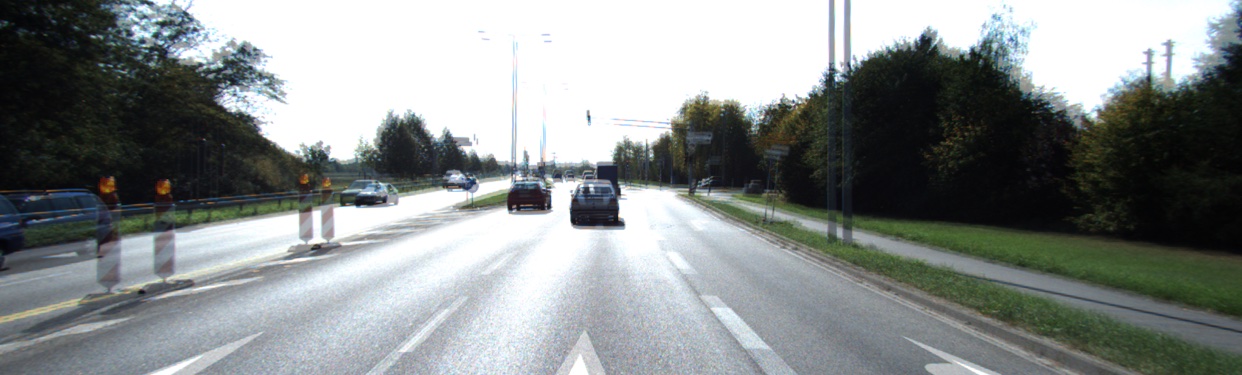} &
	\includegraphics[width=\linewidth]{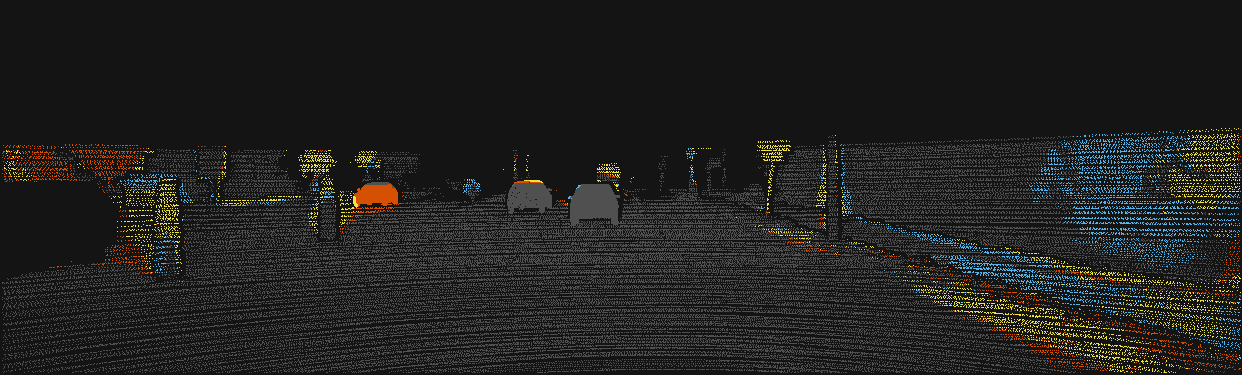} &
	\includegraphics[width=\linewidth]{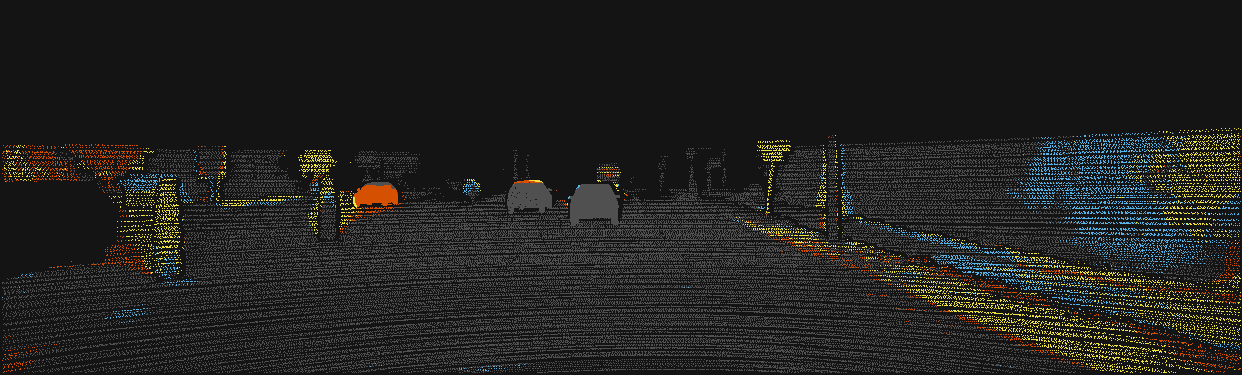} &
	\includegraphics[width=\linewidth]{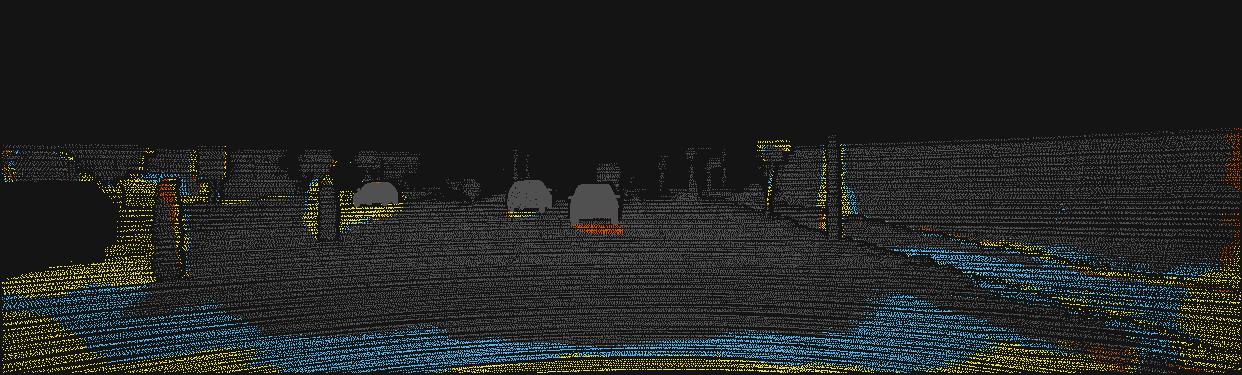} &
	\includegraphics[width=\linewidth]{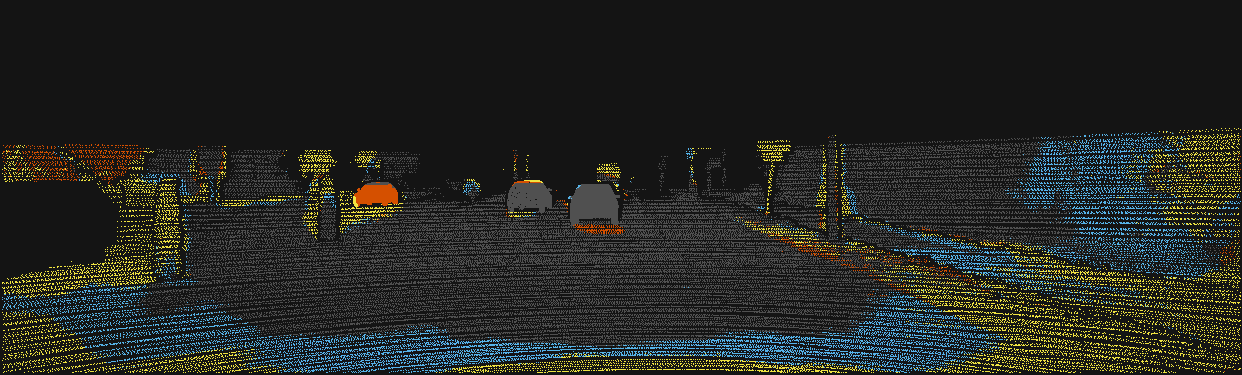} \\
	\includegraphics[width=\linewidth]{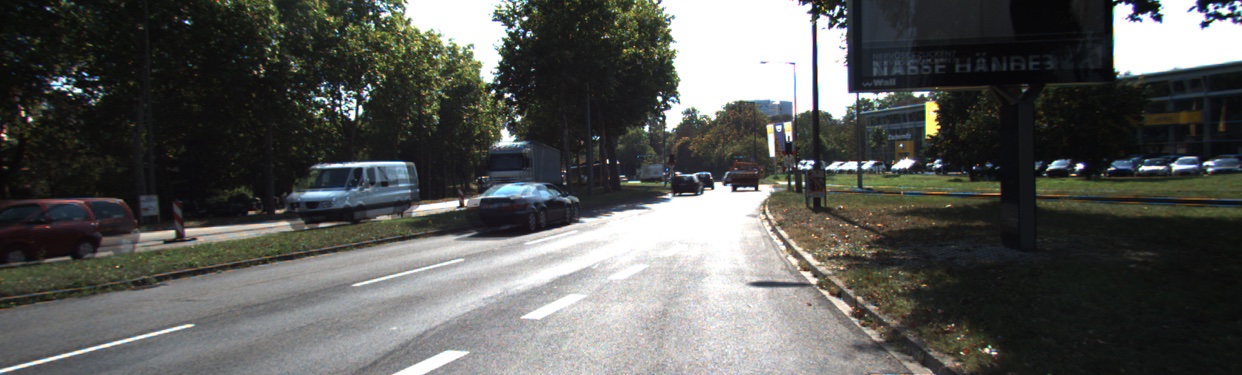} &
	\includegraphics[width=\linewidth]{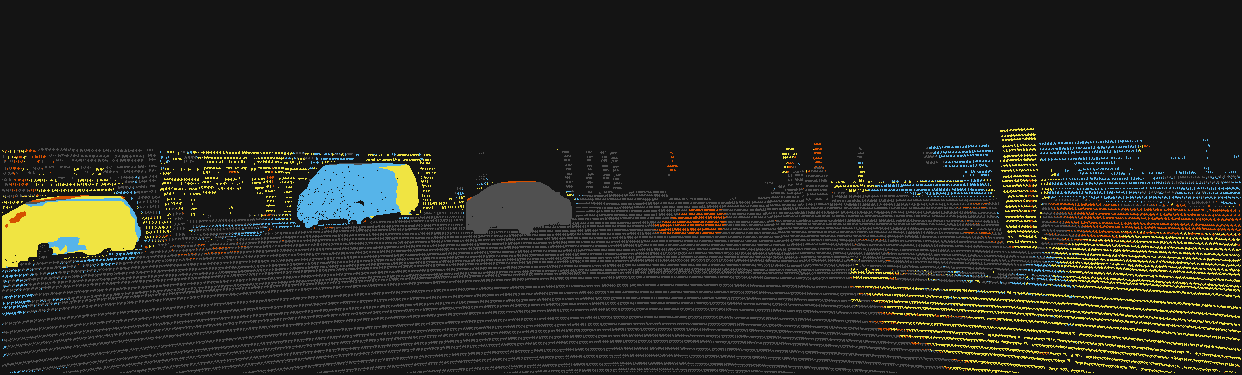} &
	\includegraphics[width=\linewidth]{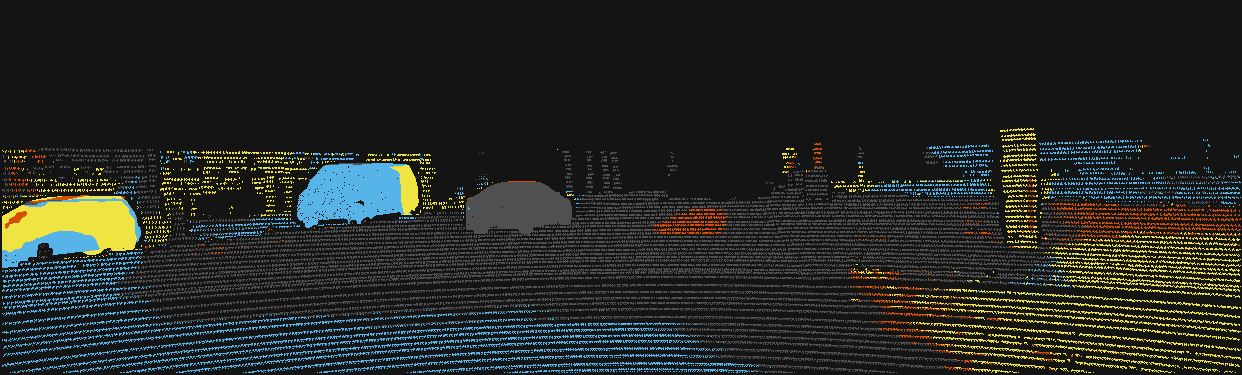} &
	\includegraphics[width=\linewidth]{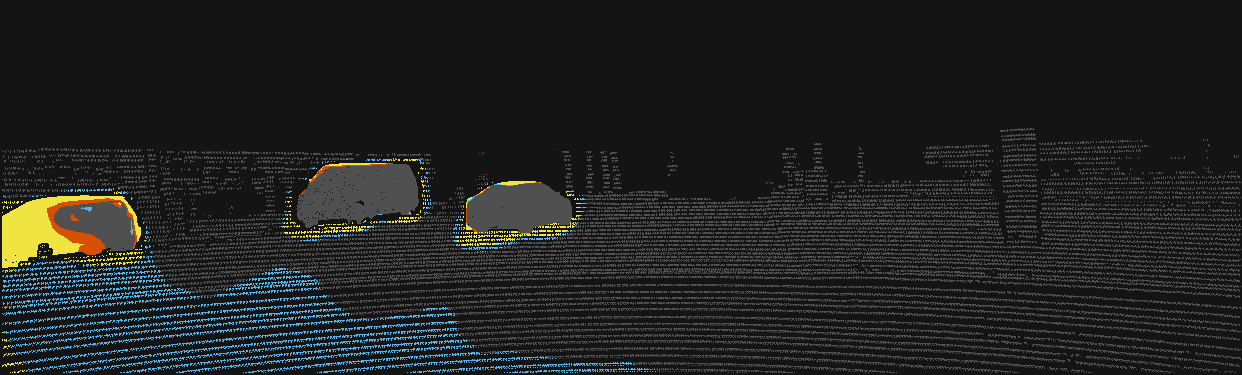} &
	\includegraphics[width=\linewidth]{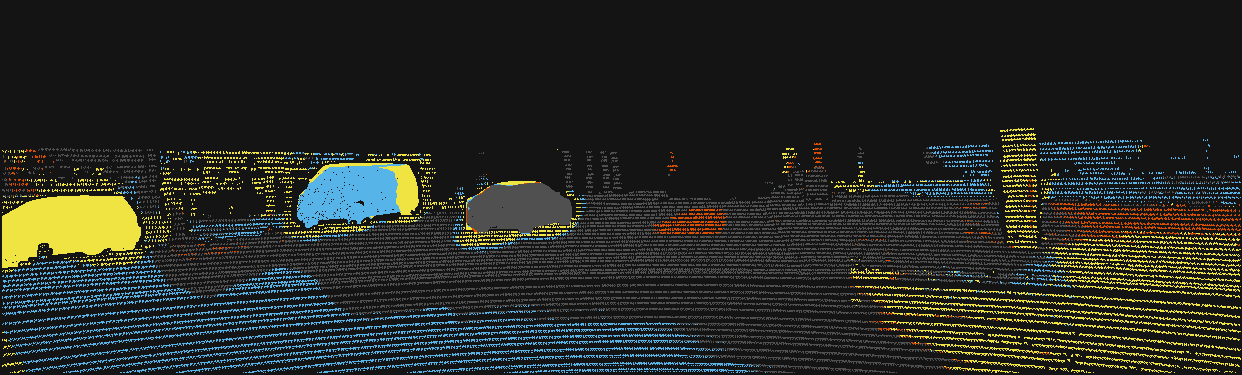} \\
	\includegraphics[width=\linewidth]{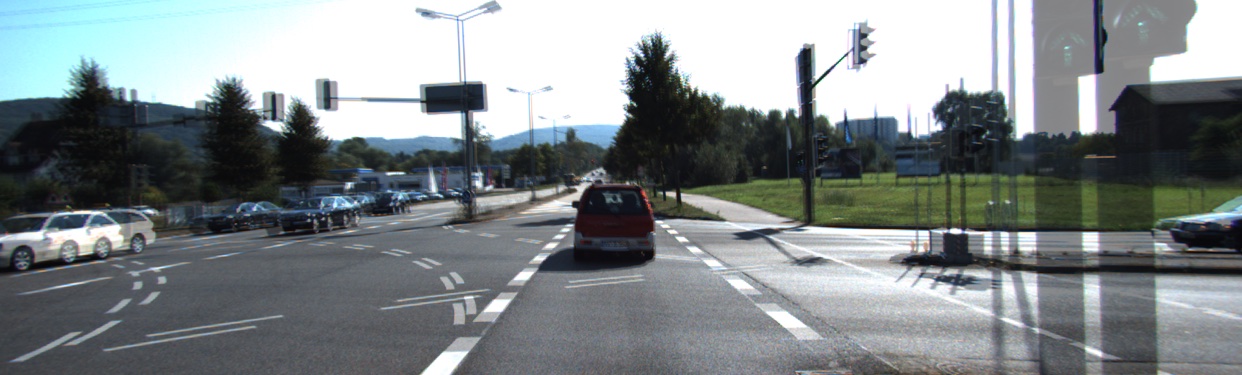} &
	\includegraphics[width=\linewidth]{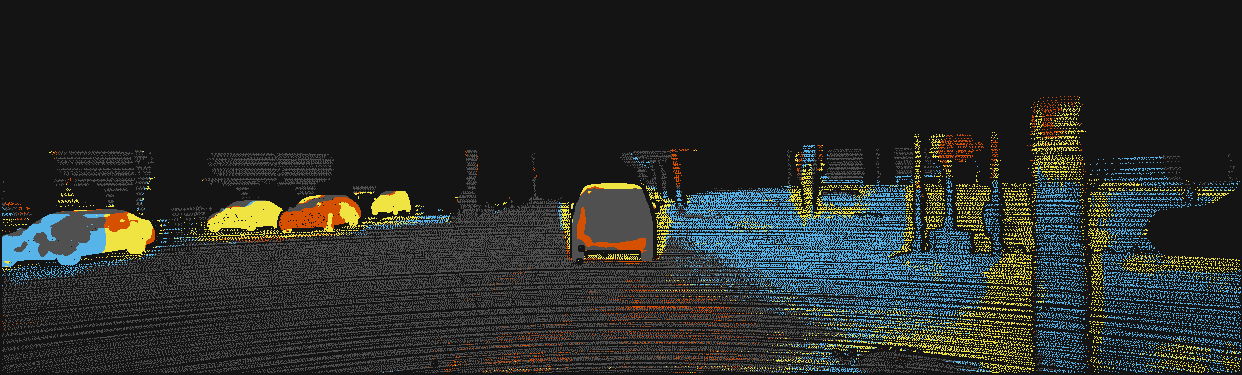} &
	\includegraphics[width=\linewidth]{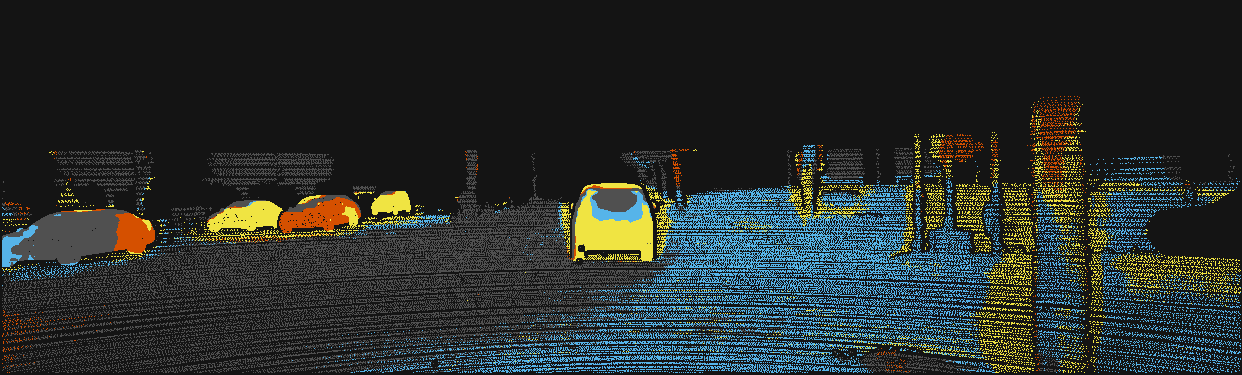} &
	\includegraphics[width=\linewidth]{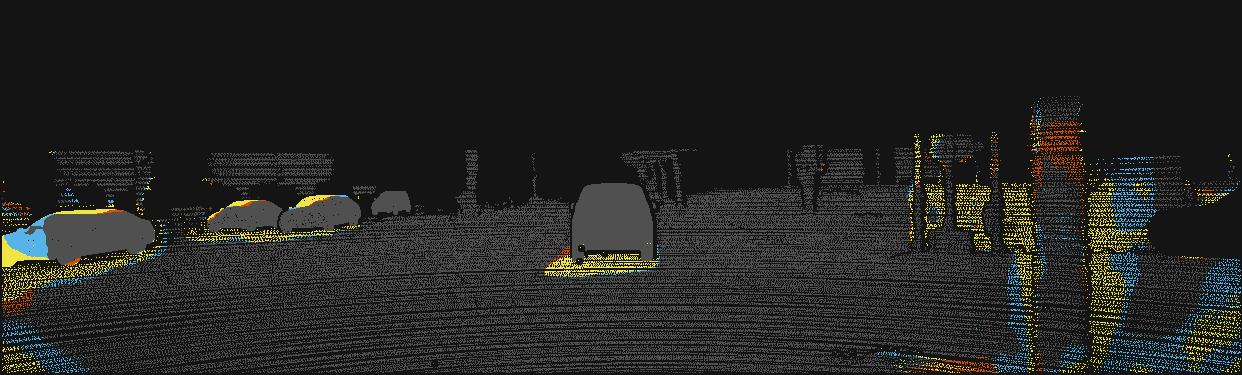} &
	\includegraphics[width=\linewidth]{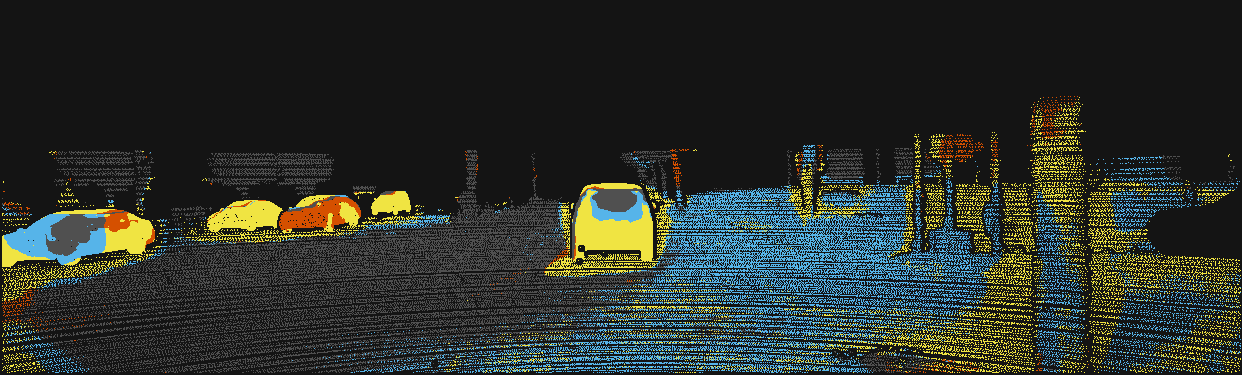} \\
	\includegraphics[width=\linewidth]{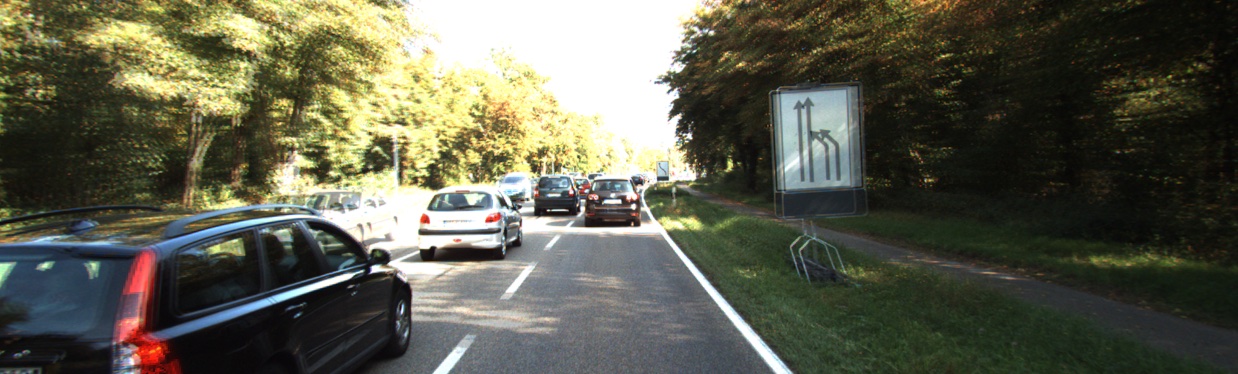} &
	\includegraphics[width=\linewidth]{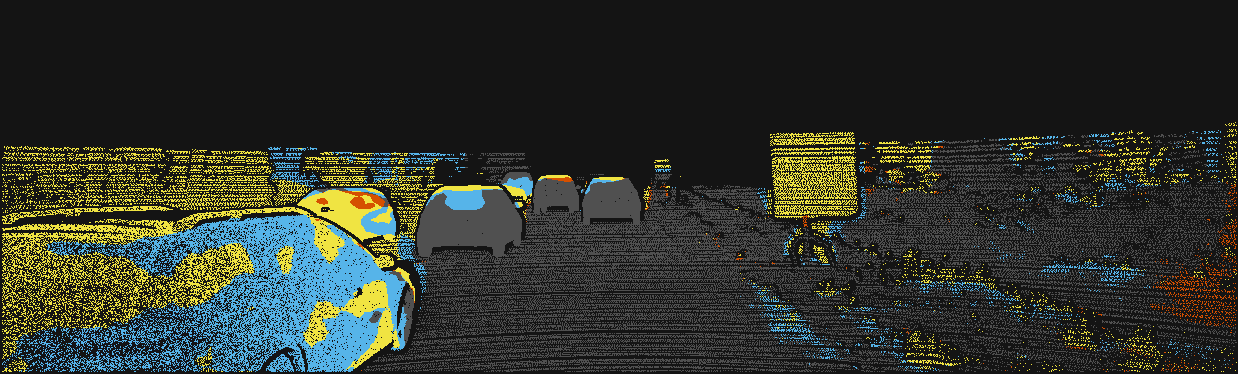} &
	\includegraphics[width=\linewidth]{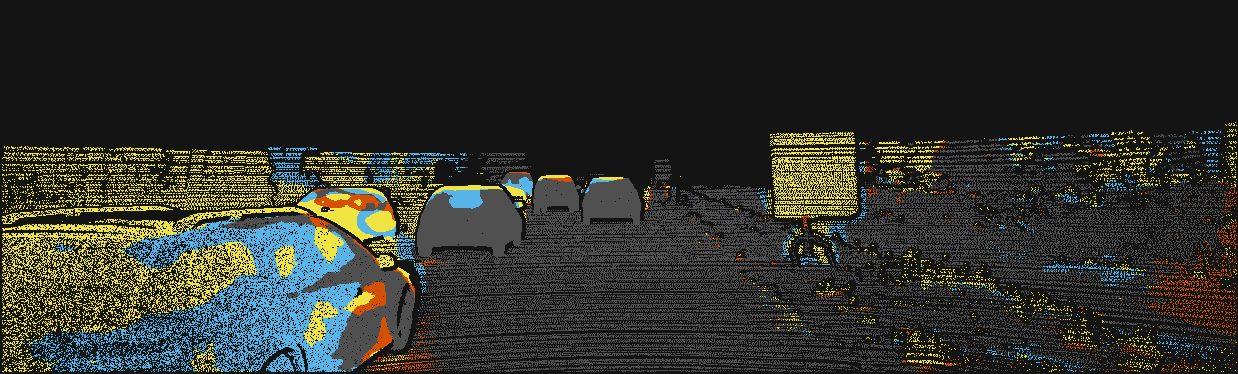} &
	\includegraphics[width=\linewidth]{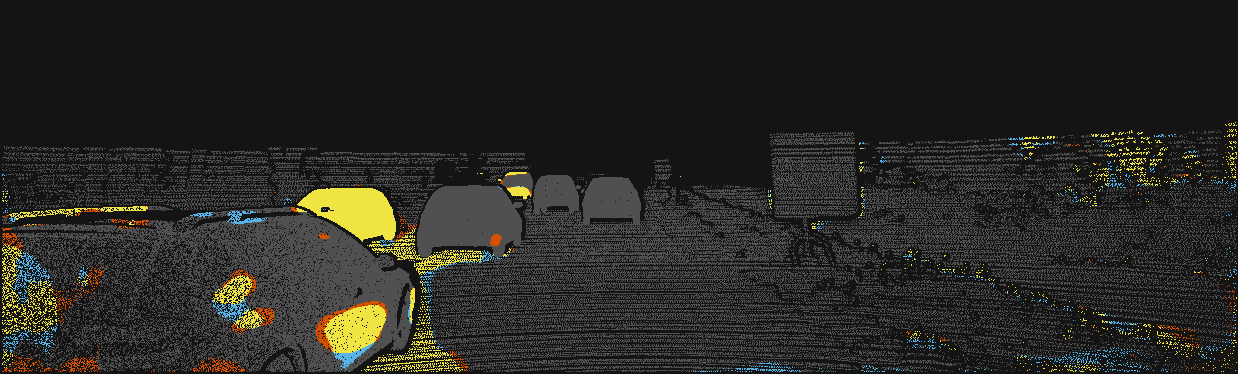} &
	\includegraphics[width=\linewidth]{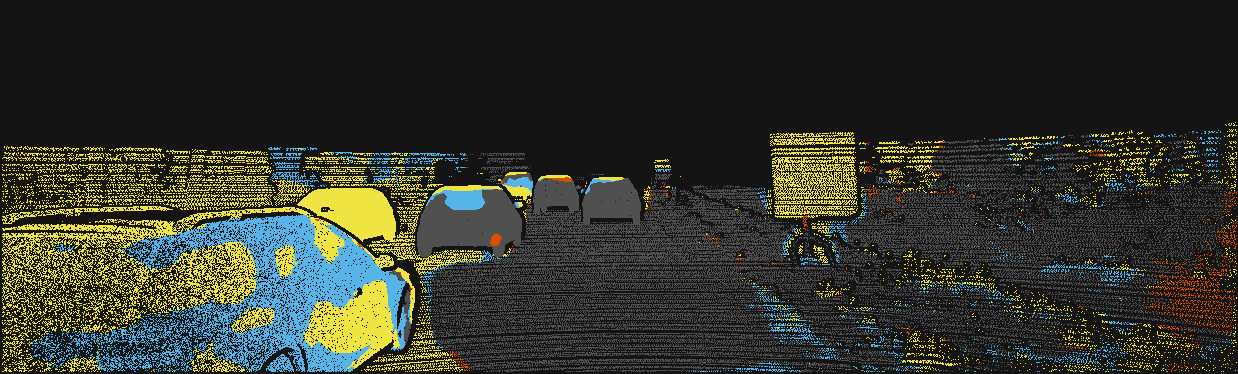} \\
	\includegraphics[width=\linewidth]{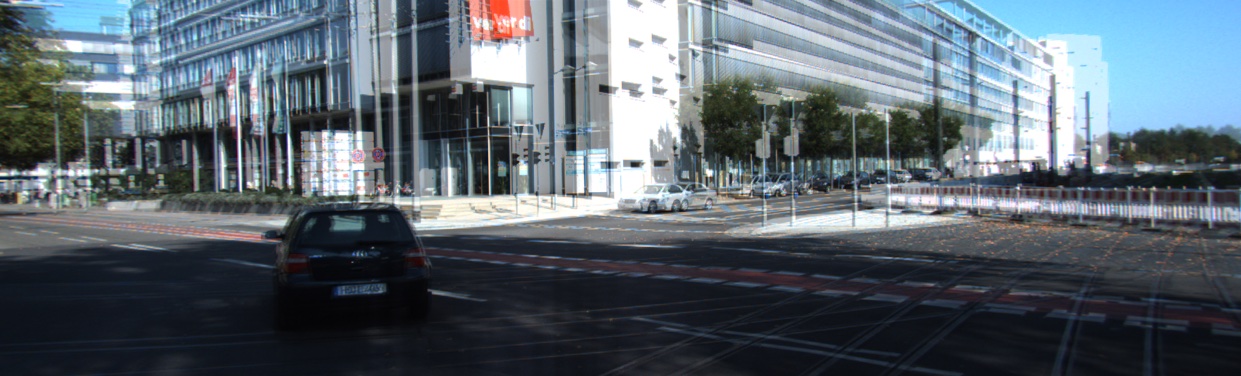} &
	\includegraphics[width=\linewidth]{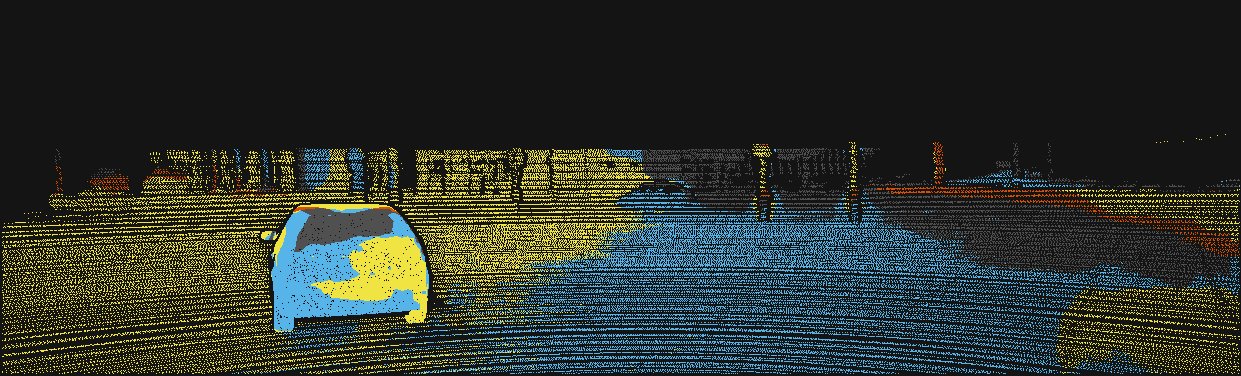} &
	\includegraphics[width=\linewidth]{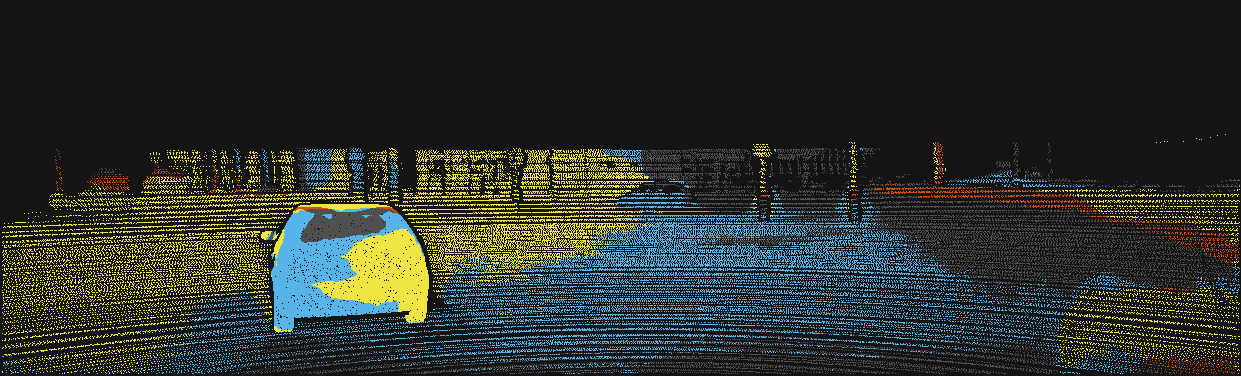} &
	\includegraphics[width=\linewidth]{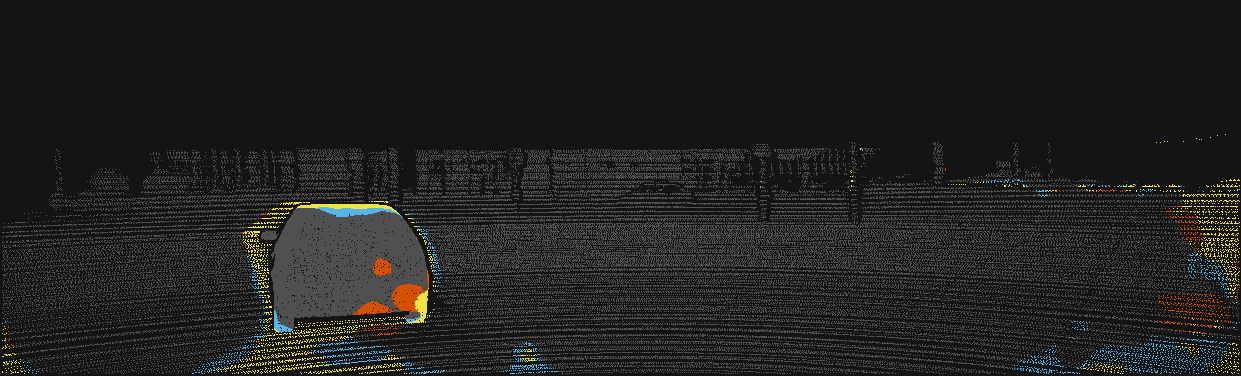} &
	\includegraphics[width=\linewidth]{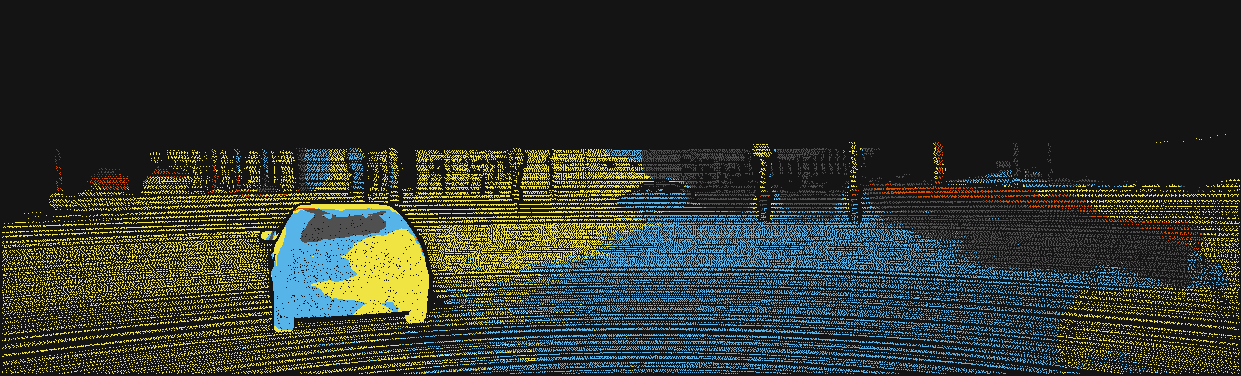} \\
	 [2em] \\
	(a) Overlayed input images & (b) Disparity error map & (c) Disparity change error map & (d) Optical flow error map & (e) Scene flow error map \\
\end{tabular}
\caption{\textbf{Qualitative comparison with the direct baseline of \cite{Hur:2020:SSM}}: Each scene shows \emph{(a)} overlayed input images and error maps for \emph{(b)}~disparity, \emph{(c)} disparity change, \emph{(d)} optical flow, and \emph{(e)} scene flow. Please refer to the color code above the images.}
\label{supp:fig:qualitative_comp}
%\vspace{-0.5em}
\end{figure*}

In \cref{supp:fig:qualitative_comp}, we provide additional qualitative comparisons with the direct two-frame baseline \cite{Hur:2020:SSM} as in \cref{subsec:qual_comp} in the main paper.
Supporting the same conclusion as in the main paper, our approach produces more accurate 3D scene flow on out-of-bound regions, foreground objects, and planar road surfaces.

% !TEX root = ../arxiv.tex
\begin{figure*}[!t]
\centering
\scriptsize
\setlength\tabcolsep{0.1pt}
\renewcommand{\arraystretch}{0.2}
\begin{tabular}{>{\centering\arraybackslash}m{.03\textwidth} >{\centering\arraybackslash}m{.158\textwidth} >{\centering\arraybackslash}m{.158\textwidth} >{\centering\arraybackslash}m{.158\textwidth}
				>{\centering\arraybackslash}m{.015\textwidth} >{\centering\arraybackslash}m{.158\textwidth} >{\centering\arraybackslash}m{.158\textwidth} >{\centering\arraybackslash}m{.158\textwidth}}

	& \multicolumn{3}{c}{\small \textbf{Successful cases}} & & \multicolumn{3}{c}{\small  \textbf{Failure cases}} \\ [0.5em] \\
	\multirow{3}{*}{\small  \rotatebox[origin=c]{90}{nuScenes \cite{Caesar:2020:NSA} }}
	& \includegraphics[width=\linewidth]{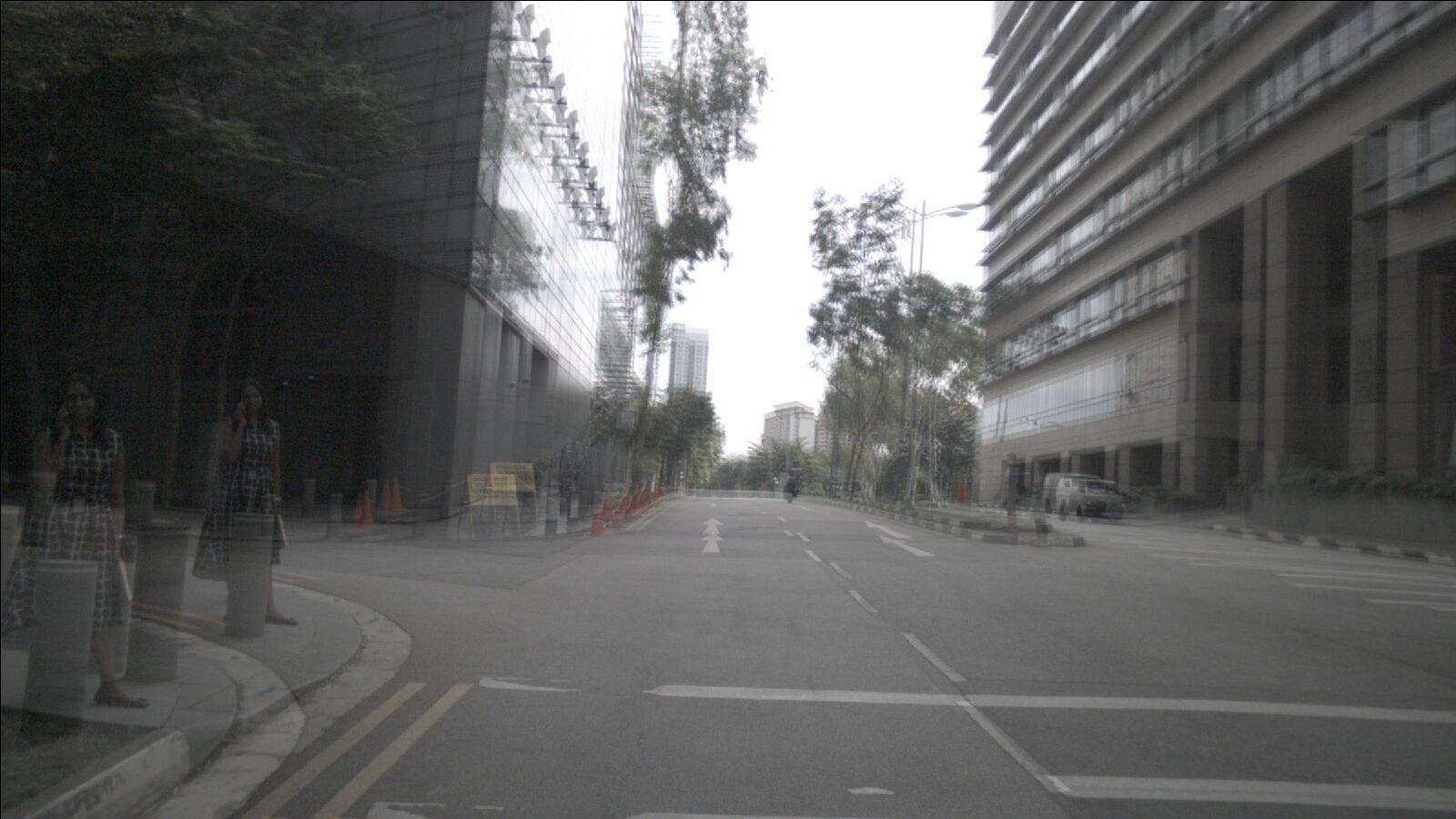} &
	\includegraphics[width=\linewidth]{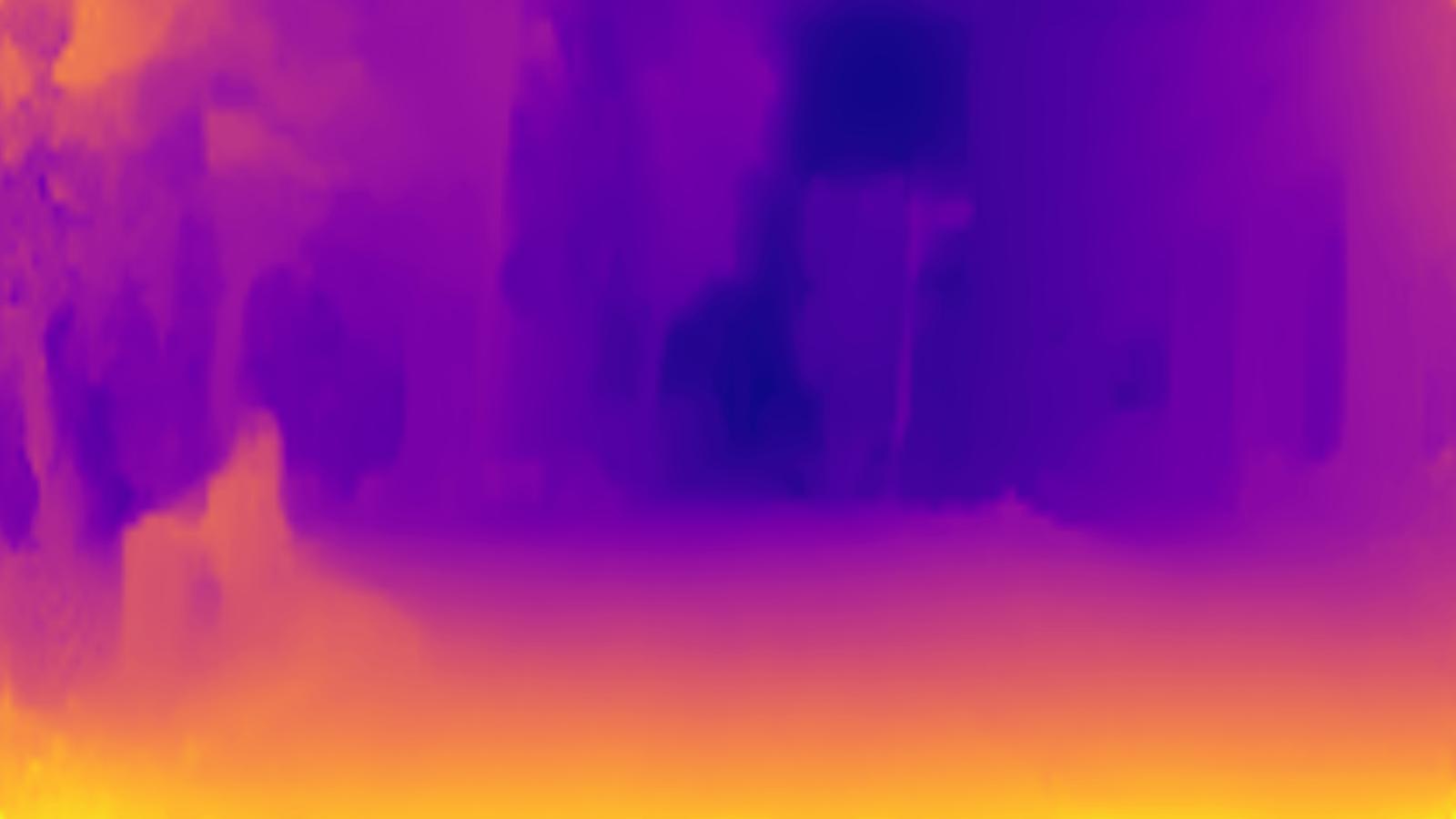} &
	\includegraphics[width=0.95\linewidth]{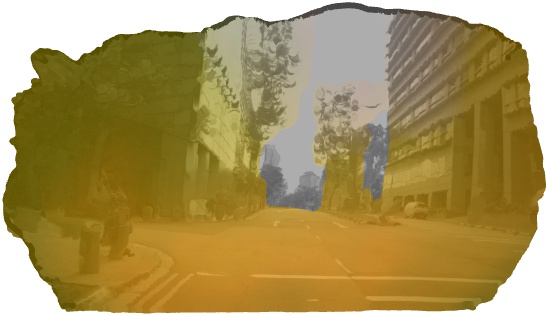} &
	& \includegraphics[width=\linewidth]{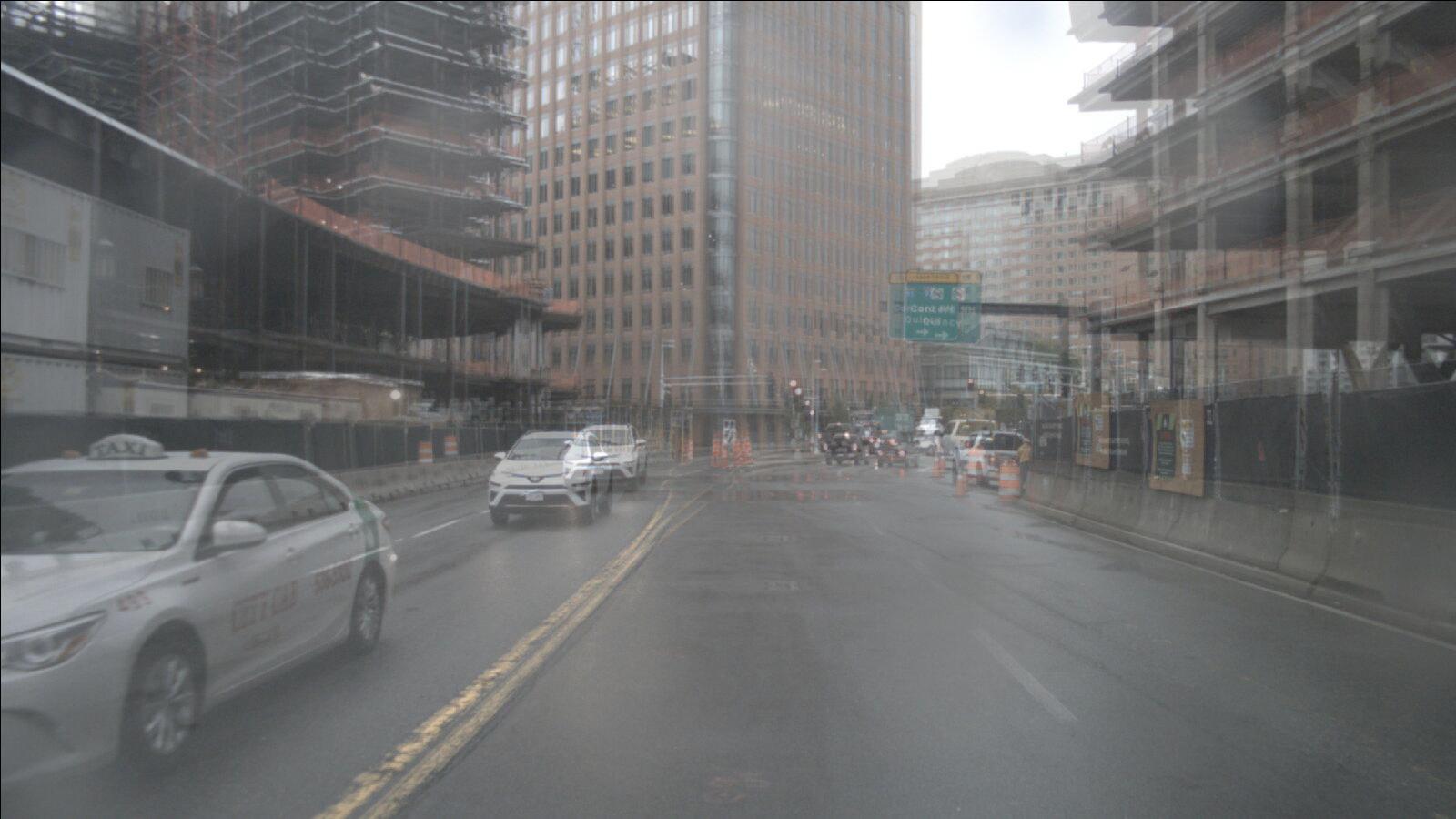} &
	\includegraphics[width=\linewidth]{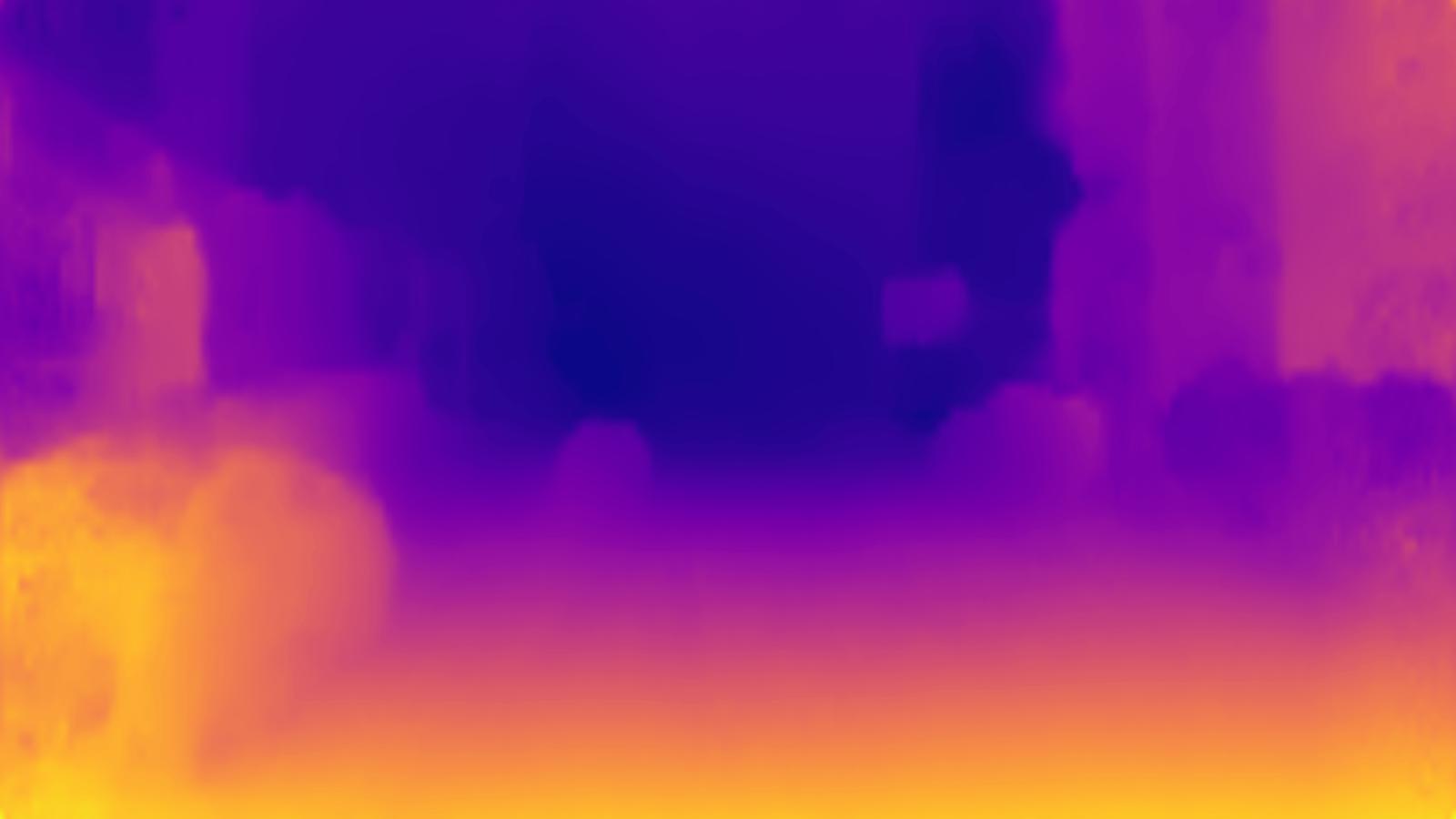} &
	\includegraphics[width=0.95\linewidth]{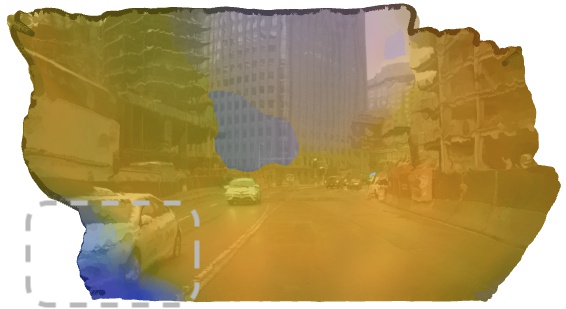} \\

	& \includegraphics[width=\linewidth]{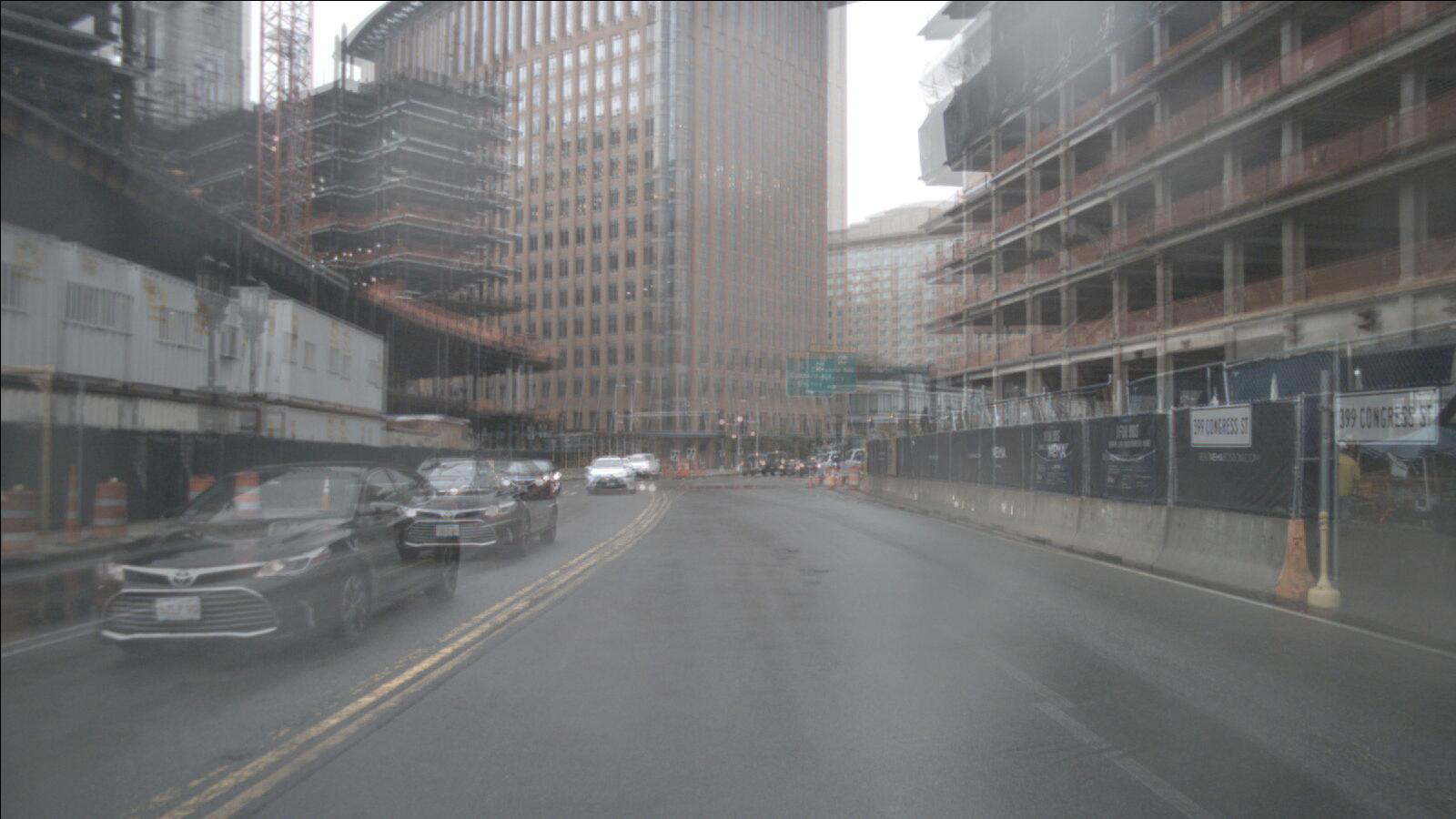} &
	\includegraphics[width=\linewidth]{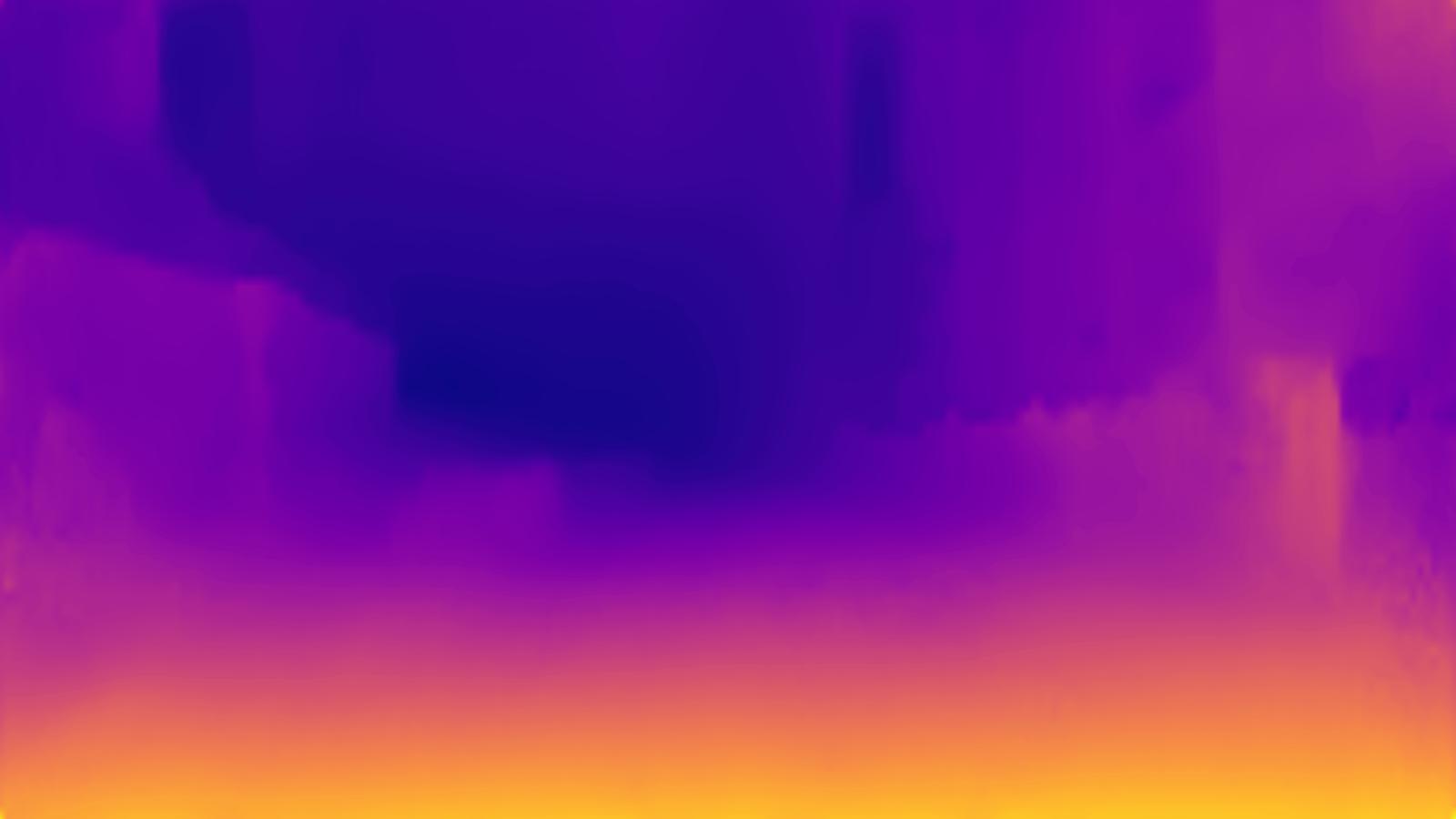} &
	\includegraphics[width=0.95\linewidth]{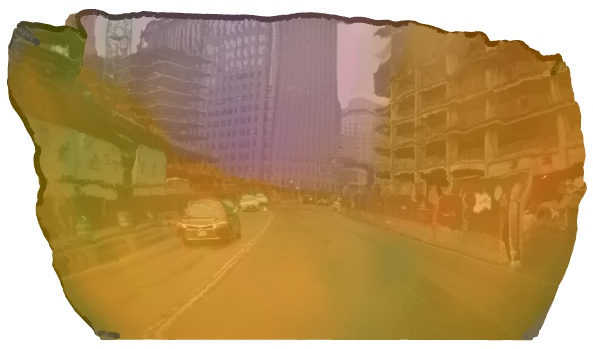} &
	& \includegraphics[width=\linewidth]{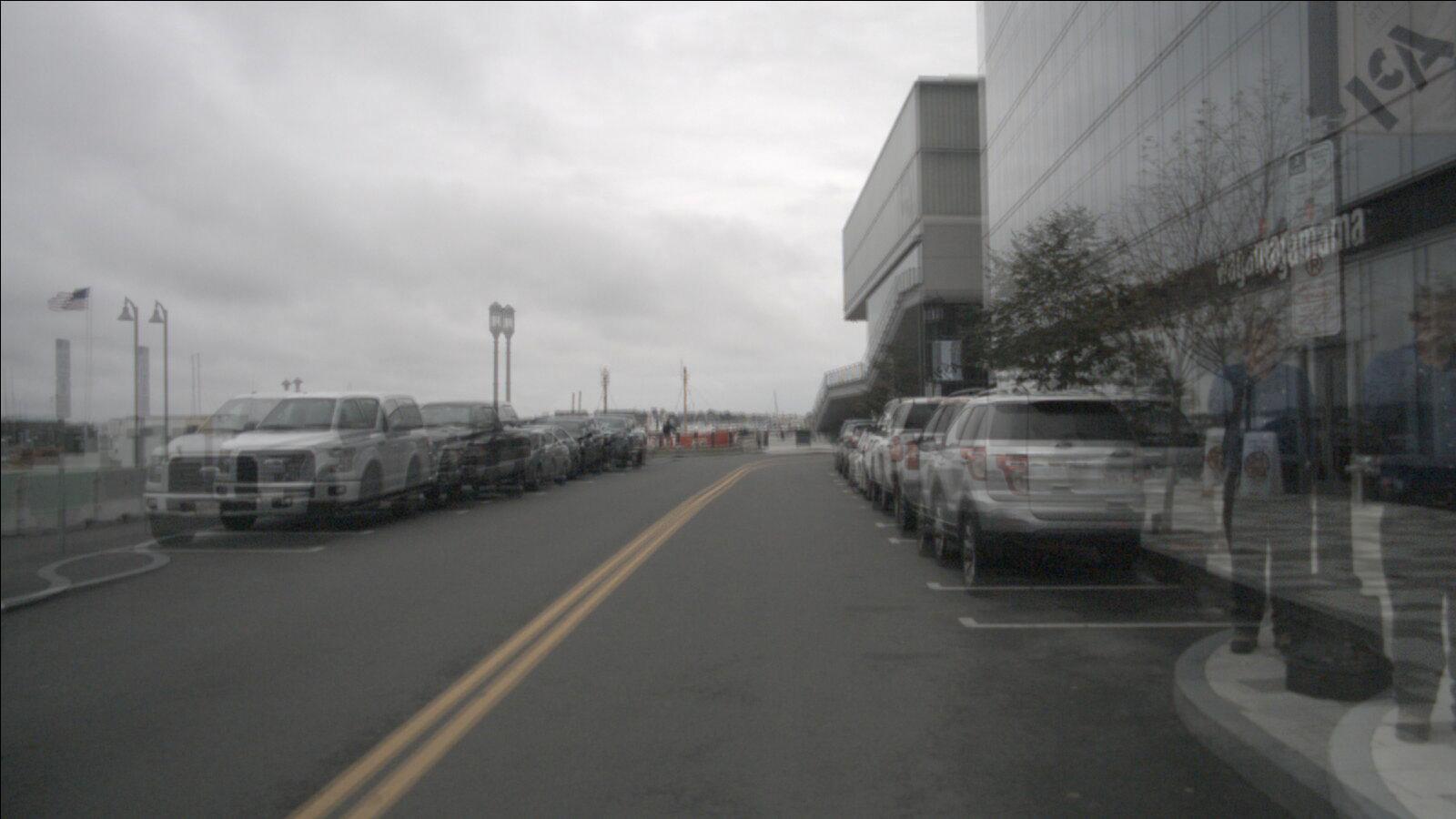} &
	\includegraphics[width=\linewidth]{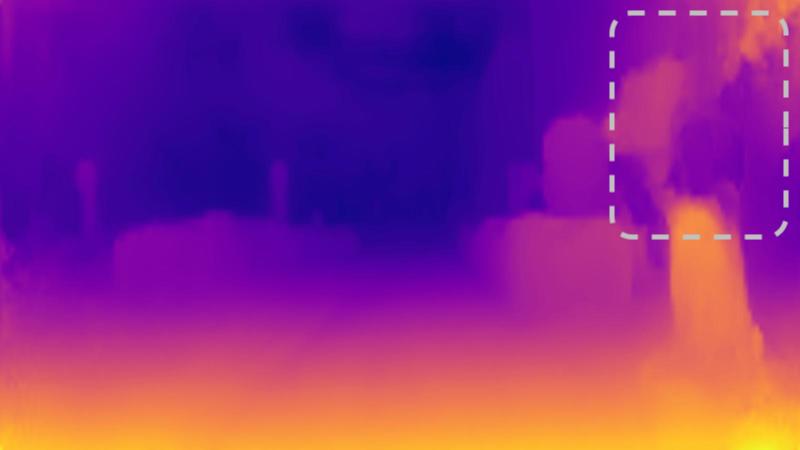} &
	\includegraphics[width=0.95\linewidth]{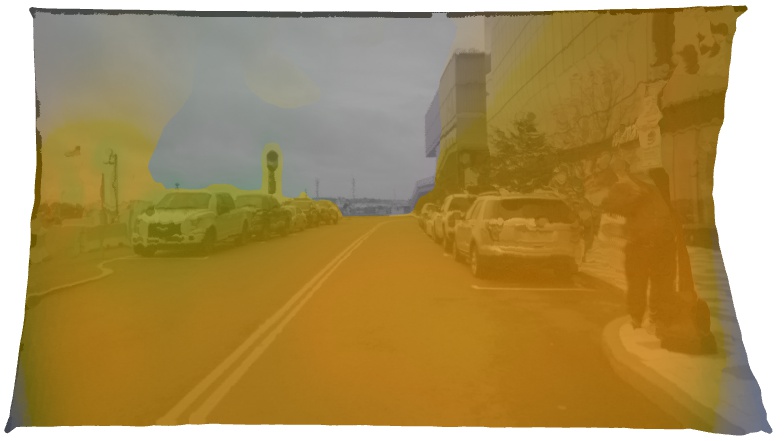} \\ [3.5em] \\

	% & \includegraphics[width=\linewidth]{figures/rebuttal/nuscenes/nu20_overlay.jpg} &
	% \includegraphics[width=\linewidth]{figures/rebuttal/nuscenes/nu20_disp.jpg} &
	% \includegraphics[width=0.95\linewidth]{figures/rebuttal/nuscenes/nu20_vis.jpg} &
	% & \includegraphics[width=\linewidth]{figures/supp_generalization/nuscenes/nu40_overlay.jpg} &
	% \includegraphics[width=\linewidth]{figures/supp_generalization/nuscenes/nu40_disp.jpg} &
	% \includegraphics[width=0.95\linewidth]{figures/supp_generalization/nuscenes/nu40_vis.jpg} \\ [3.5em] \\

	\multirow{3}{*}{\small  \rotatebox[origin=c]{90}{DAVIS \cite{Perazzi:2016:ABD} }}
	& \includegraphics[width=\linewidth]{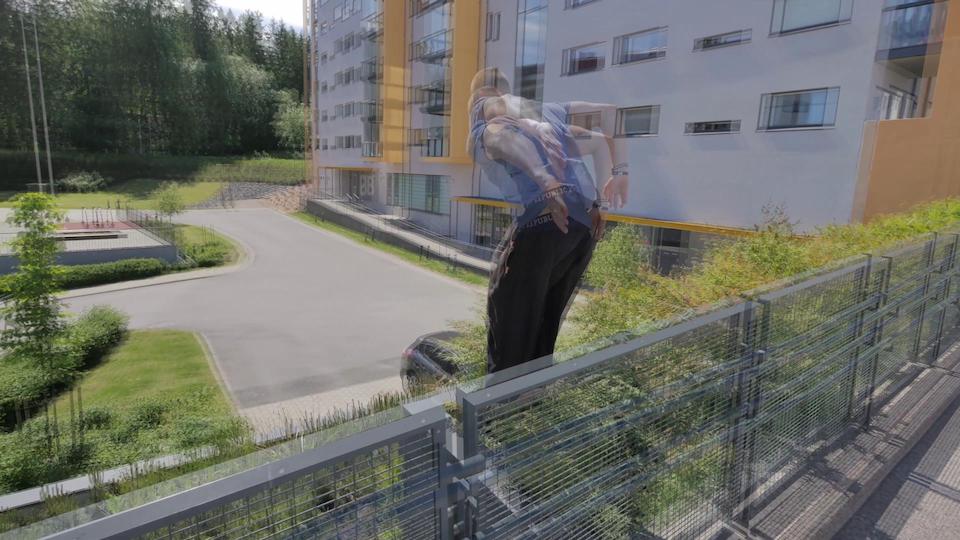} &
	\includegraphics[width=\linewidth]{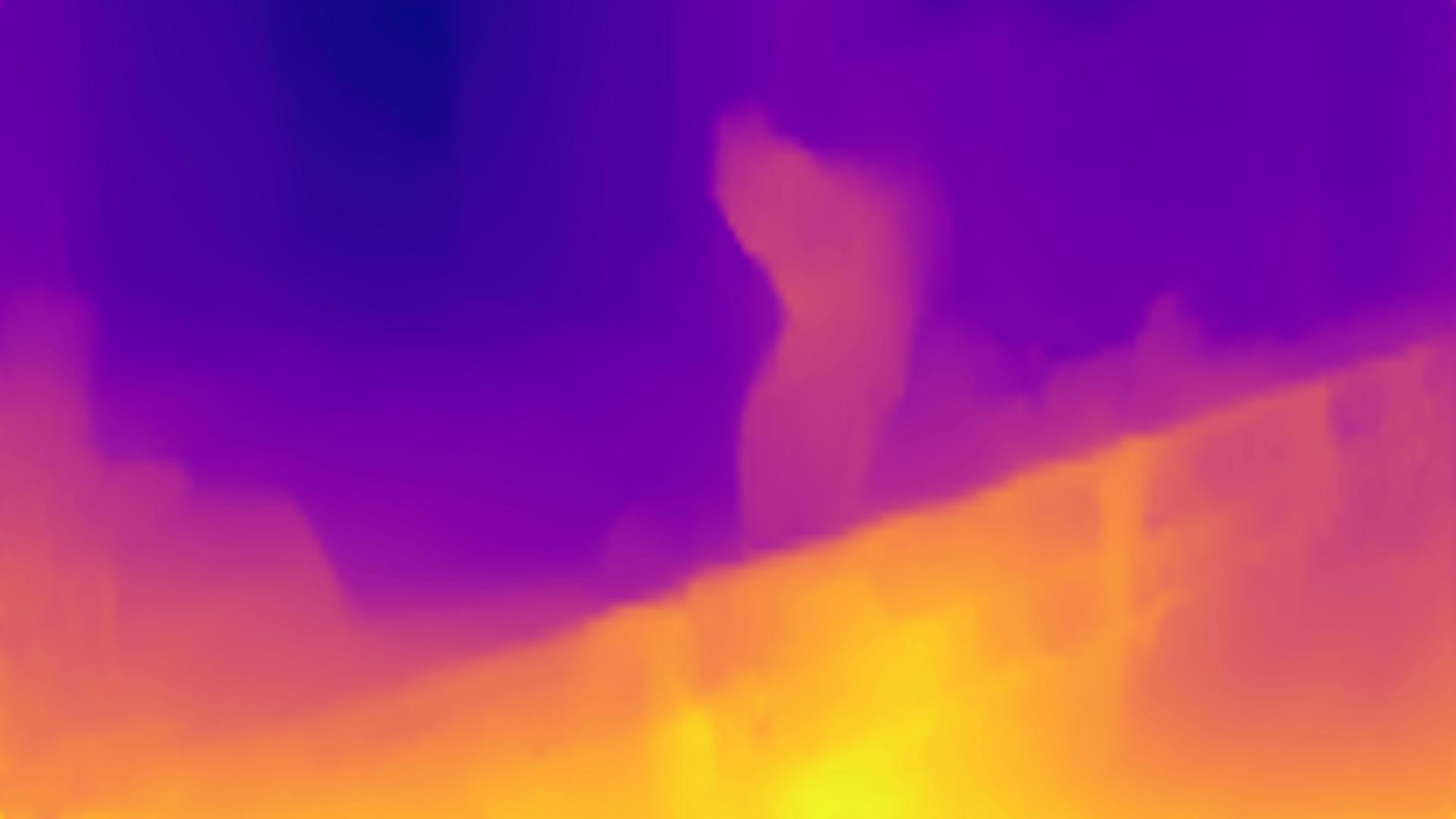} &
	\includegraphics[width=\linewidth]{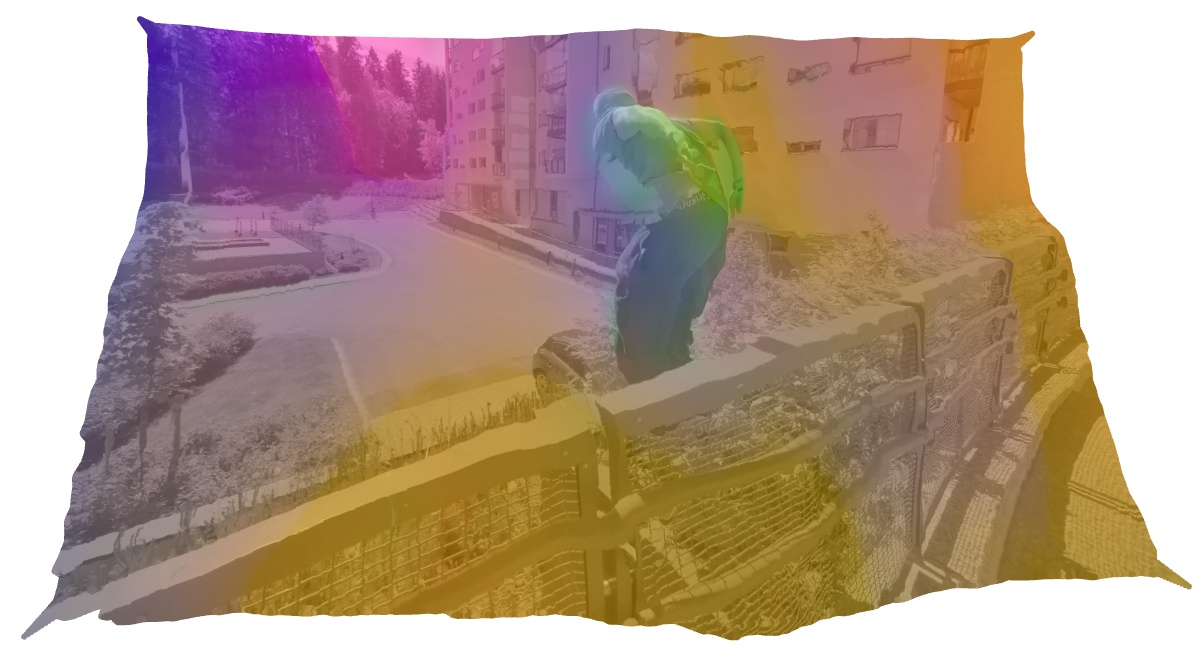} &
	& \includegraphics[width=\linewidth]{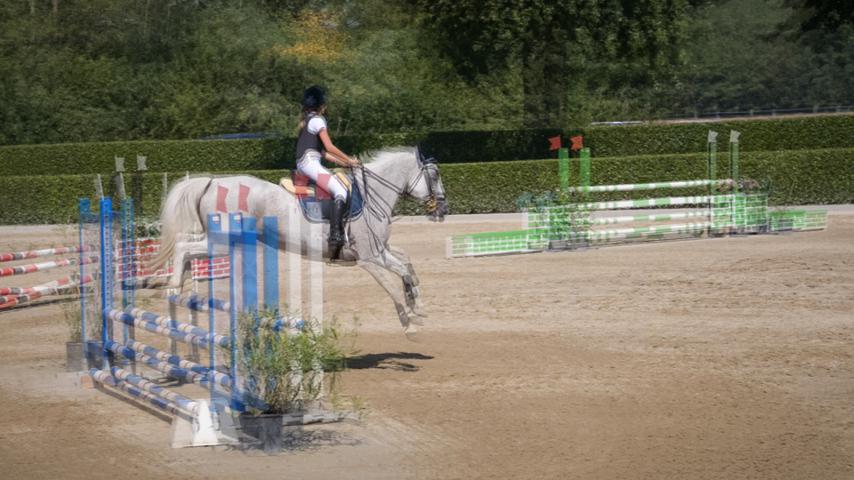} &
	\includegraphics[width=\linewidth]{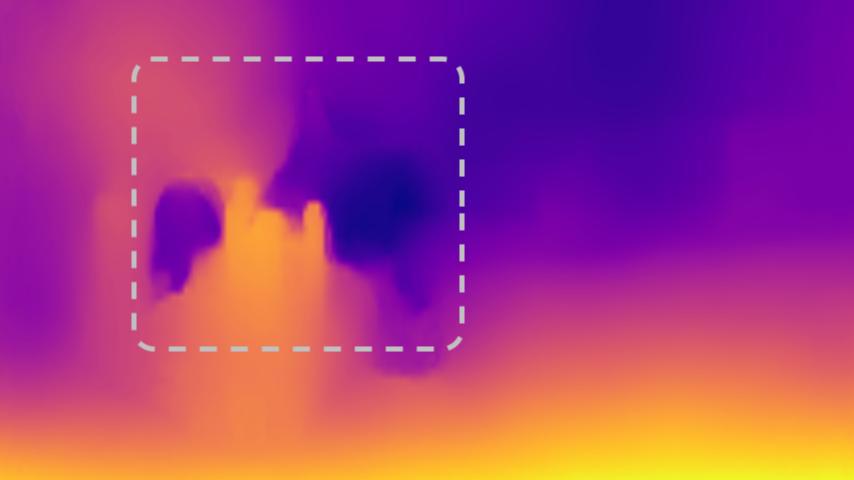} &
	\includegraphics[width=\linewidth]{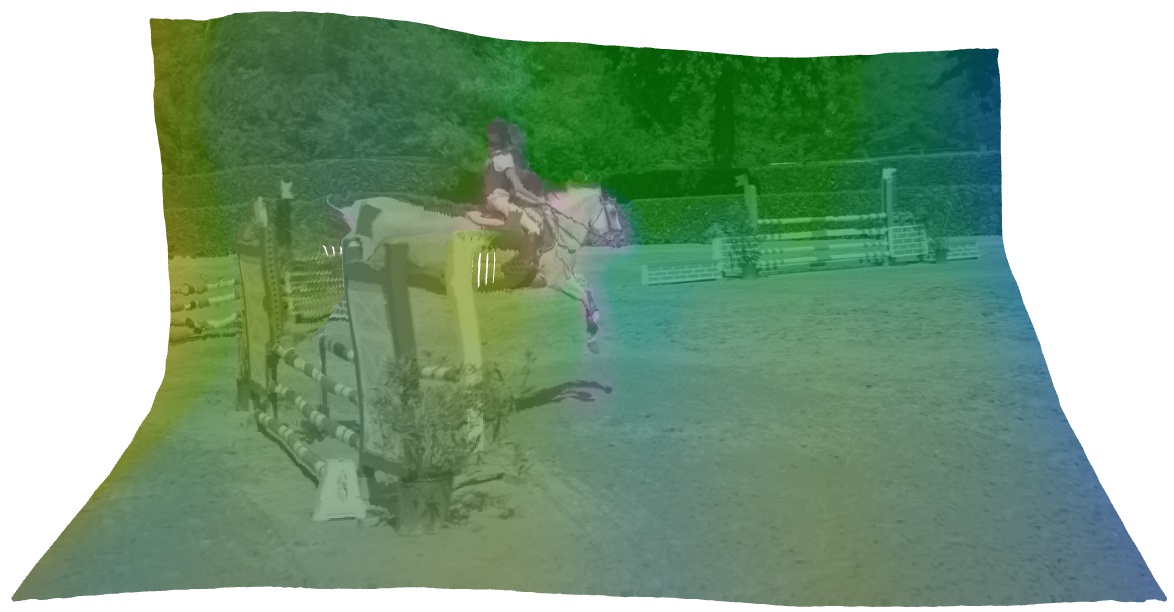} \\

	& \includegraphics[width=\linewidth]{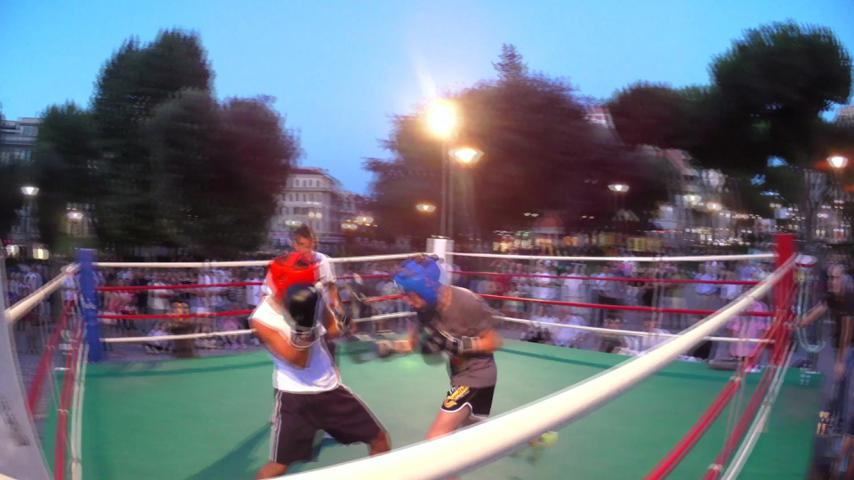} &
	\includegraphics[width=\linewidth]{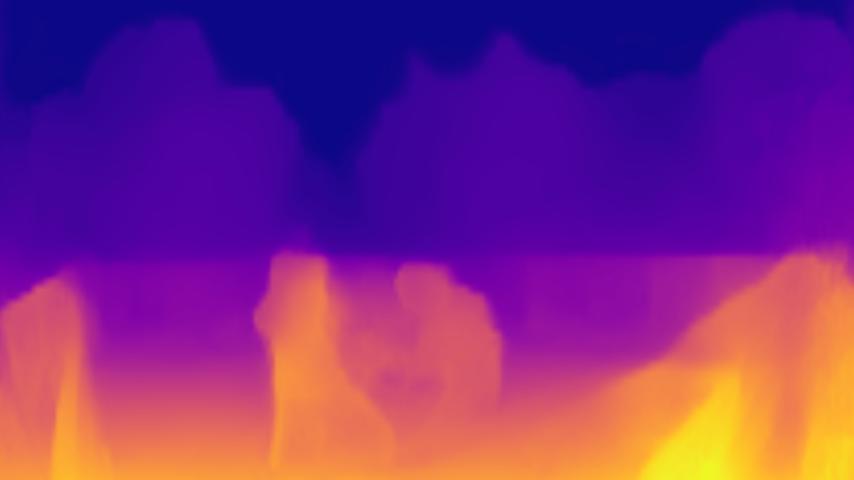} &
	\includegraphics[width=\linewidth]{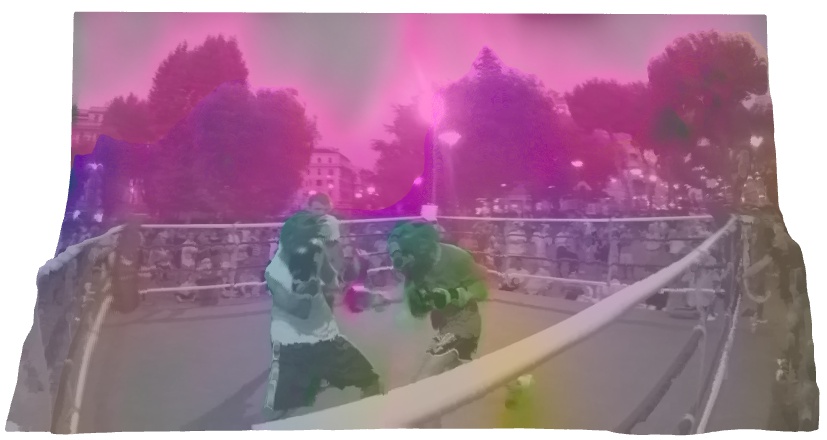} &
	& \includegraphics[width=\linewidth]{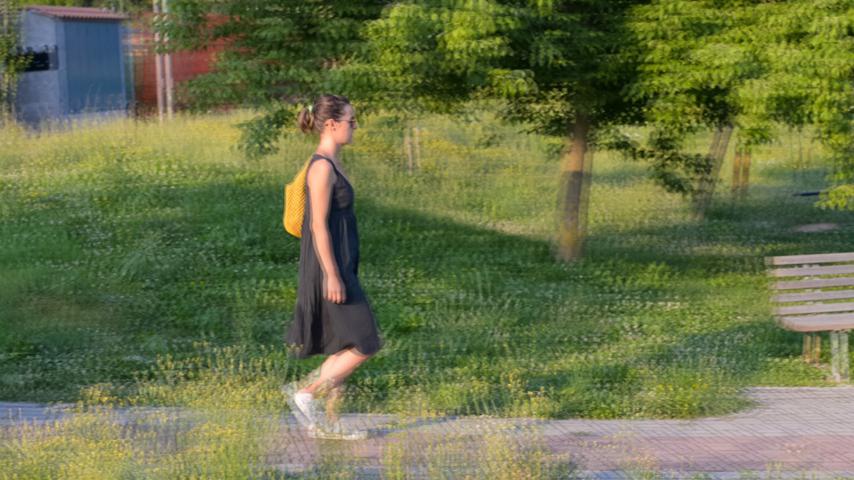} &
	\includegraphics[width=\linewidth]{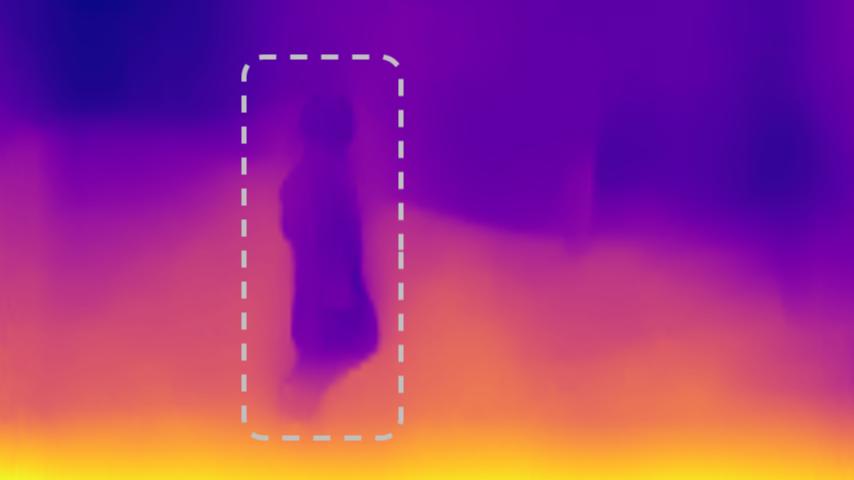} &
	\includegraphics[width=\linewidth]{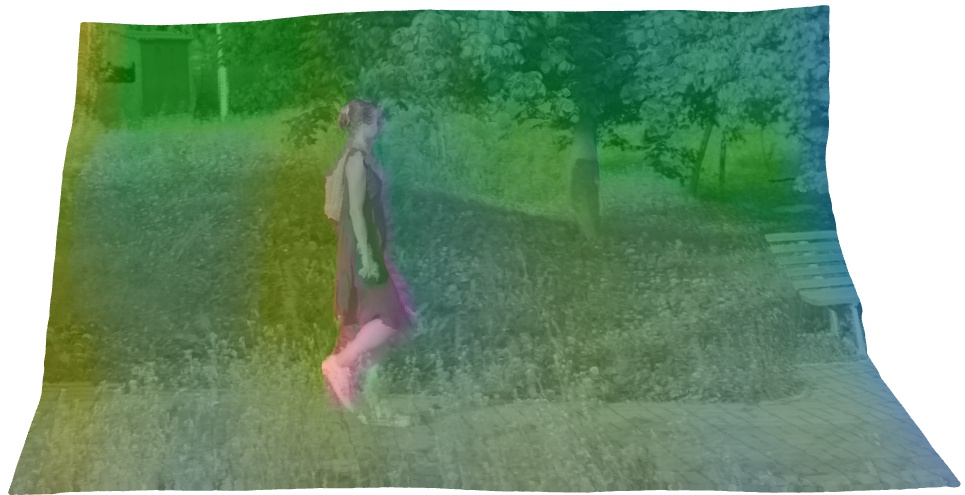} \\ [3.5em] \\

	% & \includegraphics[width=\linewidth]{figures/supp_generalization/davis/bike-packing_00012_overlay.jpg} &
	% \includegraphics[width=\linewidth]{figures/supp_generalization/davis/bike-packing_00012_disp.jpg} &
	% \includegraphics[width=0.95\linewidth]{figures/supp_generalization/davis/bike-packing_00012_vis.jpg} &
	% & \includegraphics[width=\linewidth]{figures/supp_generalization/davis/cat-girl_00078_overlay.jpg} &
	% \includegraphics[width=\linewidth]{figures/supp_generalization/davis/cat-girl_00078_disp.jpg} &
	% \includegraphics[width=\linewidth]{figures/supp_generalization/davis/cat-girl_00078_vis.jpg} \\ [3.5em] \\

	\multirow{3}{*}{\small  \rotatebox[origin=c]{90}{Driving \cite{Mayer:2016:ALD}}}
	& \includegraphics[width=\linewidth]{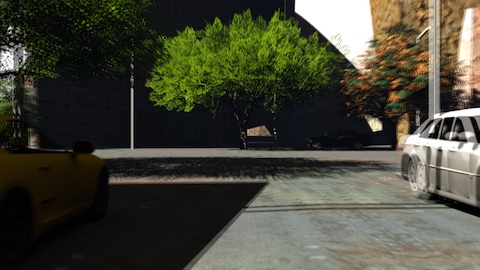} &
	\includegraphics[width=\linewidth]{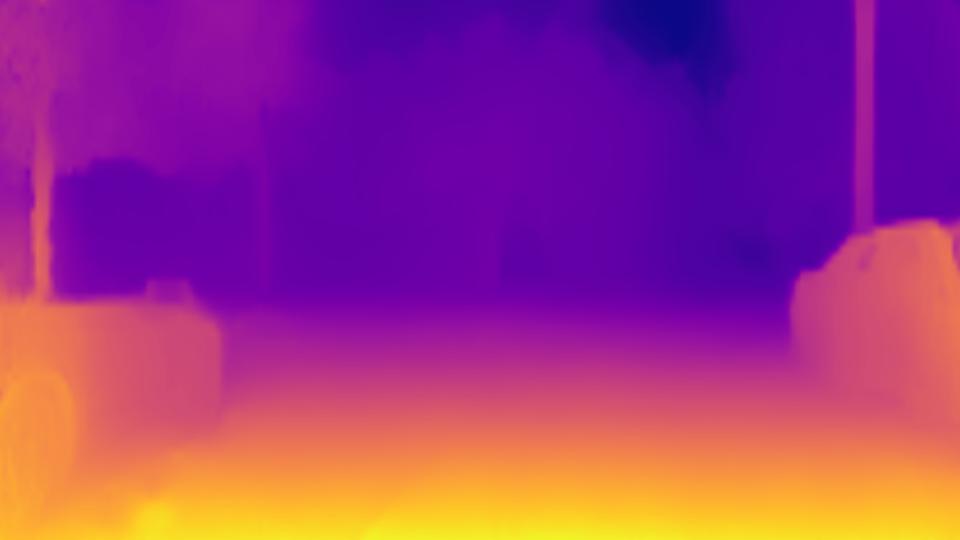} &
	\includegraphics[width=0.92\linewidth]{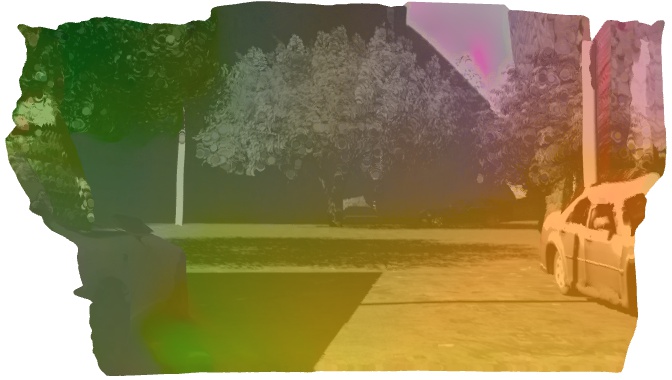} &
	& \includegraphics[width=\linewidth]{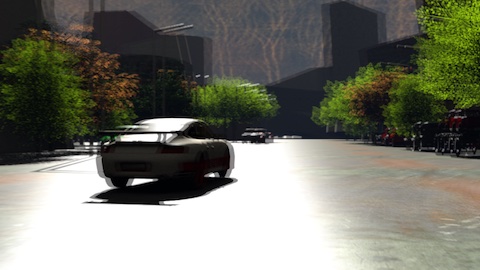} &
	\includegraphics[width=\linewidth]{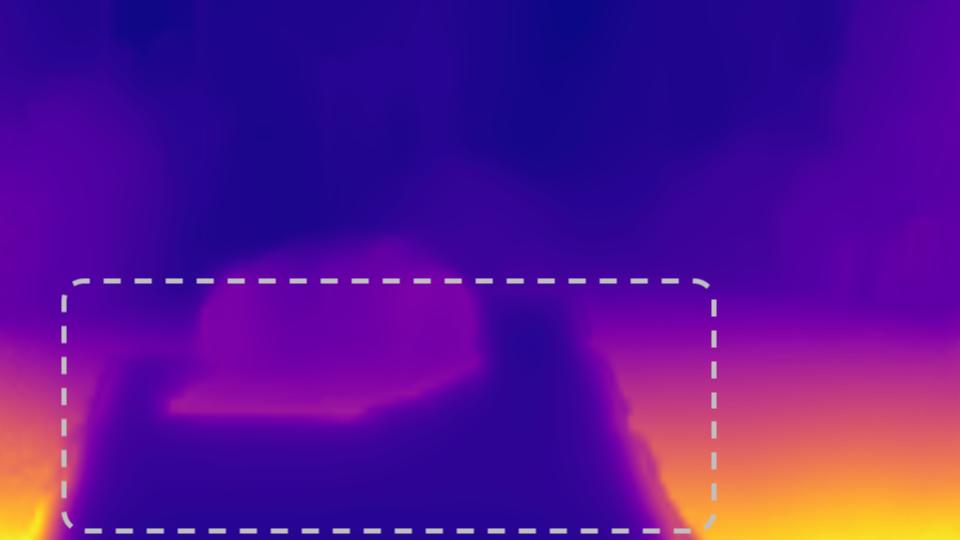} &
	\includegraphics[width=0.88\linewidth]{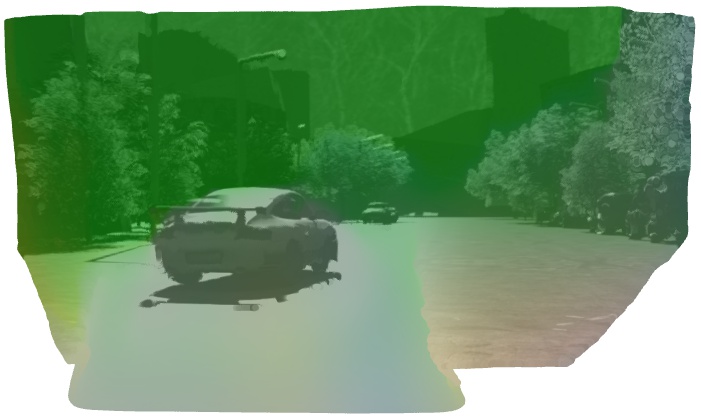} \\

	& \includegraphics[width=\linewidth]{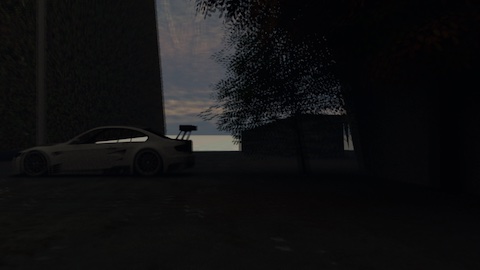} &
	\includegraphics[width=\linewidth]{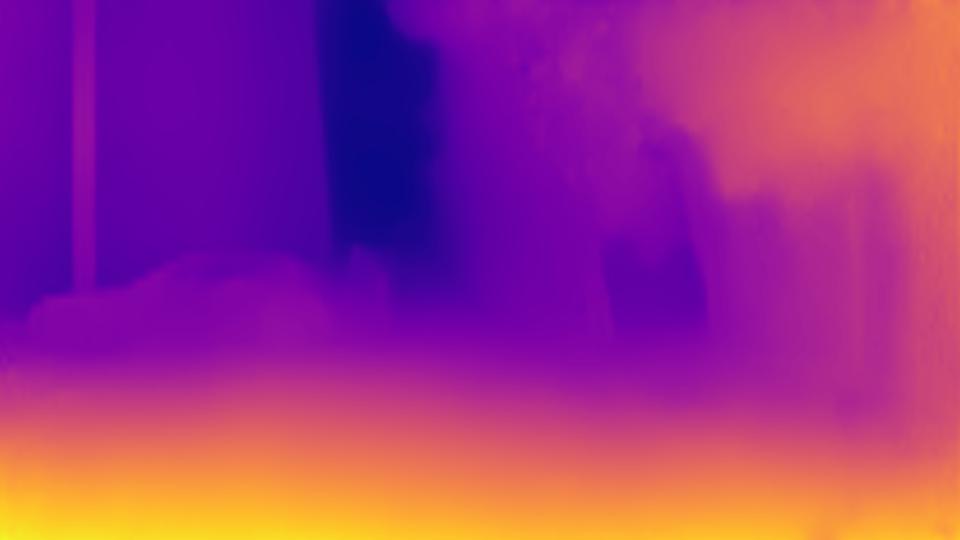} &
	\includegraphics[width=0.9\linewidth]{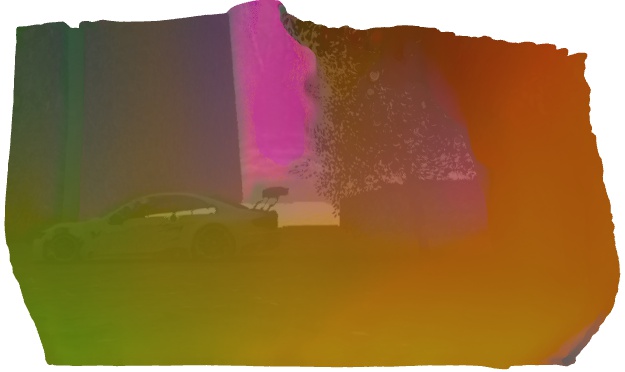} &
	& \includegraphics[width=\linewidth]{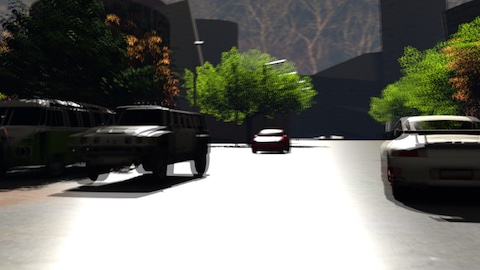} &
	\includegraphics[width=\linewidth]{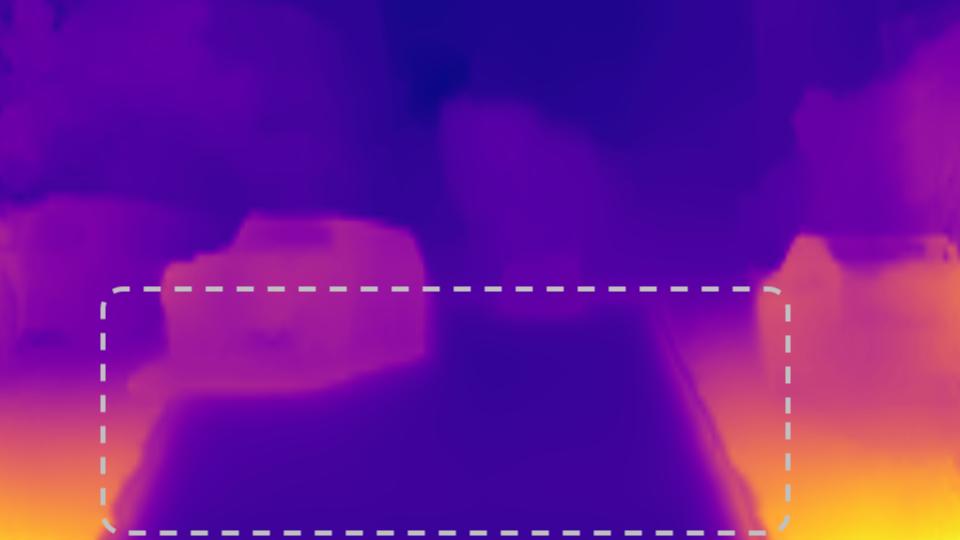} &
	\includegraphics[width=0.9\linewidth]{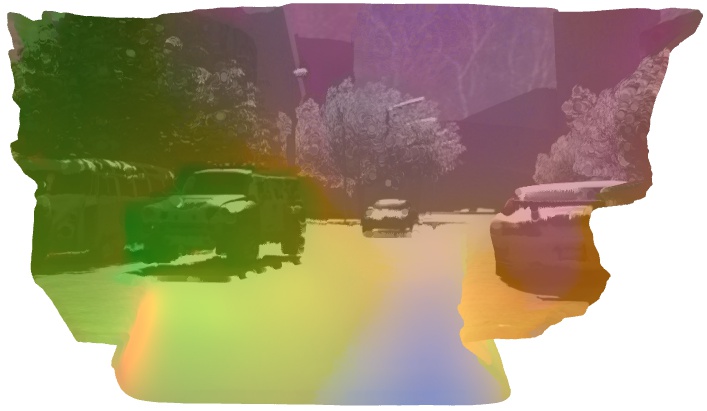} \\ [3.5em] \\

	\multirow{3}{*}{\small  \rotatebox[origin=c]{90}{Monkaa \cite{Mayer:2016:ALD}}}
	& \includegraphics[width=\linewidth]{figures/supp_generalization/monkaa/funnyworld_camera2_x2_0459_overlay.jpg} &
	\includegraphics[width=\linewidth]{figures/supp_generalization/monkaa/funnyworld_camera2_x2_0459_disp.jpg} &
	\includegraphics[width=\linewidth]{figures/supp_generalization/monkaa/funnyworld_camera2_x2_0459_vis.jpg} &
	& \includegraphics[width=\linewidth]{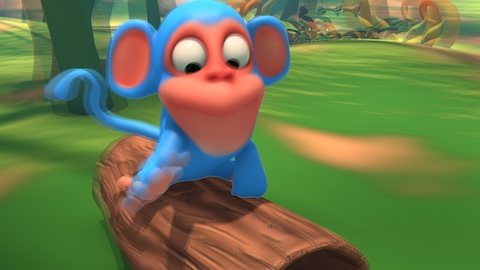} &
	\includegraphics[width=\linewidth]{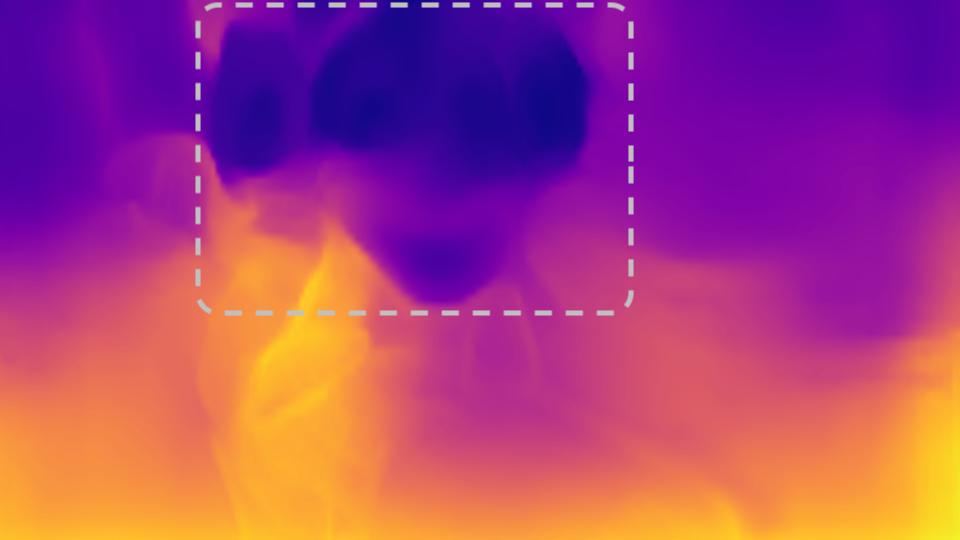} &
	\includegraphics[width=0.87\linewidth]{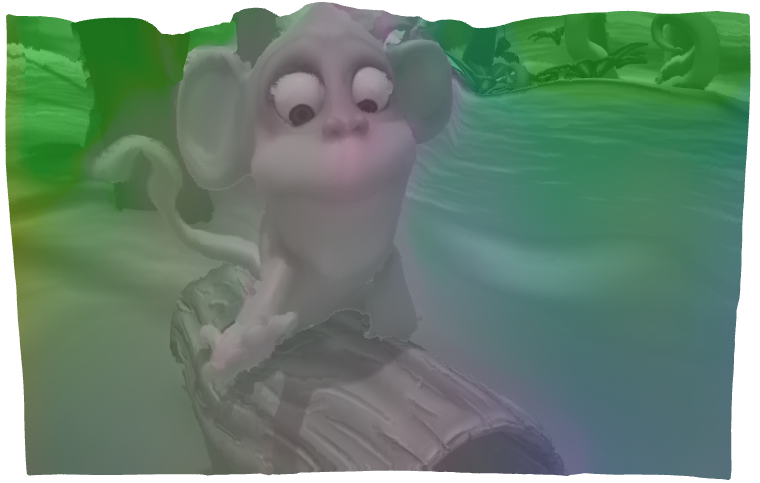} \\

	& \includegraphics[width=\linewidth]{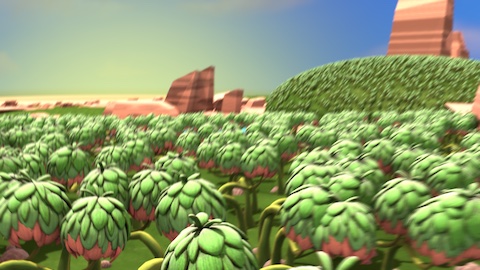} &
	\includegraphics[width=\linewidth]{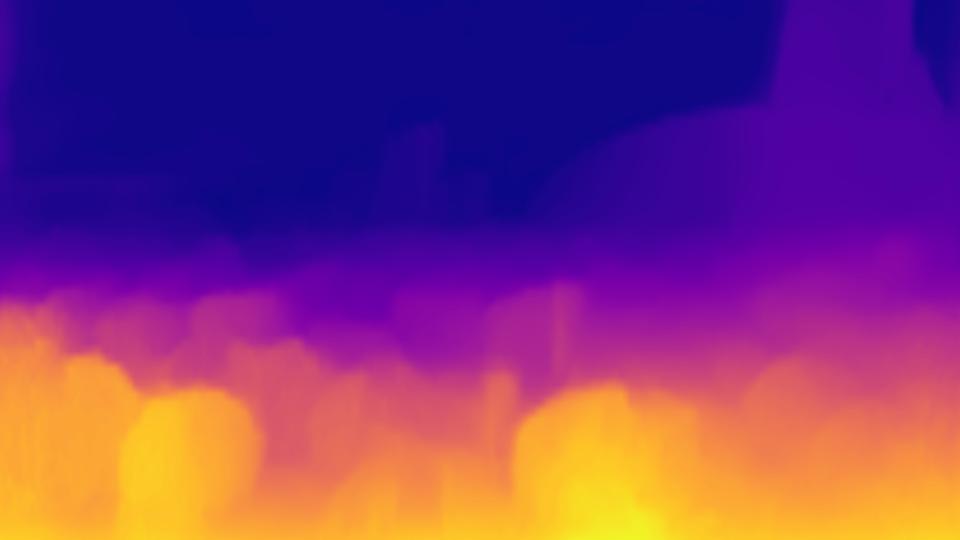} &
	\includegraphics[width=0.9\linewidth]{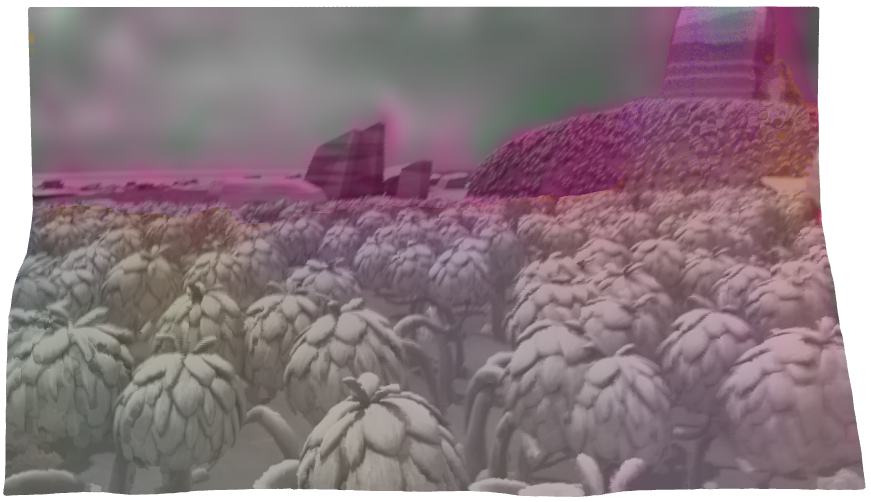} &
	& \includegraphics[width=\linewidth]{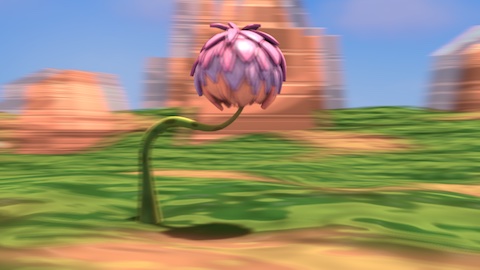} &
	\includegraphics[width=\linewidth]{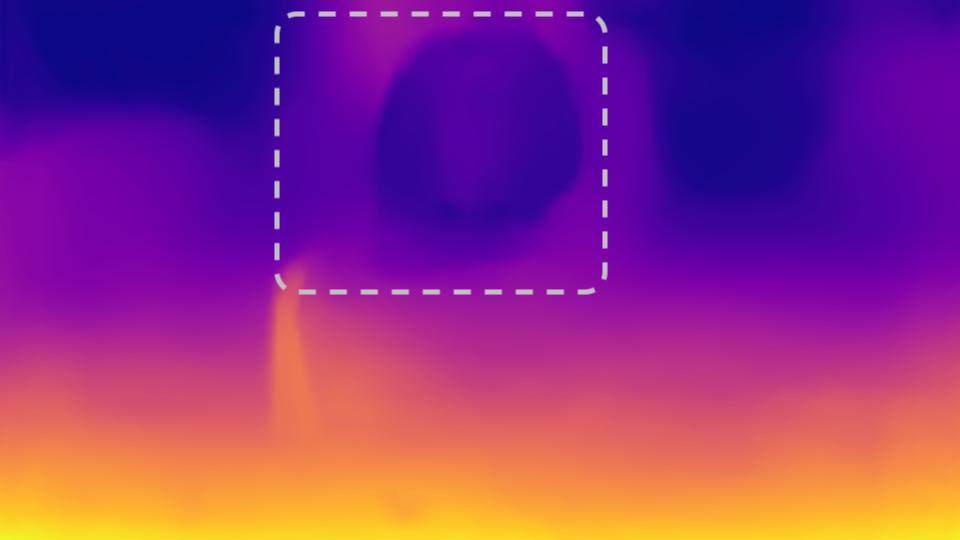} &
	\includegraphics[width=0.9\linewidth]{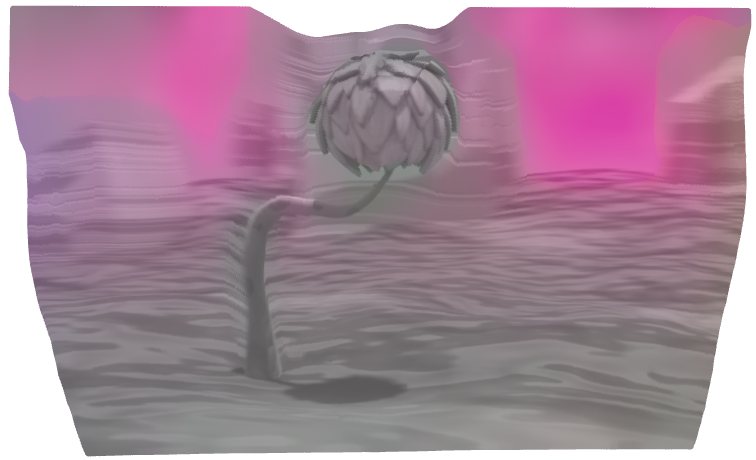} \\ [3.5em] \\

	& (a) Overlayed input images & (b) Depth map & (c) Scene flow visualization & & (a) Overlayed input images & (b) Depth map & (c) Scene flow visualization \\ [1em]
\end{tabular}
\caption{\textbf{Generalization to nuScenes, DAVIS, Driving, and Monkaa datasets}: Each scene shows \emph{(a)} overlayed input images, \emph{(b)}~depth, and \emph{(c)} 3D scene flow visualization. The \emph{left side} demonstrates good generalization of our model to nuScenes, DAVIS, Driving, and Monkaa datasets. Nonetheless, failure cases do exist, provided on the \emph{right side} and highlighted with dashed gray squares. A typical failure mode is inaccurate depth estimation of foreground objects that are not seen in the training set as well as reflective road surfaces.}
\label{fig:supp:generalization}
\vspace{-0.5em}
\end{figure*}

\section{Generalization to Other Datasets}
\label{supp:subsec:generalization}

\begin{table}[]
\centering
\footnotesize
\setlength\tabcolsep{3pt}
% \rotatebox[origin=c]{90}{Occ}
\begin{tabular*}{\columnwidth}{@{\extracolsep{\fill}}c@{\hskip 5em}S[table-format=2.3]S[table-format=2.3]@{}}
	\toprule
	Dataset & {Optical flow EPE} & {Scene flow EPE} \\
	\midrule
	Driving & 3.6613 & 0.8845 \\
	Monkaa & 12.8881 & 4.2982 \\
	\bottomrule
\end{tabular*}
\caption{\textbf{Scene flow accuracy on Driving and Monkaa datasets \cite{Mayer:2016:ALD} using the End-Point Error (EPE) metric}: The accuracy of our model is generally low in the synthetic domain but is better on Driving than on Monkaa.}
\label{table:driving_monkaa_eval}
\vspace{-0.5em}
\end{table}

\begin{figure*}[!t]
\centering
\scriptsize
\setlength\tabcolsep{0.1pt}
\renewcommand{\arraystretch}{0.2}
\begin{tabular}{>{\centering\arraybackslash}m{.158\textwidth} >{\centering\arraybackslash}m{.158\textwidth} >{\centering\arraybackslash}m{.158\textwidth}
				>{\centering\arraybackslash}m{.015\textwidth} >{\centering\arraybackslash}m{.158\textwidth} >{\centering\arraybackslash}m{.158\textwidth} >{\centering\arraybackslash}m{.158\textwidth}}

	\includegraphics[width=\linewidth]{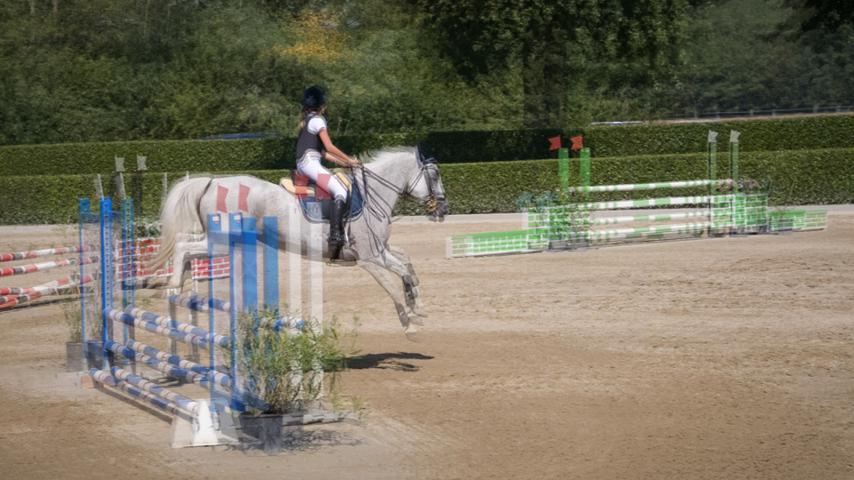} &
	\includegraphics[width=\linewidth]{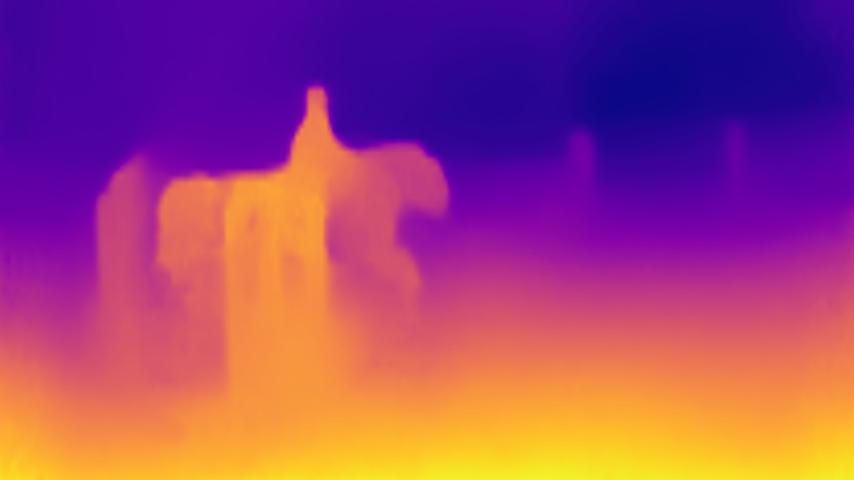} &
	\includegraphics[width=\linewidth]{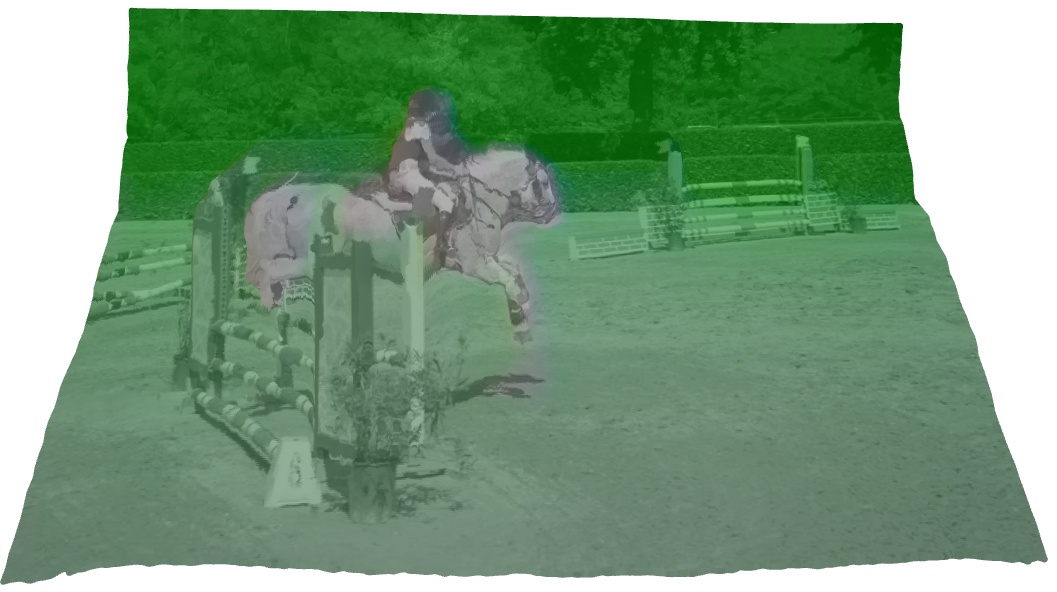} & &

	\includegraphics[width=\linewidth]{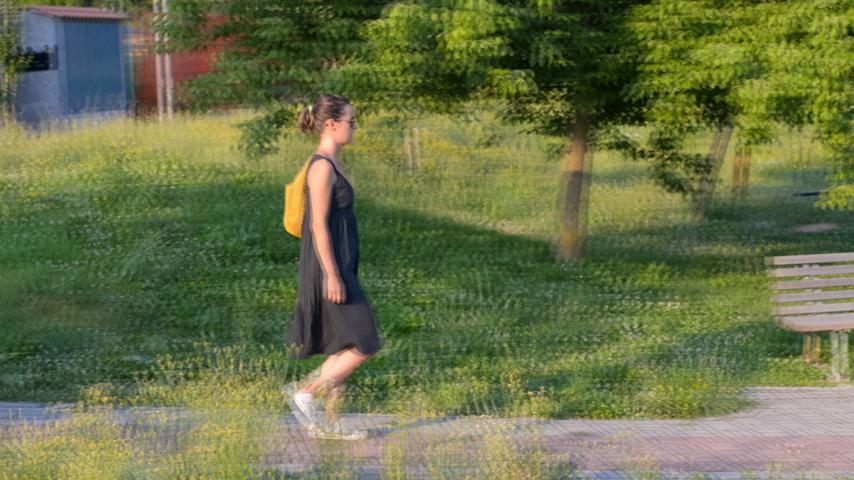} &
	\includegraphics[width=\linewidth]{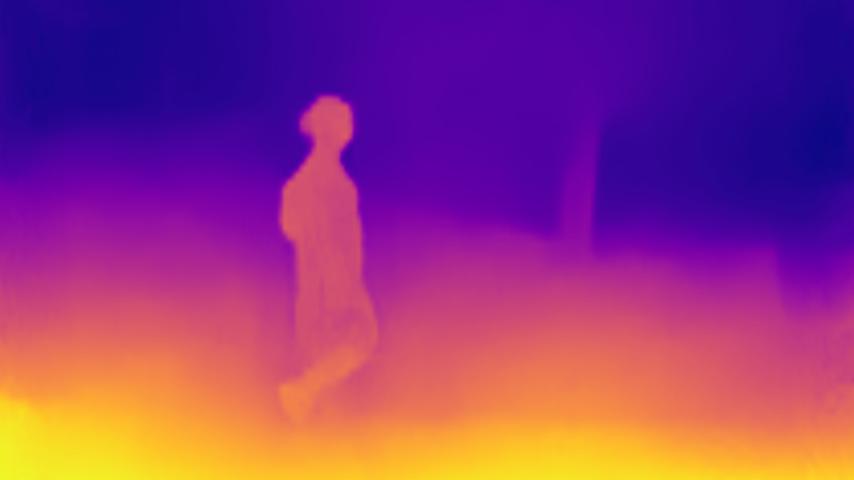} &
	\includegraphics[width=\linewidth]{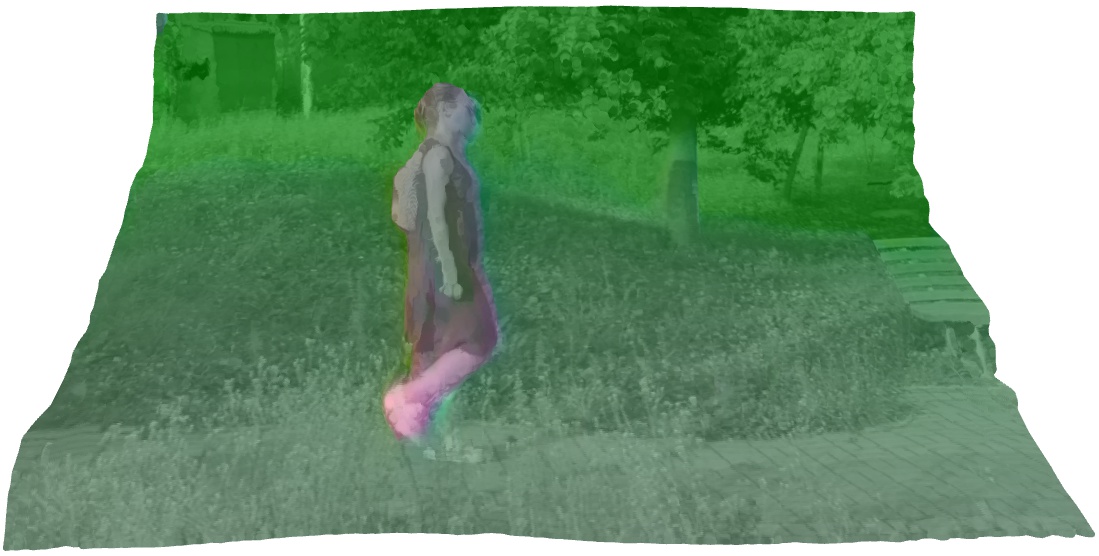} \\

	\includegraphics[width=\linewidth]{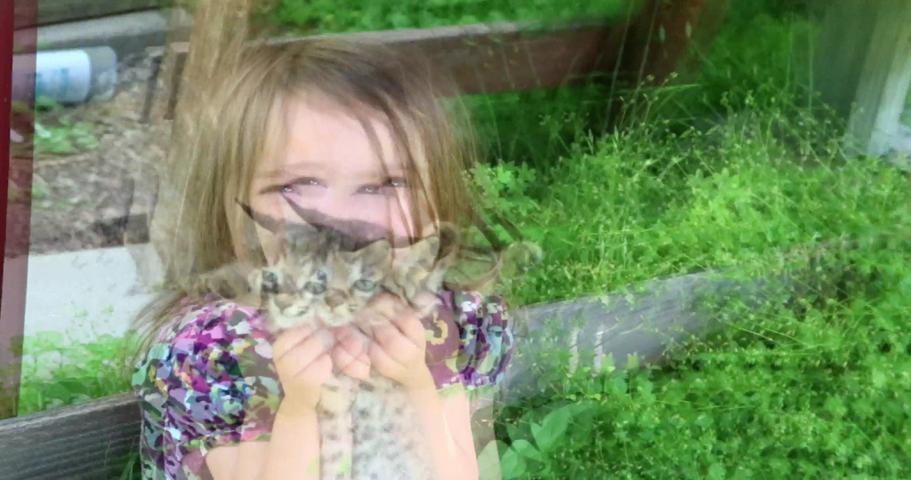} &
	\includegraphics[width=\linewidth]{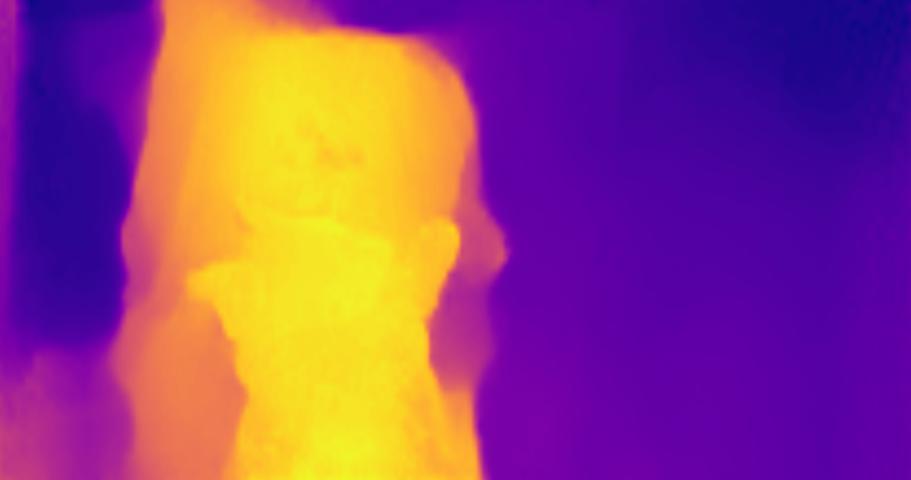} &
	\includegraphics[width=0.84\linewidth]{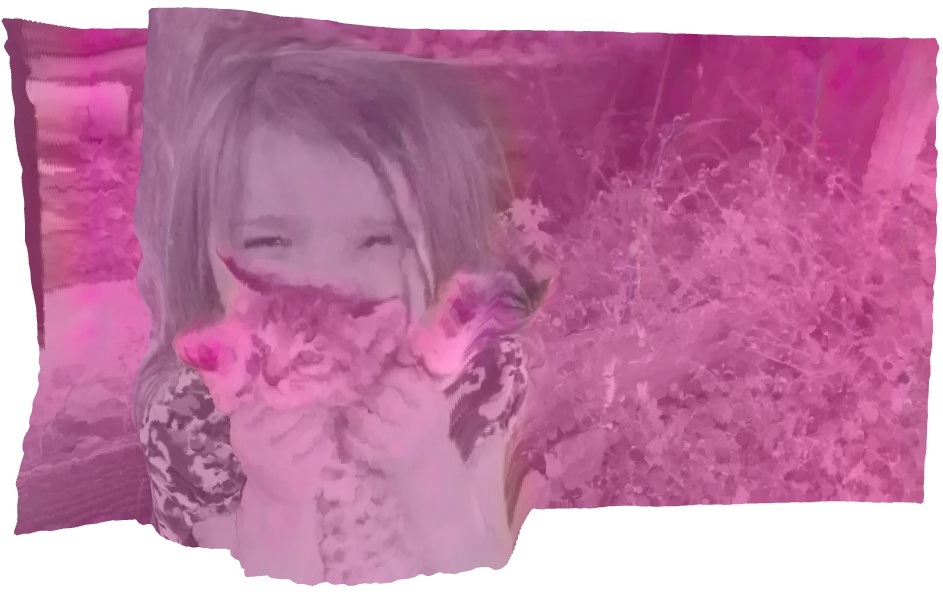} & &

	\includegraphics[width=\linewidth]{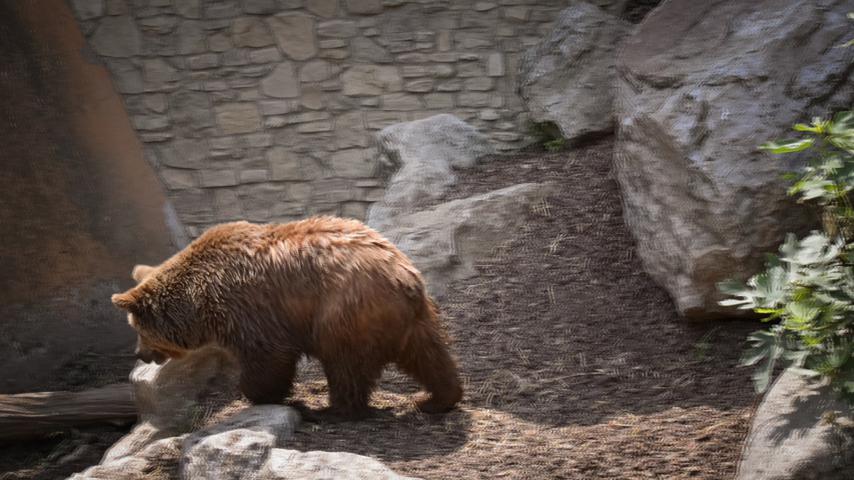} &
	\includegraphics[width=\linewidth]{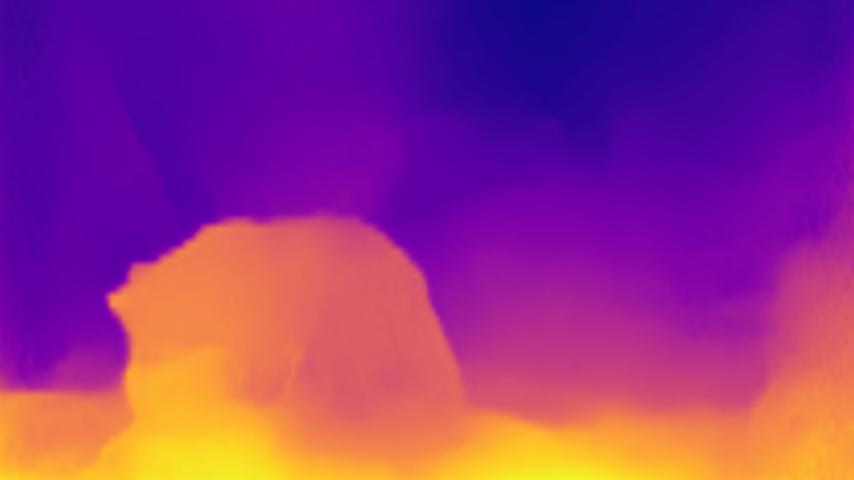} &
	\includegraphics[width=\linewidth]{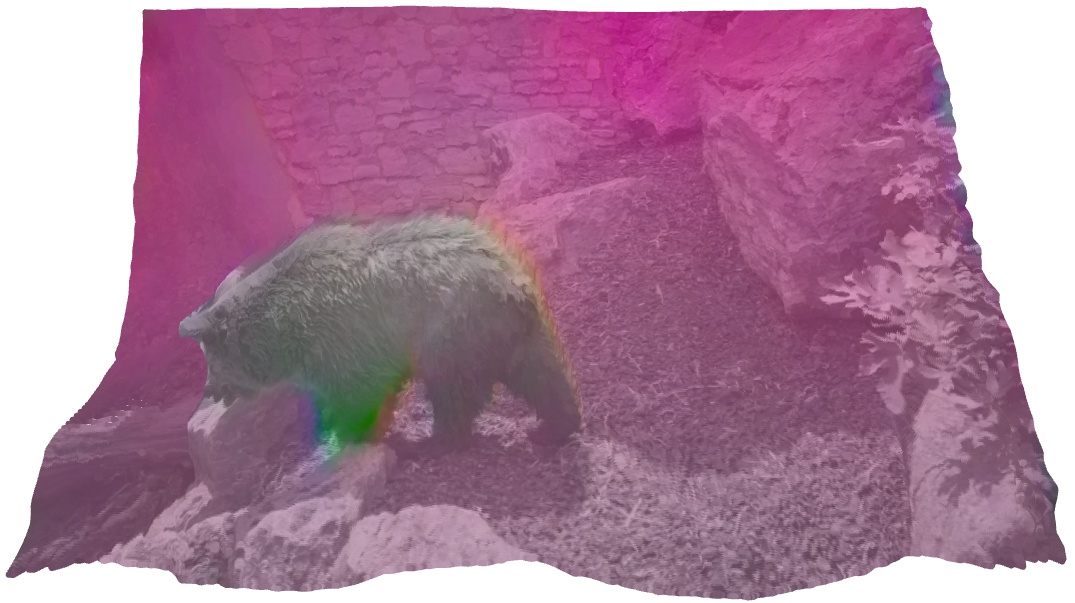} \\

	\includegraphics[width=\linewidth]{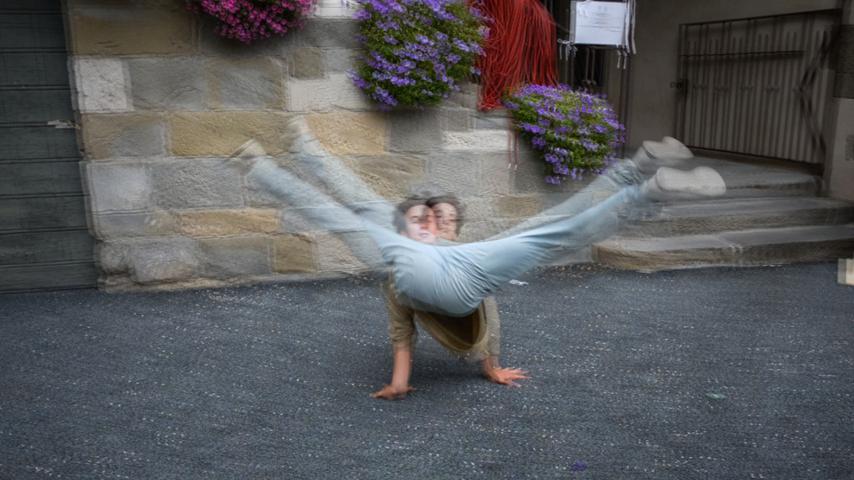} &
	\includegraphics[width=\linewidth]{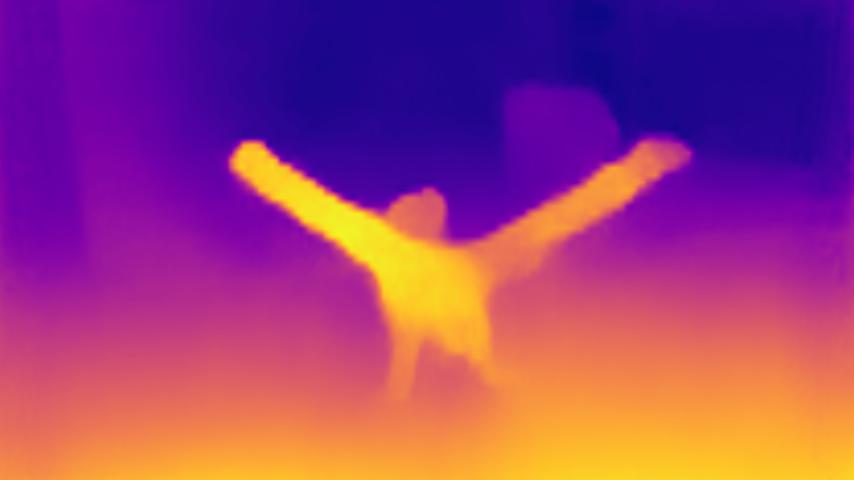} &
	\includegraphics[width=\linewidth]{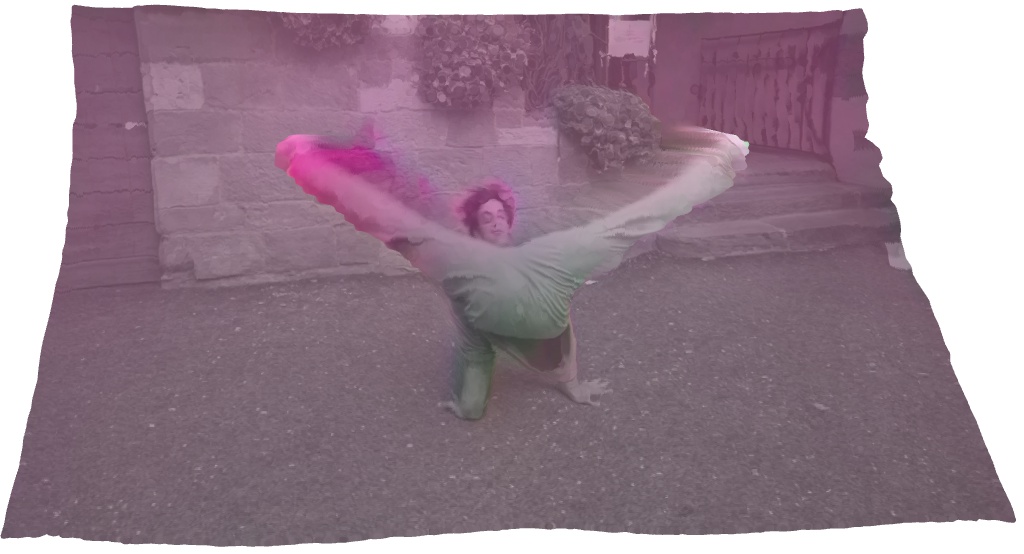} & &

	\includegraphics[width=\linewidth]{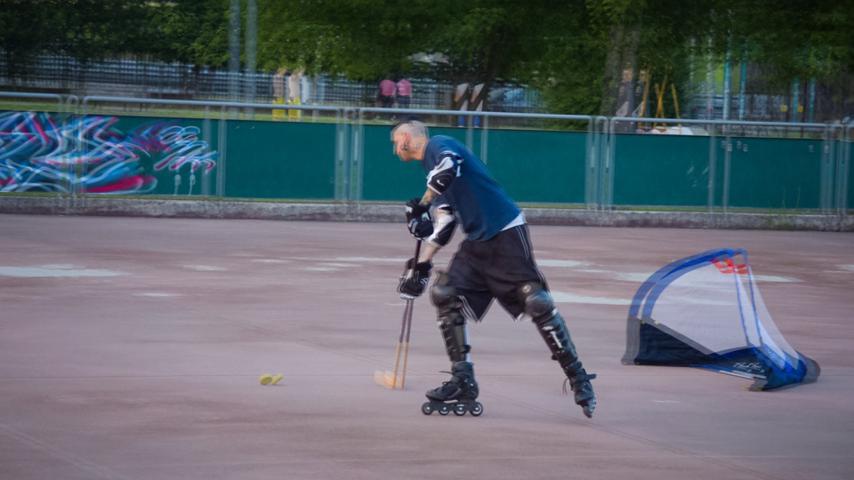} &
	\includegraphics[width=\linewidth]{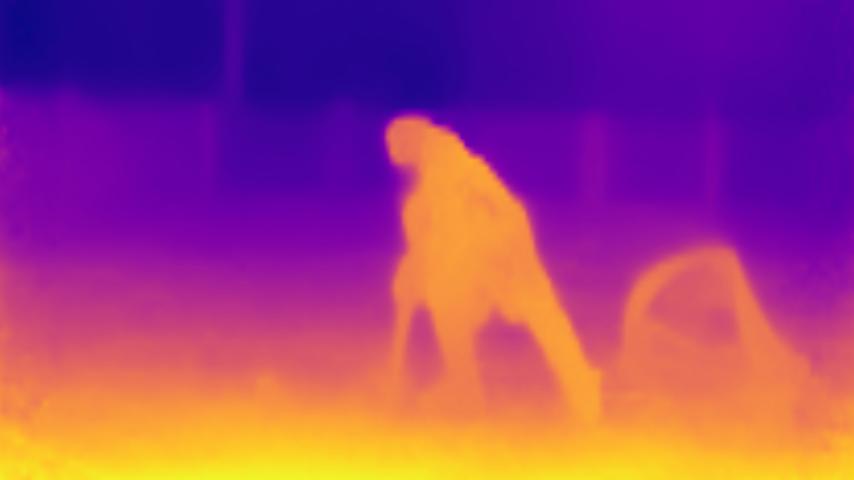} &
	\includegraphics[width=\linewidth]{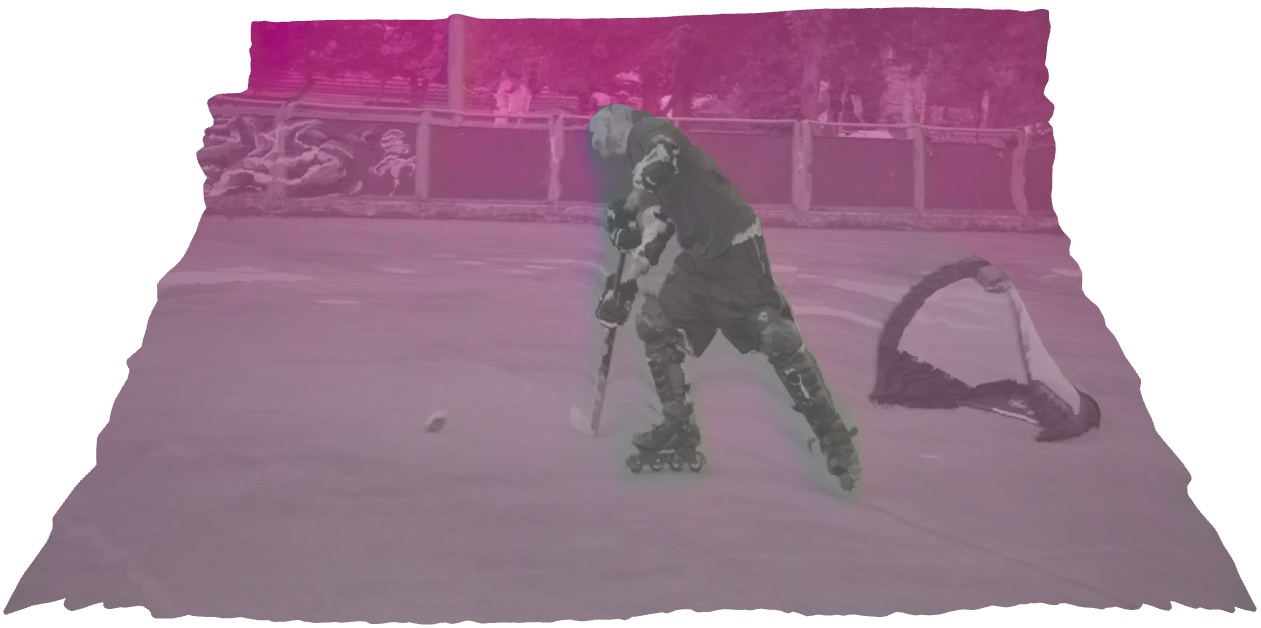} \\

	\includegraphics[width=\linewidth]{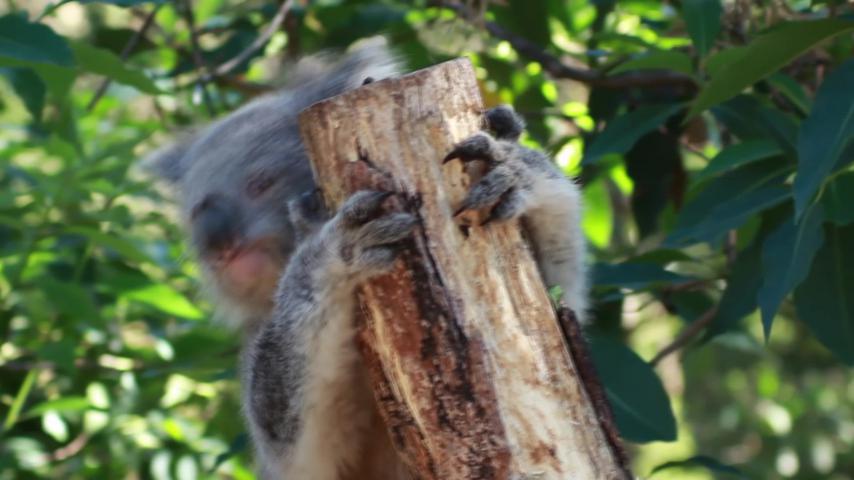} &
	\includegraphics[width=\linewidth]{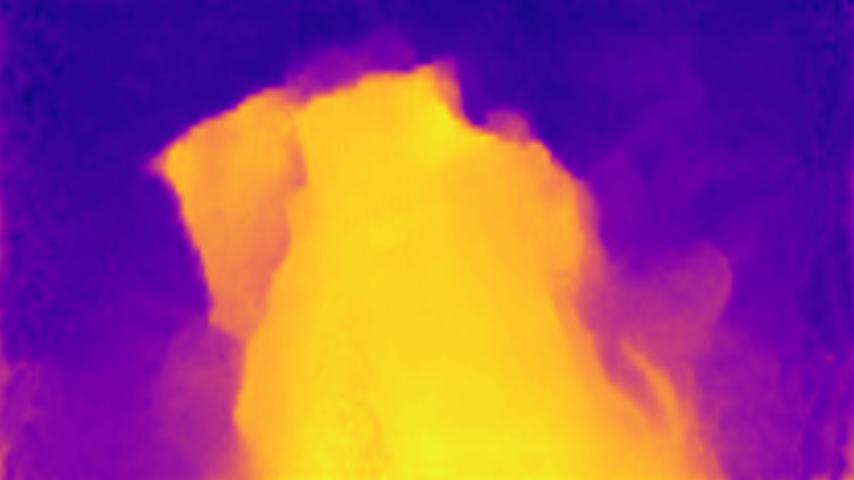} &
	\includegraphics[width=0.9\linewidth]{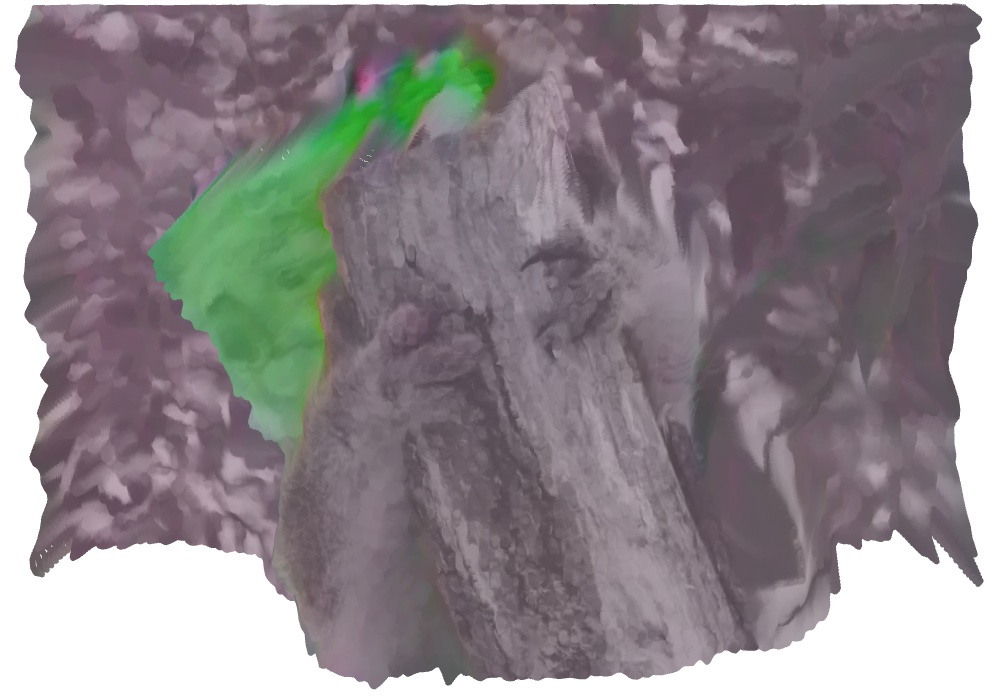} & &

	\includegraphics[width=\linewidth]{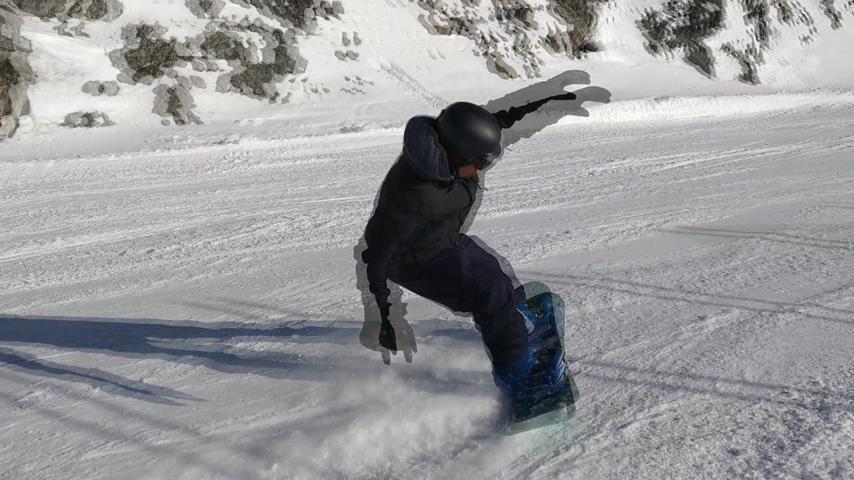} &
	\includegraphics[width=\linewidth]{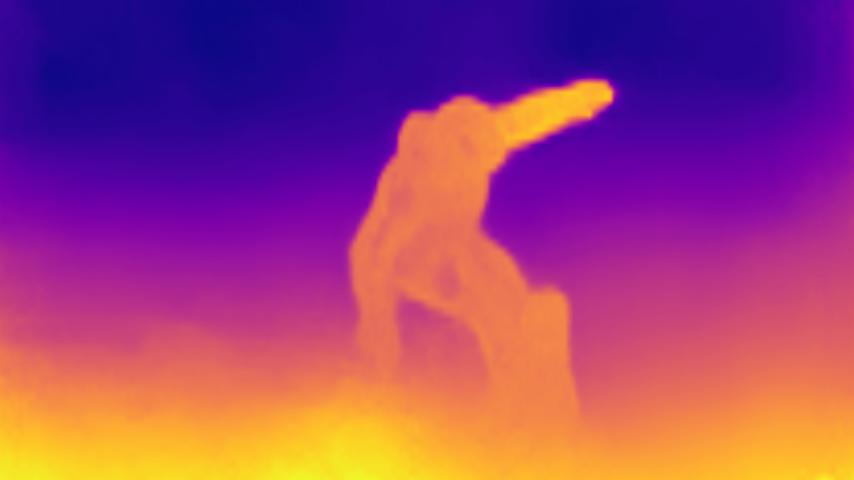} &
	\includegraphics[width=\linewidth]{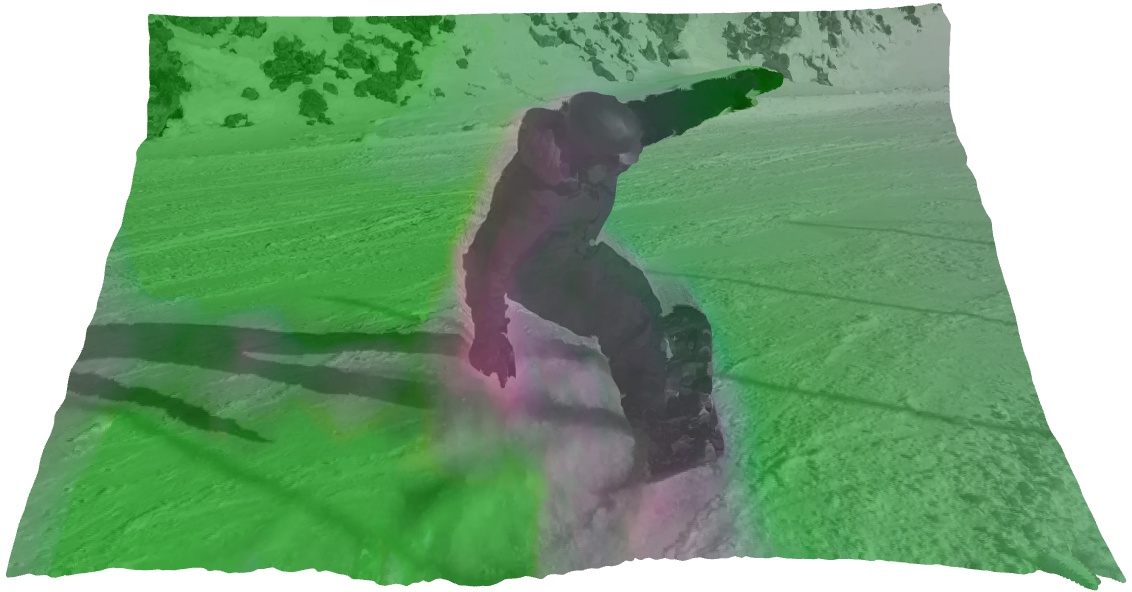} \\

	(a) Overlayed input images & (b) Depth map & (c) Scene flow visualization & & (a) Overlayed input images & (b) Depth map & (c) Scene flow visualization \\ [1em]
\end{tabular}
\caption{\textbf{Self-supervised learning in the wild and generalization to DAVIS dataset \cite{Perazzi:2016:ABD}}: Our model trained on the WSVD dataset \cite{Wang:2019:WSV} (a large amount of web videos) generalizes well to diverse scenes from the DAVIS dataset \cite{Perazzi:2016:ABD}.}
\label{fig:supp:in_the_wild}
\vspace{-0.5em}
\end{figure*}

Continuing from \cref{subsec:qual_comp} of the main paper, we provide more qualitative results on the nuScenes \cite{Caesar:2020:NSA}, DAVIS \cite{Perazzi:2016:ABD}, Driving and Monkaa \cite{Mayer:2016:ALD} datasets.
\cref{fig:supp:generalization} provides both successful cases and failure cases on those three datasets, respectively.
Our model, trained only on the KITTI dataset, generalizes well to the nuScenes dataset \cite{Caesar:2020:NSA}, which is reasonably close in domain (\ie,~driving scenes).
However, there exist some failure cases with inaccurate depth estimation as well as occasional artifacts on the image boundary.
On the DAVIS \cite{Perazzi:2016:ABD} dataset, our model generalizes surprisingly well to completely unseen domains, yet depth estimation on unseen objects (\eg,~horse, cat) can sometimes fail.

In the synthetic domain (Driving and Monkaa \cite{Mayer:2016:ALD} datasets), however, our model demonstrates less accurate results, as can be expected.
Typical failure cases are again inaccurate depth estimation on completely unseen synthetic objects or reflective road surfaces.
In \cref{table:driving_monkaa_eval}, we evaluate the scene flow accuracy of our model on the Driving and Monkaa \cite{Mayer:2016:ALD} datasets, using the End-Point-Error (EPE) metric.
Though the accuracy is quite low in general, the accuracy on Driving is much better than that on Monkaa as can be expected.

These overall results suggest that the accuracy of our self-supervised model depends on the training domain as well as the presence of target objects in the training dataset.
From this observation, we can conclude that better generalization requires to train the model on a dataset with both diverse domains and objects.
% In the next section, we demonstate the result of training on a large amount of unlabled videos with diverse scenes and objects.

% !TEX root = ../arxiv.tex

\section{Self-Supervised Learning in the Wild}

Through self-supervised learning, our method can in principle leverage vast amounts of unlabeled stereo web videos.
However unlike training on a single, calibrated dataset (\eg, KITTI), this comes with several new technical challenges.
Each stereo video is captured with different camera intrinsics and stereo configurations, whose values are even unknown.
Without knowing them, the self-supervised loss in \cref{eq:total_loss} in the main paper cannot be directly applied because it assumes a fixed (or given) focal length and stereo baseline.
% depth scale is determined by the stereo pair with fixed (or given) baseline.
We provide preliminary experiments to assess the feasibility of this scenario.

To train the network despite these unknowns, we first assume all videos share the same focal length.
Then, we normalize the output disparity (say, $d_\text{norm}$) to be in a fixed, normalized scale and use it for the scene flow loss.
For the disparity loss, we further linearly transform the disparity $d_\text{norm}$ to match the actual disparity scale of each given stereo input:
\begin{equation}
d_\text{actual} = a_\text{scale} \cdot d_\text{norm} + b_\text{scale}.
\label{eq:scale_disparity}
\end{equation}
To obtain the coefficients $a_\text{scale}$ and $b_\text{scale}$, we estimate optical flow between the stereo pair using our network, take the horizontal flow as pseudo disparity $d_\text{pseudo}$, and use the least squares between the pseudo disparity $d_\text{pseudo}$ and the normalized disparity $d_\text{norm}$.
Though our network now outputs disparity and scene flow on a normalized scale, it is still able to estimate optical flow (by projecting scene flow to image coordinates) on the correct scale due to being supervised by the 2D view-synthesis proxy loss.

For our preliminary experiments, we use the WSVD dataset \cite{Wang:2019:WSV}, which is a collection of stereo videos from YouTube, for training and test on the DAVIS \cite{Perazzi:2016:ABD} dataset.

\myparagraph{Training dataset preparation.}
When training on such diverse data collected on the web, it is important to make sure that the dataset is free of outliers.
For preparing the training data, we carefully pre-process the WVSD dataset by first discarding videos with low resolution, poor image quality, texts, or watermarks.
We further discard videos having vertical disparity, lens distortion, and narrow stereo baselines for better stereo supervision.
Then, we check every frame and remove black-colored edges on the image boundaries, if applicable.
Also, we find that many videos contain static scenes; thus we sample every \nth{4} frame and \nth{2} sequence for having more dynamic motion in the training sequences.
This pre-processing step, in the end, results in \num{58}\si{\kilo} training images from about \num{1.5}\si{\mega} raw frames.
Given the pre-trained model on KITTI, we further train the model on this curated dataset for \num{300}\si{\kilo} iteration steps.

\myparagraph{Result and discussion.}
\cref{fig:supp:in_the_wild} demonstrates the test result on the DAVIS \cite{Perazzi:2016:ABD} dataset.
Comparing to our KITTI-trained model (\cf~\cref{fig:generalization}), our model trained on the WSVD dataset \cite{Wang:2019:WSV} is able to correctly estimate depth on diverse scenes from the DAVIS \cite{Perazzi:2016:ABD} dataset.
This also confirms our observation from the generalization analysis in \cref{supp:subsec:generalization} that better generalization can be achieved by training on a dataset with diverse scenes and objects.

However, our model unfortunately fails to correctly estimate scene flow: the network outputs $z$ components of the scene flow that are nearly zero and only estimates $x$ and $y$ components (refer the scene flow color coding in \cref{fig:color_coding}).
We also observe that the 3D reconstruction loss $L_\text{sf,pt}$ in \cref{eq:sf_3d_dist} does not converge at all.
We conjecture that the failure comes from the issue that the model does not estimate scale-consistent depth but instead normalized depth for each frame, which makes it difficult to determine the correct $z$ component of the scene flow for each sequence.
This connects to a current limitation that our approach requires a fixed (or given) focal length and a stereo baseline for training.
However, we expect that this can be overcome once the network is able to output scale-consistent depth across sequences or videos while being trained on videos with diverse camera settings.
We leave this for future work.

% \begin{figure*}[]
% \centering
% \scriptsize
% \setlength\tabcolsep{0.1pt}
% \renewcommand{\arraystretch}{0.2}
% \begin{tabular}{>{\centering\arraybackslash}m{.16\textwidth} >{\centering\arraybackslash}m{.16\textwidth} >{\centering\arraybackslash}m{.16\textwidth} }

{\small

\let\oldthebibliography=\thebibliography
\let\oldendthebibliography=\endthebibliography
\renewenvironment{thebibliography}[1]{%
     \oldthebibliography{#1}%
     \setcounter{enumiv}{ 80 }%
}{\oldendthebibliography}

\bibliographystyle{bib/ieee_fullname}

}

\end{document}